\documentclass[twoside,11pt]{article}

\usepackage{blindtext}
\usepackage{booktabs}
\usepackage{algorithm}  \usepackage{algpseudocode}
\usepackage{jmlr2e}
\usepackage{float}
\usepackage{booktabs}
\usepackage{multirow}
\usepackage{rotating}
\providecommand{\sci}[2]{$#1{\times}10^{#2}$}

\usepackage{amssymb}
\usepackage{amsmath}
\usepackage{diagbox}

\usepackage{lastpage}

\ShortHeadings{Why Rectified Power Unit Networks Fail and How to Improve It}{Kim and Kang}
\firstpageno{1}

\begin{document}

\title{Why Rectified Power Unit Networks Fail and How to Improve It: An Effective Field Theory Perspective}

\author{\name Taeyoung Kim \email  taeyoungkim@kias.re.kr \\
       \addr Center for AI and Natural Sciences\\
        Korea Institute for Advanced Study\\
       Seoul 02455, South Korea
       \AND
       \name Myungjoo Kang \email mkang@snu.ac.kr \\
       \addr Department of Mathematical Science\\
       Seoul National University\\
       Seoul 08826, South Korea}

\maketitle

\begin{abstract}
The Rectified Power Unit (RePU) activation function, a differentiable generalization of the Rectified Linear Unit (ReLU), has shown promise in constructing neural networks due to its smoothness properties. However, deep RePU networks often suffer from critical issues such as vanishing or exploding values during training, rendering them unstable regardless of hyperparameter initialization. Leveraging the perspective of effective field theory, we identify the root causes of these failures and propose the Modified Rectified Power Unit (MRePU) activation function. MRePU addresses RePU’s limitations while preserving its advantages, such as differentiability and universal approximation properties. Theoretical analysis demonstrates that MRePU satisfies the criticality conditions in the limit of a small kernel, and a finite-kernel analysis yields an initialization rule for deep MRePU networks. Experiments on polynomial regression, physics-informed neural networks (PINNs) and real-world vision tasks show that MRePU networks initialized by this rule remain trainable at depths at which RePU networks fail, while their accuracy relative to standard activations depends on the task.
\end{abstract}


\section{Introduction}
\label{sec1}
\subsection{Activation Functions}
\label{subsec1.1}
An activation function is a function that acts on the output of each layer of a neural network. The remarkable success of deep neural networks is closely related to the choice of appropriate nonlinear activation functions, and there has been extensive discussion solely on the research of activation functions. Early studies on neural networks include the perceptron (\cite{rosenblatt58}), where the activation function is the sign function, and it was composed of only one or two layers. The perceptron had the drawback that its derivative was almost everywhere zero, making it difficult to apply the backpropagation algorithm. \citet{cybenko89} showed that networks with a continuous sigmoidal activation have the universal approximation property, and \citet{Leshno93} showed that any continuous activation that is not a polynomial suffices.
In the 1980s and 1990s, smooth activations like the sigmoid activation and tanh activation, along with their variations, were proposed (\cite{Rumelhart86}, \cite{Narayan97}, \cite{LeCun02}). These activations had the disadvantage of gradient vanishing when stacking layers deeply. To overcome this drawback, the Rectified Linear Unit (ReLU) was proposed by (\cite{Nair10}), whose derivative is a non-zero constant. Variations of ReLU, such as PReLU, GELU, and ELU, have also been proposed in the 2010s (\cite{He15}, \cite{Hendrycks16}, \cite{Clervert15}).
Recently, there have been attempts to find activations tailored to specific tasks to impart inductive bias to the neural network structure according to the task (\cite{Ramachandran17}, \cite{Sun24}).

\subsection{Rectified Power Unit}
\label{subsec1.2}
One of the attempts to provide good regularity and inductive bias to the architecture is the Rectified Power Unit (RePU), a generalization of ReLU (\cite{Li20}). Early results using RePU include the application of ReQU, a type of RePU, in the DeepRitz method, which is a neural network approach for solving Partial Differential Equations (PDEs) (\cite{E18}). RePU has also been proposed to construct differentiable neural networks, and error bounds related to this have been analyzed (\cite{Li20}). According to research, using a RePU network allows for the optimal approximation of smooth functions in Sobolev space with optimal depth and width, and the optimal approximation error bound is also provided in \cite{Li20}. Similarly, another study derived that a shallow RePU network can approximate multivariate functions of a certain function class with finite neural network weight norms and possibly unbounded width (\cite{Abdeljawad22}). Furthermore, it was shown that the derivatives of a network composed of RePU can be expressed as a mixture of various types of RePU, and an improved upper bound on the complexity of these derivatives was derived. This demonstrates the capability of RePU by showing the risk bound for the deep score matching estimator (DSME) (\cite{Shen23}). Despite these results highlighting the good regularity of RePU neural networks, experimental evidence shows that RePU networks suffer from exploding or vanishing values and fail to train properly when the layers are deep.

\subsection{Effective Field Theory of Neural Networks}
\label{subsec1.3}
The language of quantum field theory (QFT) is useful for analyzing systems with a very high degree of freedom. One such example is artificial neural networks. It is known that the distribution of neural network ensembles follows a Gaussian process when the width is infinite, corresponding to the free field in the QFT framework (\cite{Neal96}, \cite{Rasmussen04}, \cite{Roberts22}, \cite{Banta24}, \cite{Halverson21}). In practice, neural networks with finite width exhibit non-Gaussianity in their distributions as layers are stacked, allowing the distribution of neurons to be calculated perturbatively, as in weakly-interacting theory (\cite{Banta24}, \cite{Halverson21}). Interestingly, when the width of the neural network is finite, the propagation of information through the neural network can be understood as a renormalization group (RG) flow. By analyzing the recursive relations of the kernels in RG flow, hyperparameters can be tuned to a criticality state that ensures the stability of the network (\cite{Roberts22}). Additionally, this analysis allows for the classification of activation functions into universality classes based on the behavior of the RG flow (\cite{Roberts22}). Some types of activation functions, according to this classification, cannot satisfy criticality conditions from the outset.

\subsection{Our Contribution}
\label{subsec1.4}
This paper provides a comprehensive analysis of the Rectified Power Unit (RePU) activation function and its limitations through the lens of effective field theory. Based on these insights, we propose the Modified Rectified Power Unit (MRePU) activation function, which addresses the drawbacks of RePU while preserving its advantages. Our contributions are summarized as follows:
\begin{itemize}
    \item \textbf{Theoretical Analysis of RePU's Limitations:}
        We derive the susceptibility of RePU activations and demonstrate that they fail to satisfy criticality conditions, leading to instability in deep neural networks. This analysis identifies why RePU networks experience vanishing or exploding kernels during forward propagation, irrespective of hyperparameter initialization.
    \item \textbf{Proposal of MRePU:} Based on this analysis, we propose the Modified Rectified Power Unit (MRePU), which retains the differentiability and the approximation properties of RePU. As the kernel tends to zero, both susceptibilities of MRePU tend to $C_W$, so that $C_W=1$ is critical in this limit. Unlike for tanh, a small kernel of an MRePU network grows under the recursion at $C_W=1$, because the coefficient $c_p=3p(2p-1)$ of its kernel map is positive; Section~4.3 quantifies this growth.
    \item \textbf{Approximation Properties of MRePU Networks:}
        We theoretically prove that MRePU networks possess universal approximation properties for differentiable functions, akin to RePU networks.
        Furthermore, MRePU networks exhibit efficient polynomial approximation.
    \item \textbf{Trainability and accuracy:} We distinguish whether a network can be trained from how accurate it is once trained. In physics-informed neural networks for four PDEs at three depths, on MNIST with up to 32 layers and on CIFAR-10 with six residual architectures, MRePU networks initialized by our rule trained where RePU networks diverged or stalled. Their accuracy was comparable to that of GELU and SiLU on the Poisson and Helmholtz equations at five and seven hidden layers and lower on the other tasks, and on CIFAR-10 MRePU trained only in architectures that keep the kernel at its inputs small.
    \item \textbf{An initialization rule and its range of validity:} Corollary~9 describes MRePU only in the limit $K\to0$. We expand the kernel map at finite $K$ and show that the rescaled depth $\tau=c_pK_0L$ determines the kernel and the gradients at initialization (Proposition~10). From this we derive Algorithm~1, which sets the input scale and $C_W$ from the depth, the degree and the input range. We state which parts of this are proven and which are empirical: on MNIST, the largest $\tau$ at which training succeeds decreases with depth, and the default $\tau=10^{-2}$ trained networks of every depth we tested.
    \item \textbf{Neural tangent kernel:} The perpendicular susceptibility also governs the neural tangent kernel at initialization, so that the conclusions above carry over to the NTK framework; we confirm this numerically (Section~4.4).
\end{itemize}

\newpage
\section{Preliminary}
\label{sec2}
In this section, we define the neural network and activation functions that will be the subject of our analysis, focusing particularly on the Rectified Power Unit (RePU). Additionally, we explore the effective field theory of neural networks, which will be the primary tool for our analysis.
\subsection{Overview of Neural Networks and Activation Functions}
\label{subsec2.1}
We primarily deal with neural networks having a fully connected network (FCN) architecture in this paper. An FCN is composed of affine transformations with hyperparameters weights and biases at each layer, followed by a nonlinear activation function applied component-wise. Specifically, it is structured as defined below:

\noindent
{\bf Definition 1 (Fully Connected Network (FCN)).} { A Fully Connected Network \( f(x;\theta): \mathbb{R}^{n_{in}} \rightarrow \mathbb{R}^{n_{out}} \) is defined by the following recursive equations:
\begin{equation}
\begin{split}
    z_{i}^{(1)}(x_{\alpha}) &:= \sum_{j=1}^{n_{0}} W_{ij}^{(1)} x_{j;\alpha} + b_{i}^{(1)}, \quad \text{for} \quad i=1,\ldots,n_{1}, \\
    z_{i}^{(l+1)}(x_{\alpha}) &:= \sum_{j=1}^{n_{l}} W_{ij}^{(l+1)} \sigma\Big(z_{j}^{(l)}(x_{\alpha})\Big) + b_{i}^{(l+1)}, \quad \text{for} \quad i=1,\ldots,n_{l+1}; \\
    &\quad l=1,\ldots,L-1.
\end{split} \label{eq:1}
\end{equation}
where \( n_{in} \) is the input dimension of the neural network, \( n_{out} \) is the output dimension of the neural network, and each \( n_{l} \) is the width (i.e., the number of nodes at the \( l \)-th layer). \( L \) is the depth of stacked layers. The parameters $(b_{i}^{(l)})_{i=1,...,n_{l}}$, $(W_{ij}^{(l)})_{i=1,...,n_{l},j=1,...,n_{l-1}}$ are bias vectors and weight matrices, respectively. \( z_{i}^{(l)} \) is called the \textbf{preactivation at the \( l \)-th layer}. \( \sigma: \mathbb{R} \rightarrow \mathbb{R} \) is an \textbf{activation function} that acts on each node of the preactivations.}

In the previous definition of neural networks, we also explained what an activation function is. The connections between the nodes of a neural network can be seen as a type of graph, and this graph representing the connections of the neural network is called the \textbf{architecture}. Besides the architecture, there are several factors that determine the properties of a neural network (training dynamics, inductive bias, approximation properties, etc.). In this paper, we focus on the initialization of the weight and bias parameters and the activation function.

\noindent
{\bf Definition 2 (Initialization distribution of biases and weights).} {In the definition of neural networks, there are adjustable weights and biases parameters. The method of setting these parameters at the beginning of training is called \textbf{initialization}, and typically, each parameter follows a specific probability distribution, referred to as the \textbf{initialization distribution}. Although it is possible to use mathematically complex initialization distributions, it is common practice to assign each weight and bias under the condition of being independent and identically distributed (i.i.d). The probability distributions used can include Gaussian distribution, truncated Gaussian distribution, uniform distribution, and others. For wide neural networks, it is known that when the mean of the weight distribution is 0 and the variance is constant, differences arising from different types of distributions are suppressed by $\frac{1}{width}$. Therefore, for the sake of convenience in our discussion, we will set the distributions of weights and biases to follow a mean-zero Gaussian distribution:
\begin{equation}
    \begin{split}
        \mathbb{E}\Big[b_{i}^{(l)}b_{j}^{(l)}\Big]&=\delta_{ij}C_{b}^{(l)} \\
        \mathbb{E}\Big[W_{i_{1}j_{1}}^{(l)}W_{i_{2}j_{2}}^{(l)}\Big]&=\delta_{i_{1}i_{2}}\delta_{j_{1}j_{2}}\frac{C_{W}^{(l)}}{n_{l-1}}.
    \end{split} \label{eq:initial}
\end{equation}
where the set of bias variances $\{C_{b}^{(1)}, \ldots, C_{b}^{(L)}\}$ and the set of rescaled weight variances $\{C_{W}^{(1)}, \ldots, C_{W}^{(L)}\}$ are called initialization hyperparameters.
}

\noindent
{\bf Activation functions.}{ Since the inception of artificial neural networks, various types of activation functions have been proposed. An activation function is essentially a 1D real scalar-valued function that acts on the preactivation, which is the output of a node. Because the capabilities of a neural network are greatly influenced by the type of activation function used, extensive research has been conducted in this area. We provide a brief summary of the activation functions proposed so far through Table \ref{Activations}.}

\begin{table}[htp]
\centering
\begin{tabular}{l c r}
  Name & Formula & References \\
  Perceptron & $\sigma(z)=\begin{cases}
       1\text{,} &\quad\text{if } z\geq0\\
       0\text{,} &\quad\text{if }z<0 \\
     \end{cases}$ & \cite{rosenblatt58} \\ 
  Sigmoid & $\sigma(z)=\frac{1}{1+e^{-z}}$ & \citet{cybenko89} \\
  Tanh & $\sigma(z)=\tanh(z)=\frac{e^{z}-e^{-z}}{e^{z}+e^{-z}}$ & \citet{cybenko89}, \cite{LeCun02} \\
  Sin & $ \sigma(z)=\sin(z)$ & \cite{Parascandolo16}, \cite{Sitzmann20} \\
  ReLU & $\sigma(z)=\begin{cases}
       z\text{,} &\quad\text{if } z\geq0\\
       0\text{,} &\quad\text{if }z<0 \\
     \end{cases}$ & \cite{Nair10}, \cite{Glorot11} \\
  Leaky ReLU & $\sigma(z)=\begin{cases}
       z\text{,} &\quad\text{if } z\geq0\\
       \alpha z\text{,} &\quad\text{if }z<0 \\
     \end{cases}$ & \cite{He15}, \cite{Maas13}\\
  Softplus & $\sigma(z)=\log(1+e^{z})$ & \cite{Bishop06} \\
  SWISH & $\sigma(z)=\frac{z}{1+e^{-z}}$ & \cite{Ramachandran17} \\
  GELU & $\sigma(z)=\Big[\frac{1}{2}+\frac{1}{2}\text{erf}\Big(\frac{z}{\sqrt{2}}\Big)\Big]z$ & \cite{Hendrycks16} \\
\end{tabular}
\caption{Various Kinds of Activation functions}\label{Activations}
\end{table}

Now we define the RePU activation functions, which are the main objective of our work.

\noindent
{\bf Definition 3 (Rectified Power Unit (RePU)).}{ The RePU activation function is defined by the following equation:

\begin{equation*}
    \begin{split}
        \sigma(z)=\begin{cases}
       z^{p}\text{,} &\quad\text{if } z\geq0\\
       0\text{,} &\quad\text{if }z<0 \\
     \end{cases}.
    \end{split}
\end{equation*}
where $p$ is a positive integer that determines the power to which the input $z$ is raised. When $p=1$, it corresponds to ReLU, and thus RePU can be considered a generalized version of ReLU. The cases where $p=2$ and $p=3$ are referred to as Rectified Quadratic Unit (ReQU) and Rectified Cubic Unit (ReCU), respectively. For RePU with 
$p=k$, it is easy to see that it is differentiable $k-1$ times, and this property ensures that the neural network and its gradients are differentiable functions.
}

\subsection{Effective Field Theory for Neural Networks}
\label{subsec2.2}

We utilize the framework of effective field theory to understand the distribution of preactivations in a neural network ensemble sampled from the initialization distribution. To accomplish this, we first define the necessary concepts.

\noindent
{\bf Definition 4 ($M$-point correlators).}{ The moments or $M$-point correlators of a probability distribution  $p(z)=p(z_{1},...,z_{N})$  are defined as follows:
\begin{equation*}
    \mathbb{E}[z_{\mu_{1}}\cdots z_{\mu_{M}}]=\int z_{\mu_{1}}\cdots z_{\mu_{M}} p(z)d\nu(z).
\end{equation*}

The collection of all $M$-point correlators fully characterizes the probability distribution since the information from the M-point correlators allows us to compute the expected values of analytic observables.} 
For a Gaussian distribution, it is important to note that the distribution’s information can be fully specified by the 1-point correlator (mean) and the 2-point correlator (variance). Keeping this in mind, we slightly modify the $M$-point correlator to define what is known as the connected correlator (also known as cumulant). The definition is as follows:

\noindent
{\bf Definition 5 ($M$-point connected correlators).}{ For $2\leq M$, the cumulant or $M$-point connected correlators of a probability distribution  $p(z)$  are defined as follows:
\begin{equation*}
    \begin{split}
    &\mathbb{E}[z_{\mu_{1}}\cdots z_{\mu_{M}}]|_{\text{connected}}=\mathbb{E}[z_{\mu_{1}}\cdots z_{\mu_{M}}]- \\&\sum_{\text{all subdivisions of }(\mu_{1},...,\mu_{M})}\mathbb{E}[z_{\mu_{i_{1,1}}}\cdots z_{\mu_{i_{1,j_{1}}}}]|_{\text{connected}}\cdots\mathbb{E}[z_{\mu_{i_{k,1}}}\cdots z_{\mu_{i_{k,j_{k}}}}]|_{\text{connected}}.
    \end{split}
\end{equation*}

\noindent
{\bf Proposition 6 (Wick contraction).}{ Suppose $(X_{1},\dots,X_{n})$ is a zero-mean multivariate normal random vector. Then, all odd-order correlators vanish, and the even-order correlators are given by
\[
\mathbb{E}[X_{j_{1}}^{i_{1}}\dots X_{j_{n}}^{i_{n}}]=\sum_{\text{(all possible parings)}}\prod_{\text{(parings)}}\mathbb{E}[X_{p_{1}}X_{p_{2}}]
\]
}

For $M=1$, the connected correlator coincides with the regular correlator. For parity-symmetric distributions, it also coincides for $M=2$. Using Wick contraction, it is known that for a Gaussian distribution, the connected correlators for  $M > 2$  are zero. Therefore, higher-point connected correlators can be used to determine how much the distribution deviates from a Gaussian distribution. Additionally, if the connected correlators for  $M > 2$  are small, the distribution is defined as \textbf{nearly-Gaussian}. In quantum field theory, the 2-point connected correlator can be seen as an expression of the translational symmetry of a free field. In a free field, there is no interaction due to potential, so all connected correlators higher than the 2-point vanish. Conversely, if there are interactions between particles due to a potential, higher-order connected correlators become non-zero. Similarly, in the neural network we analyze, if the higher-order correlators do not vanish and instead show significant values, this implies meaningful interactions between nodes in the feature space, suggesting that feature learning is taking place.
}

Using these concepts, we can analyze the probability distribution of the preactivations of an ensemble following the initialization distribution. Specifically, since each layer depends on the previous layer, marginalization reveals that there are recursive relations between the statistics of the probability distributions of the layers.

\noindent
{\bf Induced distributions.}{
Following the notation in \eqref{eq:1}, let  $p(z^{(l)}|\mathcal{D})$ denote the probability distribution of the preactivations in the  $l$-th layer given a dataset $\mathcal{D} = \{x_{i;\alpha} \}_{i=1,...,n_{0};\alpha =1,...,N_{\mathcal{D}}}$. Here, $\alpha$ denotes the label index for the samples. For the first layer, the preactivation has  $M > 2$  connected correlators equal to 0, and has the following mean and covariance, indicating that each preactivation follows an independent Gaussian distribution.
\begin{equation*}
    \begin{split}
        \mathbb{E}[z_{i;\alpha}^{(1)}]&=\mathbb{E}\bigg[b_{i}^{(1)}+\sum_{j=1}^{n_{0}}W_{ij}^{(1)}x_{j;\alpha}\bigg]=0,\\
        \mathbb{E}[z_{i_{1};\alpha_{1}}^{(1)}z_{i_{2};\alpha_{2}}^{(1)}]&=\mathbb{E}\bigg[\bigg(b_{i_{1}}^{(1)}+\sum_{j=1}^{n_{0}}W_{i_{1}j_{1}}^{(1)}x_{j_{1};\alpha_{1}}\bigg)\bigg(b_{i_{2}}^{(1)}+\sum_{j=1}^{n_{0}}W_{i_{2}j_{2}}^{(1)}x_{j_{2};\alpha_{2}}\bigg)\bigg] \\
        &=\delta_{i_{1}i_{2}}\bigg(C_{b}^{(1)}+C_{W}^{(1)}\frac{1}{n_{0}}\sum_{j=1}^{n_{0}}x_{j;\alpha_{1}}x_{j;\alpha_{2}}\bigg)=\delta_{i_{1}i_{2}}G_{\alpha_{1}\alpha_{2}}^{(1)}.
    \end{split}
\end{equation*}
Where $G_{\alpha_{1}\alpha_{2}}^{(1)}$ is called as the \textbf{metric}, which represents the two-point correlator of the preactivations for different samples. Similarly, the conditional probability  $p(z^{(l+1)}|z^{(l)})$ is also described by Gaussian distribution with covariance $\hat{G}_{\alpha_{1}\alpha_{2}}^{(l+1)}$, where the distribution (i.e.,~$\hat{G}_{\alpha_{1}\alpha_{2}}^{(l+1)}$) depends on the condition $z^{(l)}$. We refer to this $\hat{G}_{\alpha_{1}\alpha_{2}}^{(l+1)}$ as the  \textbf{$l+1$-th layer stochastic metric}. We can consider the mean metric $G_{\alpha_{1}\alpha_{2}}^{(l+1)}$, which is the average of the stochastic metric. Additionally, we can consider the variance of $\Delta \hat{G}_{\alpha_{1}\alpha_{2}}^{(l+1)} = \hat{G}_{\alpha_{1}\alpha_{2}}^{(l+1)} - G_{\alpha_{1}\alpha_{2}}^{(l+1)}$, which is the fluctuation of the stochastic metric around the mean metric. We refer to scaled quantity of this variance as the \textbf{four-point vertex}, defined as follows:
\begin{equation*}
    V^{(l+1)}_{(\alpha_{1}\alpha_{2})(\alpha_{3}\alpha_{4})}:=n_{l}\mathbb{E}\bigg[\Delta \hat{G}_{\alpha_{1}\alpha_{2}}^{(l+1)}\Delta \hat{G}_{\alpha_{3}\alpha_{4}}^{(l+1)}\bigg]
    =n_{l}\bigg(\mathbb{E}\bigg[\hat{G}_{\alpha_{1}\alpha_{2}}^{(l+1)}\hat{G}_{\alpha_{3}\alpha_{4}}^{(l+1)}\bigg]-G_{\alpha_{1}\alpha_{2}}^{(l+1)}G_{\alpha_{3}\alpha_{4}}^{(l+1)}\bigg).
\end{equation*}
The four-point vertex is related to the 4-point connected correlator as follows:
\begin{equation}    
    \begin{split}
        &\mathbb{E}[z_{i_{1};\alpha_{1}}^{(l+1)}z_{i_{2};\alpha_{2}}^{(l+1)}z_{i_{3};\alpha_{3}}^{(l+1)}z_{i_{4};\alpha_{4}}^{(l+1)}]|_{\text{connected}}\\
        &=\frac{1}{n_{l}}[\delta_{i_{1}i_{2}}\delta_{i_{3}i_{4}}V^{(l+1)}_{(\alpha_{1}\alpha_{2})(\alpha_{3}\alpha_{4})}+\delta_{i_{1}i_{3}}\delta_{i_{2}i_{4}}V^{(l+1)}_{(\alpha_{1}\alpha_{3})(\alpha_{2}\alpha_{4})}+\delta_{i_{1}i_{4}}\delta_{i_{2}i_{3}}V^{(l+1)}_{(\alpha_{1}\alpha_{4})(\alpha_{2}\alpha_{3})}].
    \end{split} \label{fourvert}
\end{equation}
The quantities we have just defined contain significant information about the distribution of the preactivation $p(z^{(l)}|\mathcal{D})$. If we define the action $S$ of $p(z^{(l)}|\mathcal{D})$ as follows (parity symmetry can be easily checked):
\begin{equation*}
\begin{split}
        &S(z^{(l)}):=\frac{1}{2}\sum_{i=1}^{n_{l}}\sum_{\alpha_{1}\alpha_{2}\in\mathcal{D}}g^{\alpha_{1}\alpha_{2}}_{(l)}z_{i;\alpha_{1}}^{(l)}z_{i;\alpha_{2}}^{(l)} \\&-\frac{1}{8}\sum_{i_{1},i_{2}=1}^{n_{l}}\sum_{\alpha_{1},...,\alpha_{4}\in\mathcal{D}}v^{(\alpha_{1}\alpha_{2})(\alpha_{3}\alpha_{4})}_{(l)}z_{i;\alpha_{1}}^{(l)}z_{i;\alpha_{2}}^{(l)}z_{i;\alpha_{3}}^{(l)}z_{i;\alpha_{4}}^{(l)} +O((z^{(l)})^{6}).
\end{split}
\end{equation*}
It is known that the following relationship holds for the coefficients. This can be derived by comparing the coefficients of the connected correlator given by the mean metric and four-point vertex with those of the connected correlator assumed by the ansatz provided by action $S$. $g^{\alpha_{1}\alpha_{2}}_{(l+1)}$ and $v^{(\alpha_{1}\alpha_{2})(\alpha_{3}\alpha_{4})}_{(l+1)}$ (\cite{Roberts22}):
\begin{equation*}
\begin{split}
    g^{\alpha_{1}\alpha_{2}}_{(l)}&=G^{\alpha_{1}\alpha_{2}}_{(l)}+O\Big(\frac{1}{n_{l-1}}\Big),\\
    v^{(\alpha_{1}\alpha_{2})(\alpha_{3}\alpha_{4})}_{(l)}&=\frac{1}{n_{l}}V^{(\alpha_{1}\alpha_{2})(\alpha_{3}\alpha_{4})}_{(l)}+O\Big(\frac{1}{n_{l-1}^{2}}\Big).
\end{split}
\end{equation*}
Where, $G^{\alpha_{1}\alpha_{2}}_{(l)}$ is inverse of mean metric $G_{\alpha_{1}\alpha_{2}}^{(l)}$ and $V^{(\alpha_{1}\alpha_{2})(\alpha_{3}\alpha_{4})}_{(l)}$ is defined as follows:
\begin{equation*}
    V^{(\alpha_{1}\alpha_{2})(\alpha_{3}\alpha_{4})}_{(l)}:=\sum_{\beta_{1},...,\beta_{4}\in\mathcal{D}}G^{\alpha_{1}\beta_{1}}_{(l)}G^{\alpha_{2}\beta_{2}}_{(l)}G^{\alpha_{3}\beta_{3}}_{(l)}G^{\alpha_{4}\beta_{4}}_{(l)}V^{(l)}_{(\alpha_{1}\alpha_{2})(\alpha_{3}\alpha_{4})}.
\end{equation*}
}

\noindent
{\bf Kernels and Representation Group flow.}{ Now, we will investigate how the statistics of the preactivation change as they pass through the layers. We will refer to the evolution of the distribution through the layers as the \textbf{representation group flow}, following the terminology of (\cite{Roberts22}). While it is not feasible to explicitly calculate the statistics for a complex non-Gaussian distribution, it is possible to develop a perturbative theory to understand the recursive flow of the key statistics of the distribution. Before delving into this, let’s define the notation for Gaussian expectations. For an observable $\mathcal{O}(z)$  with a mean-zero Gaussian distribution with variance $\bf{K} = (K_{ij})$, the expectation is denoted as $\langle \mathcal{O}(z) \rangle_{\bf{K}}$. Now, if we define the \textbf{kernel} of the neural network through the following recursive relation:
\begin{equation}
\begin{split}
        K^{(1)}_{\alpha_{1}\alpha_{2}}&:=G^{(1)}_{\alpha_{1}\alpha_{2}}, \\
        K^{(l+1)}_{\alpha_{1}\alpha_{2}}&=C^{(l+1)}_{b}+C^{(l+1)}_{W}\langle \sigma(z_{\alpha_{1}})\sigma(z_{\alpha_{2}})\rangle_{K^{(l)}}.
\end{split} \label{kernel recur}
\end{equation}
it is known that this kernel satisfies following properties, as discussed in (\cite{Roberts22}).
\begin{equation}
    \begin{split}
        G^{(l)}_{\alpha_{1}\alpha_{2}}&=K^{(l)}_{\alpha_{1}\alpha_{2}}+O\Big(\frac{1}{n_{l-1}}\Big), \\
        V^{(l+1)}_{(\alpha_{1}\alpha_{2})(\alpha_{3}\alpha_{4})}&=\Big(C_{W}^{(l+1)}\Big)^{2}\Big[\langle \sigma(z_{\alpha_{1}})\sigma(z_{\alpha_{2}})\sigma(z_{\alpha_{3}})\sigma(z_{\alpha_{4}})\rangle_{K^{(l)}}\\
        &-\langle \sigma(z_{\alpha_{1}})\sigma(z_{\alpha_{2}})\rangle_{K^{(l)}}\langle \sigma(z_{\alpha_{3}})\sigma(z_{\alpha_{4}})\rangle_{K^{(l)}}\Big] \\
        &+\frac{(C_W^{(l+1)})^{2}}{4}\frac{n_{l}}{n_{l-1}}\sum_{\beta_{1},...,\beta_{4}\in\mathcal{D}}V^{(\beta_{1}\beta_{2})(\beta_{3}\beta_{4})}_{(l)}\Big\langle \sigma(z_{\alpha_{1}})\sigma(z_{\alpha_{2}}) \Big(z_{\beta_{1}}z_{\beta_{2}}-K_{\beta_{1}\beta_{2}}^{(l)}\Big)\Big\rangle_{K^{(l)}} \\ 
        &\times\Big\langle \sigma(z_{\alpha_{3}})\sigma(z_{\alpha_{4}})\Big(z_{\beta_{3}}z_{\beta_{4}}-K_{\beta_{3}\beta_{4}}^{(l)}\Big)\Big\rangle_{K^{(l)}} + O\Big(\frac{1}{n_{l}}\Big).
    \end{split} \label{vertexrecur}
\end{equation}

}
\noindent
{\bf Susceptibility.}{ When expanding the kernel matrix for two inputs  $(x_{i;+})$ and $(x_{i;-})$, it can be written as follows:
\begin{equation*}
    K^{(l)}=\begin{pmatrix}
K^{l}_{++} & K^{l}_{+-}\\
K^{l}_{-+} & K^{l}_{--}
\end{pmatrix}=K^{(l)}_{[0]}\begin{pmatrix}
1 & 1\\
1 & 1
\end{pmatrix}+K^{(l)}_{[1]}\begin{pmatrix}
1 & 0\\
0 & -1
\end{pmatrix}+K^{(l)}_{[2]}\begin{pmatrix}
1 & -1\\
-1 & 1
\end{pmatrix}.
\end{equation*}
Where, $K^{(l)}_{\pm\pm}:=\mathbb{E}\bigg[\frac{1}{n_{l}}\sum_{i=1}^{n_{l}}(z_{i;\pm}^{(l)})^{2}\bigg]$, $K^{(l)}_{\pm\mp}:=\mathbb{E}\bigg[\frac{1}{n_{l}}\sum_{i=1}^{n_{l}}z_{i;\pm}^{(l)}z_{i;\mp}^{(l)}\bigg]$.
The coefficients $K^{(l)}_{[\alpha]}$ are given by:
\begin{equation*}
    \begin{split}
        K_{[0]}^{(l)}&=\frac{1}{4}[K^{(l)}_{++}+K^{(l)}_{--}+2K^{(l)}_{+-}], \\
        K_{[1]}^{(l)}&=\frac{1}{2}[K^{(l)}_{++}-K^{(l)}_{--}], \\
        K_{[2]}^{(l)}&=\frac{1}{4}[K^{(l)}_{++}+K^{(l)}_{--}-2K^{(l)}_{+-}].
    \end{split}
\end{equation*}

If we express $(x_{i;+})$ and $(x_{i;-})$ as $x_{i;\pm} = \frac{x_{i;+}+x_{i;-}}{2}\pm \delta x_{i}$, it can be seen that as $(\delta x_{i})$ approaches zero, the coefficients $K^{(l)}_{[1]}, K^{(l)}_{[2]}$  also approach zero. Consequently, the coefficients can be expanded as follows:
\begin{equation*}
    \begin{split}
        K_{[0]}^{(l)}&=K_{00}^{(l)}+\delta^{2}K_{[0]}^{(l)}+O(\delta^{4}), \\
        K_{[1]}^{(l)}&=\delta K_{[1]}^{(l)}+\delta^{3}K_{[1]}^{(l)}+O(\delta^{5}), \\
        K_{[2]}^{(l)}&=\delta^{2}K_{[2]}^{(l)}+\delta^{4}K_{[2]}^{(l)}+O(\delta^{6}).
    \end{split}
\end{equation*}
Then, it is known that the following recursion can be derived (\cite{Roberts22}):
\begin{equation*}
    \begin{split}
        K_{00}^{(l+1)}&=C_{b}+C_{W}g(K_{00}^{(l)}), \\
        \delta K_{[1]}^{(l+1)}&=\chi_{\parallel}(K_{00}^{(l)})\delta K_{[1]}^{(l)}, \\
        \delta^{2} K_{[2]}^{(l+1)}&=\chi_{\perp}(K_{00}^{(l)})\delta^{2}K^{(l)}_{[2]}+h(K^{(l)}_{00})(\delta K^{(l)}_{[1]})^{2}.
    \end{split}
\end{equation*}
Where $K_{00}^{(l+1)}$ is the kernel for a single input, and each $g$, $h$, $\chi_{\parallel}$, $\chi_{\perp}$ are defined as follows:       
\begin{equation}
    \begin{split}
        g(K)&:=\langle\sigma(z)\sigma(z)\rangle_{K}, \\
        \chi_{\parallel}(K)&:=C_{W}g'(K)=\frac{C_{W}}{K}\langle z \sigma'(z)\sigma(z)\rangle_{K},\\
        \chi_{\perp}(K)&:=C_{W}\langle\sigma'(z)\sigma'(z)\rangle_{K}, \\
        h(K)&:=\frac{C_{W}}{4K^{2}}\Big\langle\sigma'(z)\sigma'(z)\Big(z^{2}-K\Big)\Big\rangle_{K}=\frac{1}{2}\frac{d}{dK}\chi_{\perp}(K).
    \end{split} \label{eq:suscep}
\end{equation}
 $K_{[1]}^{(l)}$ and $K_{[2]}^{(l)}$ represent the difference in the squared distances between two inputs and a value related to the square of the difference between the two inputs, respectively. As the inputs become closer, their leading terms exhibit a geometric ratio in susceptibility as they pass through layers, which forms the basis for naming $\chi_{\parallel}$ and  $\chi_{\perp}$ as the parallel and perpendicular susceptibility, respectively. Since each leading term follows this geometric ratio in susceptibility, if the value is less than 1, the kernel vanishes along the layer, and if it is greater than 1, it diverges. Thus, setting each susceptibility to 1 is essential for maintaining a stable kernel.

\newpage
\noindent
{\bf Summary of Framework.}{
In this section, we investigate the evolution of preactivation statistics across layers, referred to as the Representation Group flow. By analyzing the recursive kernel relation (Eq.~\eqref{kernel recur}), we derive a fixed point for the statistical flow. When the initial ensemble of the network operates near this fixed point, the propagation of small perturbations through the layers is governed by the parallel ($\chi_{\parallel}$) and perpendicular ($\chi_{\perp}$) susceptibilities. These quantities can be interpreted as the eigenvalues of the linearized flow along the respective directions around the fixed point. Consequently, for both training dynamics and inference, it is crucial that these eigenvalues are close to unity ($\chi \approx 1$). This criticality condition ensures that perturbations are neither amplified nor suppressed but are preserved across layers, thereby maintaining a stable signal propagation.
}
}

\section{Failure of RePU Activation}
\subsection{Susceptibility Calculation}
Activation functions that exhibit similar behaviors according to the RG flow are classified into \textbf{universality classes}. Universality class is a concept in statistical physics used to classify systems that exhibit similar behavior at the limit scale. According to (\cite{Roberts22}), scale-invariant activations (e.g. ReLU, LeakyReLU) form a single universality class called the self-invariant universality class. Tanh and Sin fall into the $K^{\star}=0$ universality class, while SWISH and GELU belong to the half-stable universality class. On the other hand, Perceptron, Sigmoid, and Softplus are known to not satisfy criticality at all. We will now calculate the susceptibility for the RePU activation function following Eq.~\eqref{eq:suscep} and demonstrate that RePU also fails to satisfy criticality.
For the RePU activation function of order $p$, the parallel susceptibility and perpendicular susceptibility are calculated as follows:

\begin{equation*}
    \begin{split}
        \chi_{\parallel}(K)&=\frac{C_{W}}{K}\langle z \sigma'(z)\sigma(z)\rangle_{K}=\frac{C_{W}}{K}\langle z\lfloor pz^{p-1}\rfloor_{+}\lfloor z^{p}\rfloor_{+}\rangle_{K} \\
        &=\frac{C_{W}p}{K\sqrt{2\pi K}}\int_{0}^{\infty}e^{-\frac{z^{2}}{2K}}z^{2p}dz=\frac{C_{W}p(2p-1)!!K^{p-1}}{2}.
    \end{split}
\end{equation*}

\begin{equation}
    \begin{split}
        \chi_{\perp}(K)&=C_{W}\langle \sigma'(z)\sigma'(z)\rangle_{K}=C_{W}\langle p^{2}\lfloor z^{p-1}\rfloor_{+}\lfloor z^{p-1} \rfloor_{+}\rangle_{K} \\
        &=\frac{C_{W}p^{2}}{\sqrt{2\pi K}}\int_{0}^{\infty}e^{-\frac{z^{2}}{2K}}z^{2p-2}dz=\frac{C_{W}p^{2}(2p-3)!!K^{p-1}}{2}.
    \end{split}\label{eq:REPUsus}
\end{equation}

Here the notation $n!!$ means double factorial (e.g. $7!!=7\cdot5\cdot3\cdot1$). The parallel susceptibility to perpendicular susceptibility ratio is $2p-1 : p$. Therefore, except for $p=1$ (i.e., ReLU), it is impossible to set both susceptibilities to 1 simultaneously. If we set the parallel susceptibility to 1, the perpendicular susceptibility will be less than 1, causing the distance between two inputs to vanish as they propagate through the layers. Conversely, if we set the perpendicular susceptibility to 1, the parallel susceptibility will be greater than 1, causing each input to diverge as it propagates through the layers.
\subsection{Experimental Validation}
\begin{figure}[htp]
\centering
\minipage{0.49\textwidth}
  \includegraphics[scale=0.40]{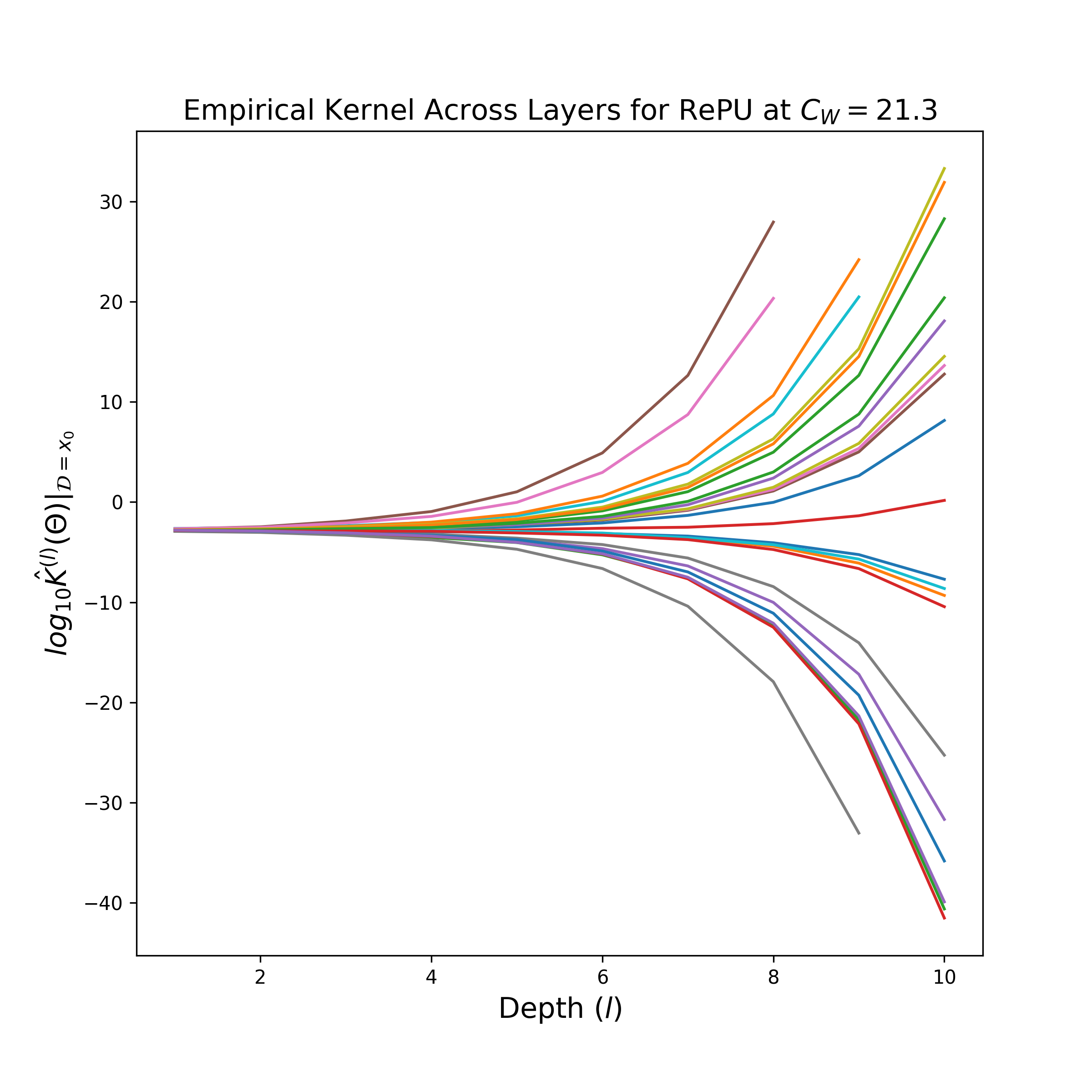}
\endminipage\hfill
\minipage{0.49\textwidth}
  \includegraphics[scale=0.40]{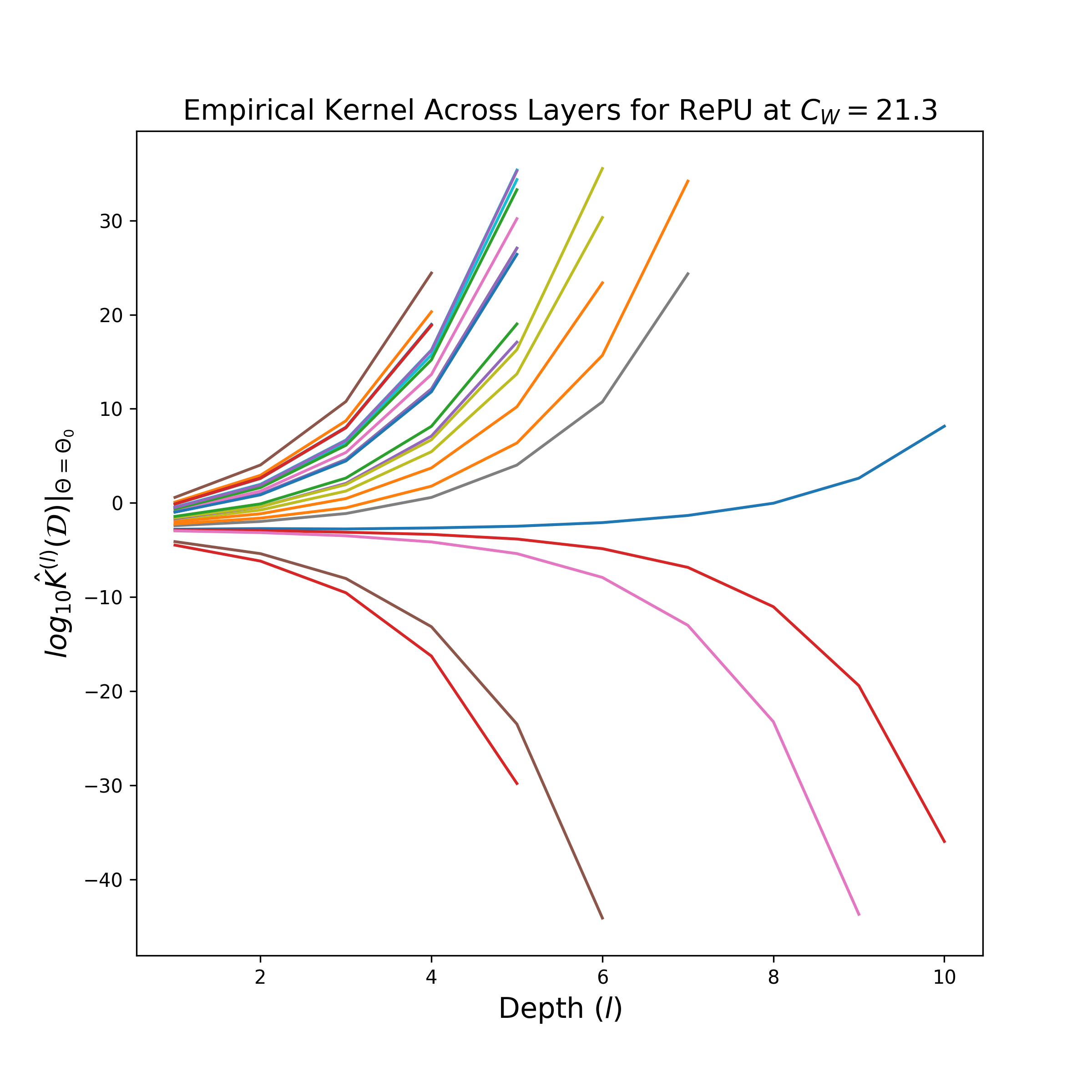}
\endminipage
\caption{Empirical Kernels at Initialization Across Layers for RePU Activation with $p=2$. \textbf{Left}: Data is fixed and randomness is in the weight parameters. \textbf{Right}: Weight parameters are fixed and data is random. Each line represents a sample.}\label{fig1}
\end{figure}

\noindent
{\bf Empirical Kernel Behavior at Initialization.}{
According to the results in section 3.1, the kernel value of Neural Networks with RePU (for $p>1$) for a single input must either explode or vanish as it propagates through the layers under the initialization distribution. In this section, we experimentally verify this. For the experiment, the neural network is configured with an input dimension of 2, an output dimension of 1, hidden layers with widths of 512, and a depth of 10 (therefore, the total number of layers, including the input and output layers, is 12). The activation function is RePU. The initialization distribution follows Eq.~\eqref{eq:initial}, with $C_{b}$ set to 0, and $C_{W}$  adjusted during the experiment. To experimentally observe the change in the kernel, we define the following empirical kernel quantities:
\begin{equation*}
    \begin{split}
        \hat{K}^{(l)}(\Theta)|_{\mathcal{D}=x_{0}}:=\frac{1}{n_{l}}\sum_{i}^{n_{l}}\Big(z^{(l)}_{i}(\Theta)|_{\mathcal{D}=x_{0}}\Big)^{2} \\
        \hat{K}^{(l)}(\mathcal{D})|_{\Theta=\Theta_{0}}:=\frac{1}{n_{l}}\sum_{i}^{n_{l}}\Big(z^{(l)}_{i}(\mathcal{D})|_{\Theta=\Theta_{0}}\Big)^{2}.
    \end{split}
\end{equation*}
Where \( z^{(l)}_{i}(\Theta)|_{\mathcal{D}=x_{0}} \) is a random variable conditioned on the input \( x_{0} \) and has randomness on \( \Theta \). Similarly, \( z^{(l)}_{i}(\mathcal{D})|_{\Theta=\Theta_{0}} \) is a random variable conditioned on the parameter \( \Theta \) and has randomness on the data input. Each component of sample in $\mathcal{D}$ follows $N(0,1)$. $x_{0}$ is also generated from $\mathcal{D}$. The experimental results for a width of 512, $p=2$, and $C_{W}=21.3$ are shown in Figs.~\ref{fig1}. Away from its fixed point $K^\star$, the kernel explodes or vanishes double-exponentially: Eq.~(4) with $g(K)=\frac{(2p-1)!!}{2}K^p$ gives $\log\bigl(K^{(l)}/K^\star\bigr)=p^{\,l-1}\log\bigl(K^{(1)}/K^\star\bigr)$. The results in the figure for $C_{W}=21.3$ show that the kernel is highly sensitive to initial conditions and tends to either double exponentially explode or vanish. According to Eqs.~\eqref{eq:suscep} and~\eqref{eq:REPUsus}, the susceptibility itself acts as a geometric ratio and is proportional to the kernel. As this ratio increases exponentially, it causes the kernel to grow double exponentially.

}

\begin{figure}[htp]
\centering
\includegraphics[scale=0.6]{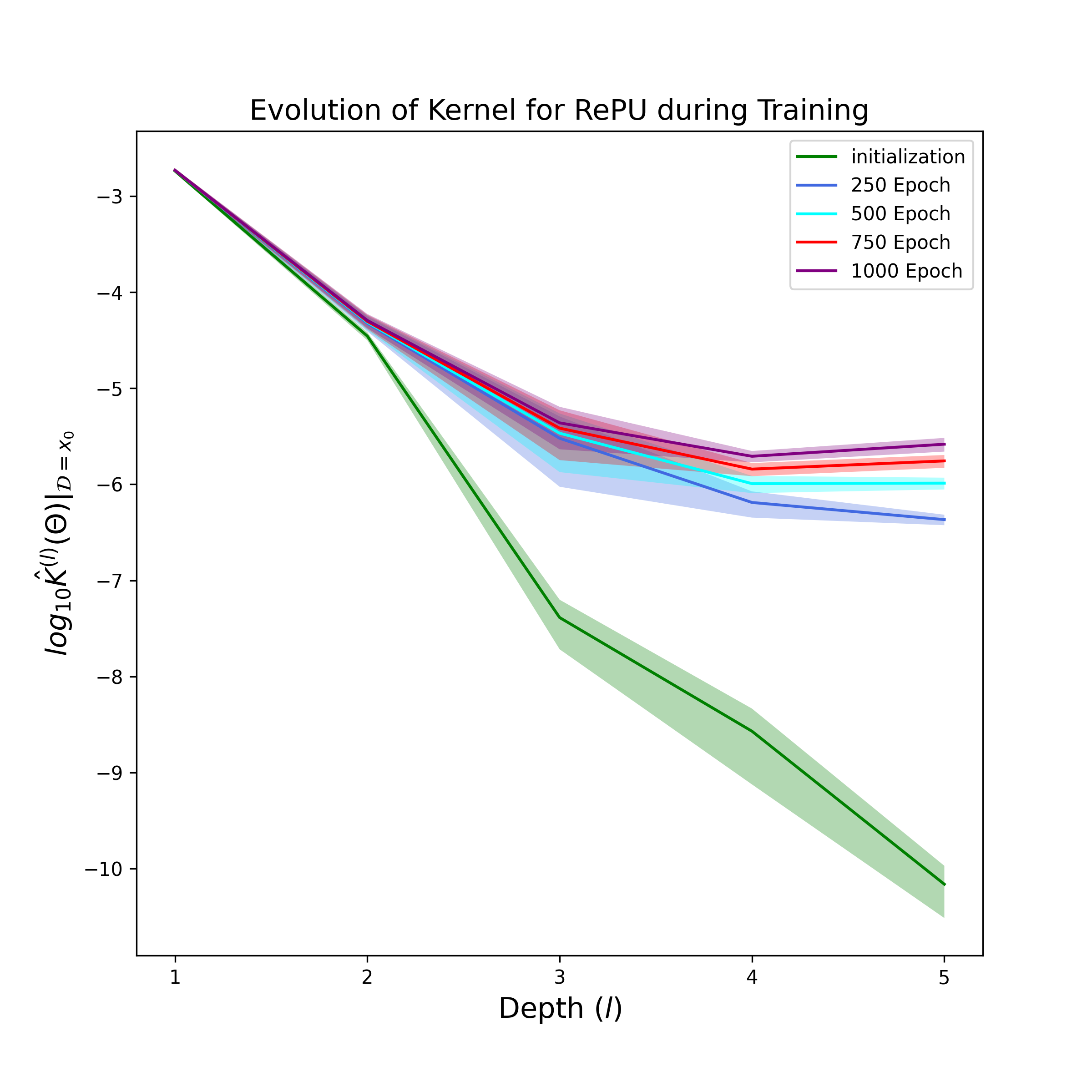}
\caption{The evolution of the mean of empirical kernels over an ensemble of 100 models for $x_{0} = (1,0)$ as training progresses for the RePU activation with $p=2$. The shaded areas represent the region between $\log_{10}(\text{mean}\pm 0.1 \times \text{standard deviation})$.}\label{Evolution of Kernel}
\end{figure}

\begin{figure}[htp]

\centering
\minipage{0.49\textwidth}
\includegraphics[scale=0.40]{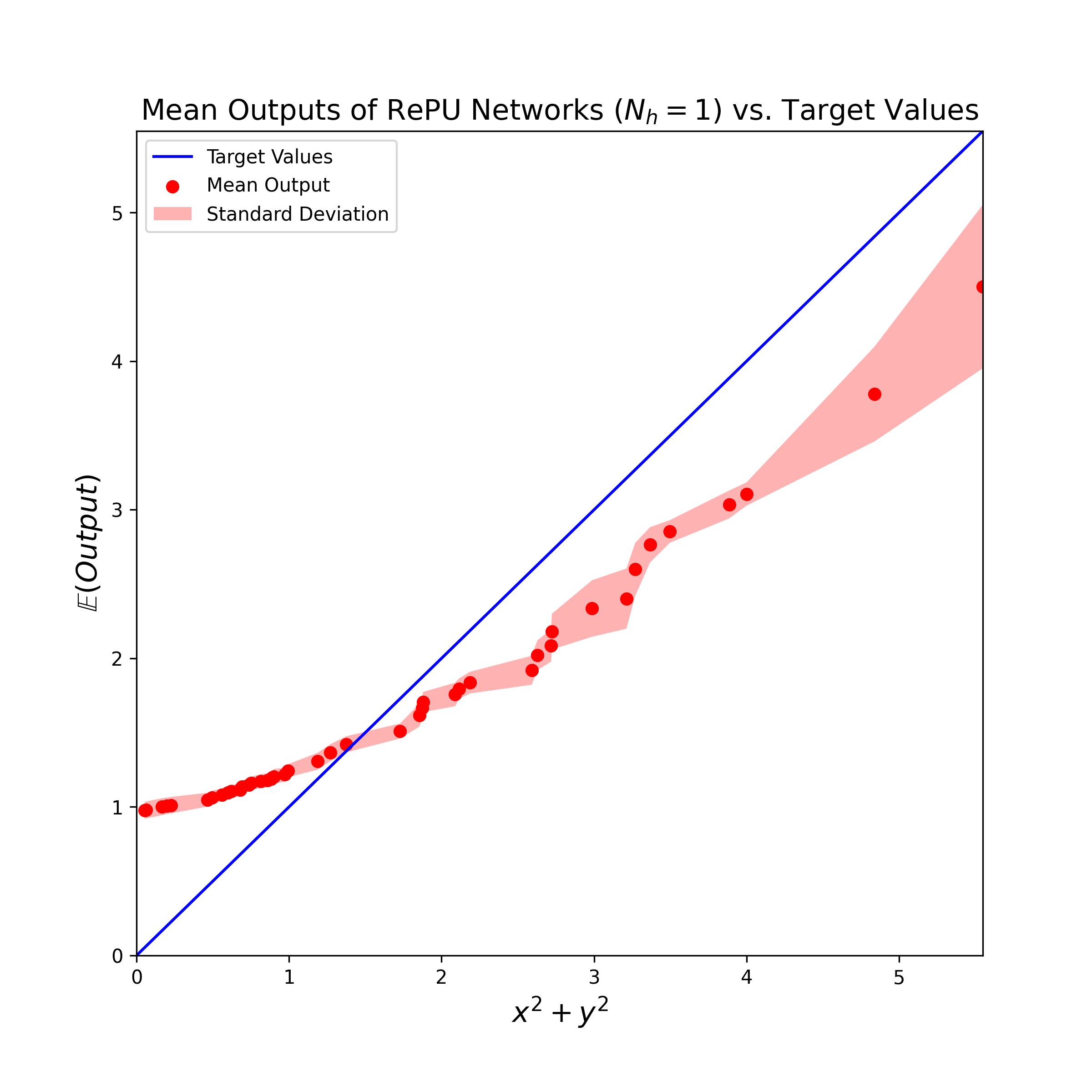}
\endminipage\hfill
\minipage{0.49\textwidth}
\includegraphics[scale=0.40]{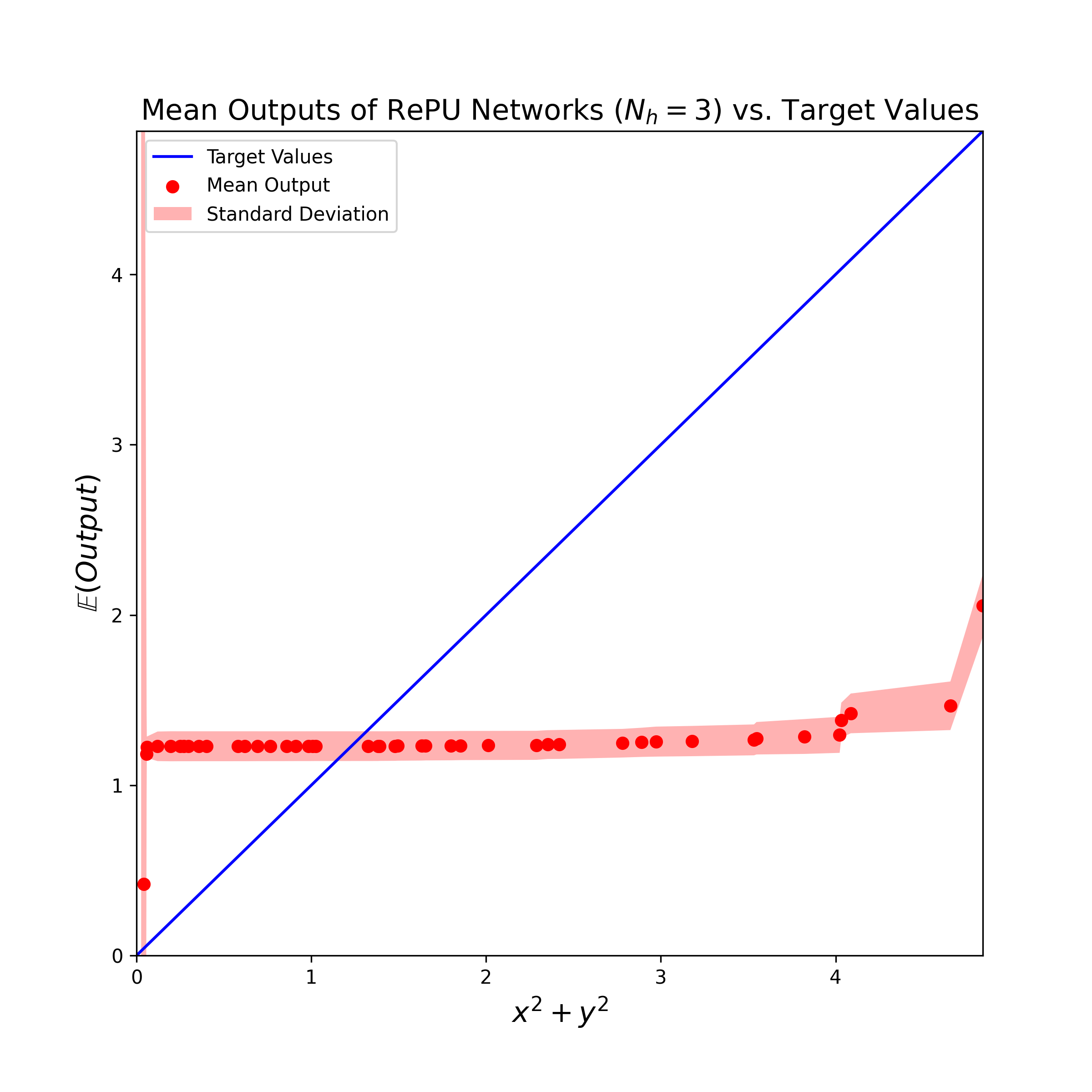}
\endminipage\par\medskip
\centering
\includegraphics[scale=0.40]{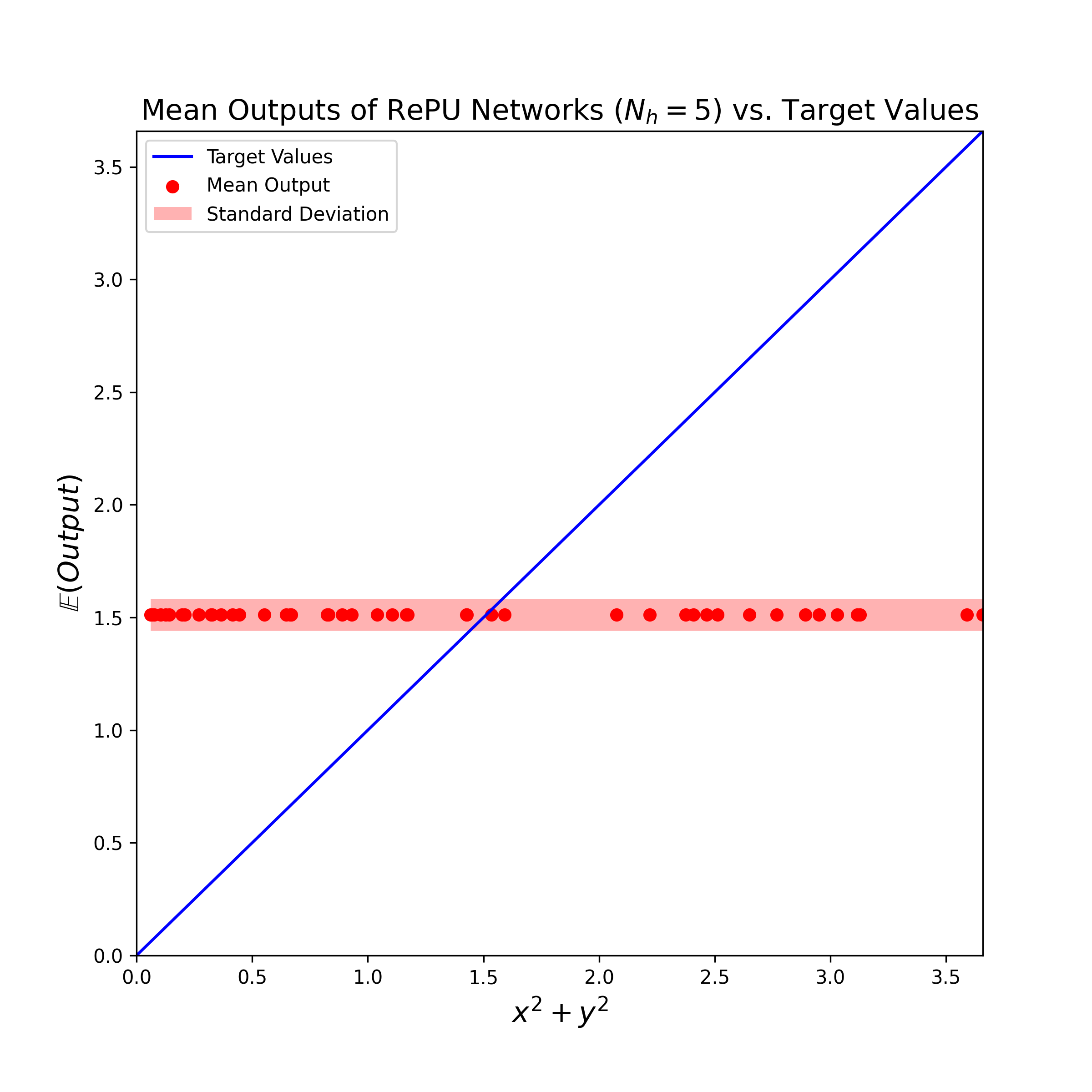}

\caption{Mean outputs over an ensemble of 100 models versus target values on random test data. The shaded areas represent the region of 1 standard deviation. \textbf{Top left}: hidden layers ($N_{h}$)= 1, \textbf{Top right}: $N_{h}$ = 3, \textbf{Bottom}: $N_{h}$ = 5.}
\label{repumeanoutput}
\end{figure}

\noindent
{\bf Dynamics of Mean Outputs and Kernels during Training.}{ In this section, we experimentally observe the dynamics of kernels in RePU neural networks during training. In the backpropagation algorithm used for updating network parameters, each factor in the chain rule is proportional to the perpendicular susceptibility at each layer, as described by the following equation.
\begin{equation*}
\begin{split}
    \mathbb{E}\bigg[\sum_{j_{1},j_{2}} \frac{dz_{i}^{(l+1)}}{dz_{j_{1}}^{(l)}}\frac{dz_{i}^{(l+1)}}{dz_{j_{2}}^{(l)}}\bigg]&=\mathbb{E}\bigg[\sum_{j_{1},j_{2}}W_{ij_{1}}^{(l+1)}W_{ij_{2}}^{(l+1)}{\sigma'}_{j_{1}}^{(l)}{\sigma'}_{j_{2}}^{(l)}\bigg] \\
    &=C_{W}^{(l+1)}\langle\sigma'(z)\sigma'(z)\rangle_{K^{(l)}}=\chi_{\perp}\big(K^{(l)}\big)
\end{split}
\end{equation*}
Through this, we can infer that in RePU neural networks, the kernel either increases or decreases double exponentially, and since the perpendicular susceptibility is proportional to this, the training dynamics become highly unstable or converge rapidly. To experimentally verify this, we designed an experiment using synthetic data. The task involved observing how well the neural network performs regression on synthetic data, where the input data are points in the plane $ (x, y) $ with $ x \sim N(0,1) $ and $ y \sim N(0,1) $, and the target is set to $ x^{2} + y^{2} $. The training dataset is composed of 100 samples. We conducted experiments with a network width of 512 and with 1, 3, and 5 hidden layers, testing cases where $ p = 2 $ and $ p = 3 $. The $ C_{W} $ values were adjusted based on the architecture; if $ C_{W} $ is too small, the kernel decreases double exponentially, and if $ C_{W} $ is too large, the kernel becomes highly unstable and eventually diverges. Therefore, we experimentally determined and set an appropriate $ C_{W} $ value.
We calculated the kernel for an ensemble of 100 neural networks, obtaining the mean and variance while excluding models where the kernel overflowed due to explosion. The results are presented in Tables \ref{t1} and \ref{t2}, with the results for 5 hidden layers and $ p=2 $ depicted in Fig.~\ref{Evolution of Kernel}. We then tested the ensemble of 100 neural network models trained under each setting on a test dataset generated from the same distribution as the training data. The results are depicted in Fig.~\ref{repumeanoutput}.
From the aggregated results of the kernel and the mean outputs on the test dataset, we observed that significant learning only occurs when there is a single hidden layer. When the network has 3 or 5 hidden layers, the kernel values quickly diminish as the layers deepen, leading to poor learning performance. Additionally, as the depth of the network increases, the mean output converges to a constant, with its standard deviation also converging to a very small value.
}
\begin{table}[htp]
\makebox[\textwidth][c]{
\centering
\small
\begin{tabular}{l c c c}
\toprule
\diagbox[width=\dimexpr \textwidth/8+8\tabcolsep\relax, height=1cm]{Epochs}{$(N_{h}, C_{W})$}
  & $(1, 2.5)$
  & $(3, 3.0)$
  & $(5, 2.6)$ \\
\midrule
At initialization
  & $(1.39\times10^{-3},\, 3\times10^{-4})$
  & $(1.28\times10^{-6},\, 1.5\times10^{-6})$
  & $(6.91\times10^{-11},\, 1.9\times10^{-10})$ \\
At 250 epochs
  & $(1.71\times10^{-3},\, 4\times10^{-4})$
  & $(2.80\times10^{-5},\, 1.2\times10^{-5})$
  & $(4.29\times10^{-7},\, 2.7\times10^{-7})$ \\
At 500 epochs
  & $(2.71\times10^{-3},\, 8\times10^{-4})$
  & $(7.44\times10^{-5},\, 5.3\times10^{-5})$
  & $(1.03\times10^{-6},\, 7.2\times10^{-7})$ \\
At 750 epochs
  & $(4.57\times10^{-3},\, 1.5\times10^{-3})$
  & $(1.35\times10^{-4},\, 9.7\times10^{-5})$
  & $(1.75\times10^{-6},\, 1.3\times10^{-6})$ \\
At 1000 epochs
  & $(6.70\times10^{-3},\, 1.6\times10^{-3})$
  & $(2.14\times10^{-4},\, 1.4\times10^{-4})$
  & $(2.62\times10^{-6},\, 2.1\times10^{-6})$ \\
\bottomrule
\end{tabular}
}
\caption{Means and Standard Deviations of the Empirical Kernel at the End of Hidden Layers. Here, \( N_{h} \) denotes the number of hidden layers, and \( C_{W} \) represents hyperparameter for the variance of weight parameters at initialization. The order of RePU is 2 in this case.} \label{t1}
\end{table}

\begin{table}[htp]
\centering
\small
\makebox[\textwidth][c]{
\begin{tabular}{l c c c}
\toprule
\diagbox[width=\dimexpr \textwidth/8+8\tabcolsep\relax, height=1cm]{Epochs}{$(N_{h}, C_{W})$}
  & $(1, 4.5)$
  & $(3, 2.9)$
  & $(5, 2.7)$ \\
\midrule
At initialization
  & $(9.01\times10^{-3},\, 4\times10^{-3})$
  & $(3.49\times10^{-10},\, 1.3\times10^{-9})$
  & $(1.22\times10^{-14},\, 8.3\times10^{-14})$ \\
At 250 epochs
  & $(1.05\times10^{-2},\, 4\times10^{-3})$
  & $(1.34\times10^{-7},\, 1.6\times10^{-7})$
  & $(1.54\times10^{-12},\, 9.0\times10^{-12})$ \\
At 500 epochs
  & $(1.14\times10^{-2},\, 5.1\times10^{-3})$
  & $(2.30\times10^{-7},\, 2.2\times10^{-7})$
  & $(1.91\times10^{-12},\, 1.0\times10^{-11})$ \\
At 750 epochs
  & $(1.24\times10^{-2},\, 6.1\times10^{-3})$
  & $(3.31\times10^{-7},\, 2.9\times10^{-7})$
  & $(2.35\times10^{-12},\, 1.2\times10^{-11})$ \\
At 1000 epochs
  & $(1.36\times10^{-2},\, 7.5\times10^{-3})$
  & $(4.53\times10^{-7},\, 4.0\times10^{-7})$
  & $(2.81\times10^{-12},\, 1.5\times10^{-11})$ \\
\bottomrule
\end{tabular}
}
\caption{Means and Standard Deviations of the Empirical Kernel at the End of Hidden Layers. Here, \( N_{h} \) denotes the number of hidden layers, and \( C_{W} \) represents hyperparameter for the variance of weight parameters at initialization. The order of RePU is 3 in this case.} \label{t2}
\end{table}

\section{Modified Rectified Power Unit (MRePU)}
In this section, we propose the modified Rectified Power Unit (MRePU) function to overcome the limitations of RePU, which cannot be stacked deeply. MRePU retains the advantages of RePU (e.g., differentiability) while fundamentally addressing the existing issues. The definition of MRePU is as follows:

\noindent
{\bf Definition 7 (Modified Rectified Power Unit (MRePU).}{ The MRePU activation function of order $p>1$ is defined by the following equation:

\begin{equation*}
    \begin{split}
        \sigma_{m;p}(z)=\begin{cases}
       z(z+1)^{p} \text{,} &\quad\text{if } z\geq-1\\
       0\text{,} &\quad\text{if }z<-1 \\
     \end{cases}.
    \end{split}
\end{equation*}
Similar to RePU, the MRePU function of order $p$ has the property of being differentiable $p-1$ times.
}

\subsection{Susceptibility Calculation}

We compute the parallel and perpendicular susceptibilities of the proposed MRePU activation. 
For an MRePU of order $p$, they are given by
\begin{equation}
\begin{split}
\chi_{\parallel}(K)
&=\frac{C_{W}}{K}\,\big\langle z\,\sigma'(z)\sigma(z)\big\rangle_{K}
=\frac{C_{W}}{K\sqrt{2\pi K}}\int_{-1}^{\infty} (z+1)^{2p-1}\big((p+1)z+1\big) z^{2}\, e^{-z^{2}/(2K)}\,\mathrm{d}z,\\[2mm]
\chi_{\perp}(K)
&=C_{W}\,\big\langle \sigma'(z)\sigma'(z)\big\rangle_{K}
=\frac{C_{W}}{\sqrt{2\pi K}}\int_{-1}^{\infty} (z+1)^{2p-2}\big((p+1)z+1\big)^{2}\, e^{-z^{2}/(2K)}\,\mathrm{d}z.
\end{split}\label{eq:MREPUsus}
\end{equation}

\noindent\textbf{Proposition 8.}
Let $\phi(t)=\frac{1}{\sqrt{2\pi}}e^{-t^{2}/2}$ and $I_n(\alpha):=\int_{\alpha}^{\infty} y^{n}\phi(y)\,\mathrm{d}y$ with $\alpha=-1/\sqrt{K}$. Then
\begin{equation}
\begin{split}
\chi_{\parallel}(K)&=C_{W}\sum_{n=2}^{2p+2} A_{n}(p)\,K^{\frac{n}{2}-1}\,I_{n}\!\Big(-\frac{1}{\sqrt{K}}\Big),\\
\chi_{\perp}(K)&=C_{W}\sum_{n=0}^{2p} B_{n}(p)\,K^{\frac{n}{2}}\,I_{n}\!\Big(-\frac{1}{\sqrt{K}}\Big),
\end{split} \label{suscepmrepu}
\end{equation}
where, with the convention $\binom{m}{r}=0$ if $r\notin\{0,\dots,m\}$,
\[
A_{n}(p)=(p+1)\binom{2p-1}{\,n-3\,}+\binom{2p-1}{\,n-2\,},\qquad
B_{n}(p)=(p+1)^{2}\binom{2p-2}{\,n-2\,}+2(p+1)\binom{2p-2}{\,n-1\,}+\binom{2p-2}{\,n\,}.
\]

\noindent\emph{Proof.}
(1) \emph{Polynomial expansion.}
Write
\[
(z+1)^{2p-1}\big((p+1)z+1\big)z^{2}
=(p+1)\sum_{j=0}^{2p-1}\binom{2p-1}{j}z^{j+3}
+\sum_{j=0}^{2p-1}\binom{2p-1}{j}z^{j+2}.
\]
Collecting the $z^{n}$ terms gives the coefficient in front of $z^{n}$ as
$A_n(p)=(p+1)\binom{2p-1}{n-3}+\binom{2p-1}{n-2}$.
Similarly,
\[
(z+1)^{2p-2}\big((p+1)z+1\big)^{2}
=(p+1)^2\sum_{j=0}^{2p-2}\binom{2p-2}{j}z^{j+2}
+2(p+1)\sum_{j=0}^{2p-2}\binom{2p-2}{j}z^{j+1}
+\sum_{j=0}^{2p-2}\binom{2p-2}{j}z^{j},
\]
whence the $z^{n}$ coefficient is
$B_n(p)=(p+1)^2\binom{2p-2}{n-2}+2(p+1)\binom{2p-2}{n-1}+\binom{2p-2}{n}$.

(2) \emph{Reduction to $I_n$.}
For any integer $n\ge0$, the change of variables $y=z/\sqrt{K}$ yields
\[
\int_{-1}^{\infty} z^{n}\frac{e^{-z^{2}/(2K)}}{\sqrt{2\pi K}}\,\mathrm{d}z
=K^{n/2}\int_{-1/\sqrt{K}}^{\infty} y^{n}\phi(y)\,\mathrm{d}y
=K^{n/2}I_n\!\Big(-\frac{1}{\sqrt{K}}\Big).
\]
Combining (1)–(2) and summing termwise gives the stated formulas.
\hfill$\square$

\noindent\textbf{Corollary 9 (small-$K$ limit).}
As $K\to0$,
\[
\chi_{\parallel}(K)\;\longrightarrow\;C_W,\qquad
\chi_{\perp}(K)\;\longrightarrow\;C_W.
\]

\noindent\emph{Proof.}
Let $\beta=1/\sqrt{K}\to\infty$. Since $I_n(-\beta)=\int_{-\beta}^{\infty}y^n\phi(y)\,\mathrm{d}y$ and $\phi$ is integrable with polynomial weights, dominated convergence gives
\[
\lim_{\beta\to\infty} I_n(-\beta)=\int_{-\infty}^{\infty}y^n\phi(y)\,\mathrm{d}y
=\begin{cases}
0,& n\ \text{odd},\\
( n-1)!!,& n\ \text{even}.
\end{cases}
\]
In $\chi_{\parallel}(K)=C_W\sum_{n=2}^{2p+2} A_n(p)\,K^{\frac{n}{2}-1} I_n(-\beta)$,
all terms with $n>2$ vanish because $K^{\frac{n}{2}-1}\to0$, while for $n=2$ we have
$K^{0}=1$, $I_2(-\beta)\to 1$, and $A_2(p)=(p+1)\binom{2p-1}{-1}+\binom{2p-1}{0}=1$.
Hence $\lim_{K\to0}\chi_{\parallel}(K)=C_W$.
Likewise, in $\chi_{\perp}(K)=C_W\sum_{n=0}^{2p} B_n(p)\,K^{\frac{n}{2}} I_n(-\beta)$,
the leading term is $n=0$ with $K^{0}=1$, $I_0(-\beta)\to1$, and
$B_0(p)=(p+1)^2\binom{2p-2}{-2}+2(p+1)\binom{2p-2}{-1}+\binom{2p-2}{0}=1$;
all $n\ge1$ terms vanish.
Thus $\lim_{K\to0}\chi_{\perp}(K)=C_W$.
\hfill$\square$

\paragraph{Numerics.}
Fig.~4 shows $\chi_\parallel$, $\chi_\perp$ and their ratio for $C_W=1$; as $K\to0$ the ratio tends to $1$, in agreement with Corollary~9. In this limit both susceptibilities equal those of a linear activation. The next order, however, has the opposite sign to that of the $K^\star=0$ universality class of \citep{Roberts22}. Writing $\sigma(z)=\sum_k\sigma_kz^k/k!$, that class has $a_1=\sigma_3/\sigma_1+\tfrac34(\sigma_2/\sigma_1)^2<0$ (e.g.\ $a_1=-2$ for tanh), so that a small kernel decays toward zero. MRePU has $a_1=3p(2p-1)>0$, so that at $C_W=1$ a small kernel grows. Section~4.3 quantifies this growth and shows how to control it.

\begin{figure}[htp]
\centering
\minipage{0.49\textwidth}
\includegraphics[scale=0.29]{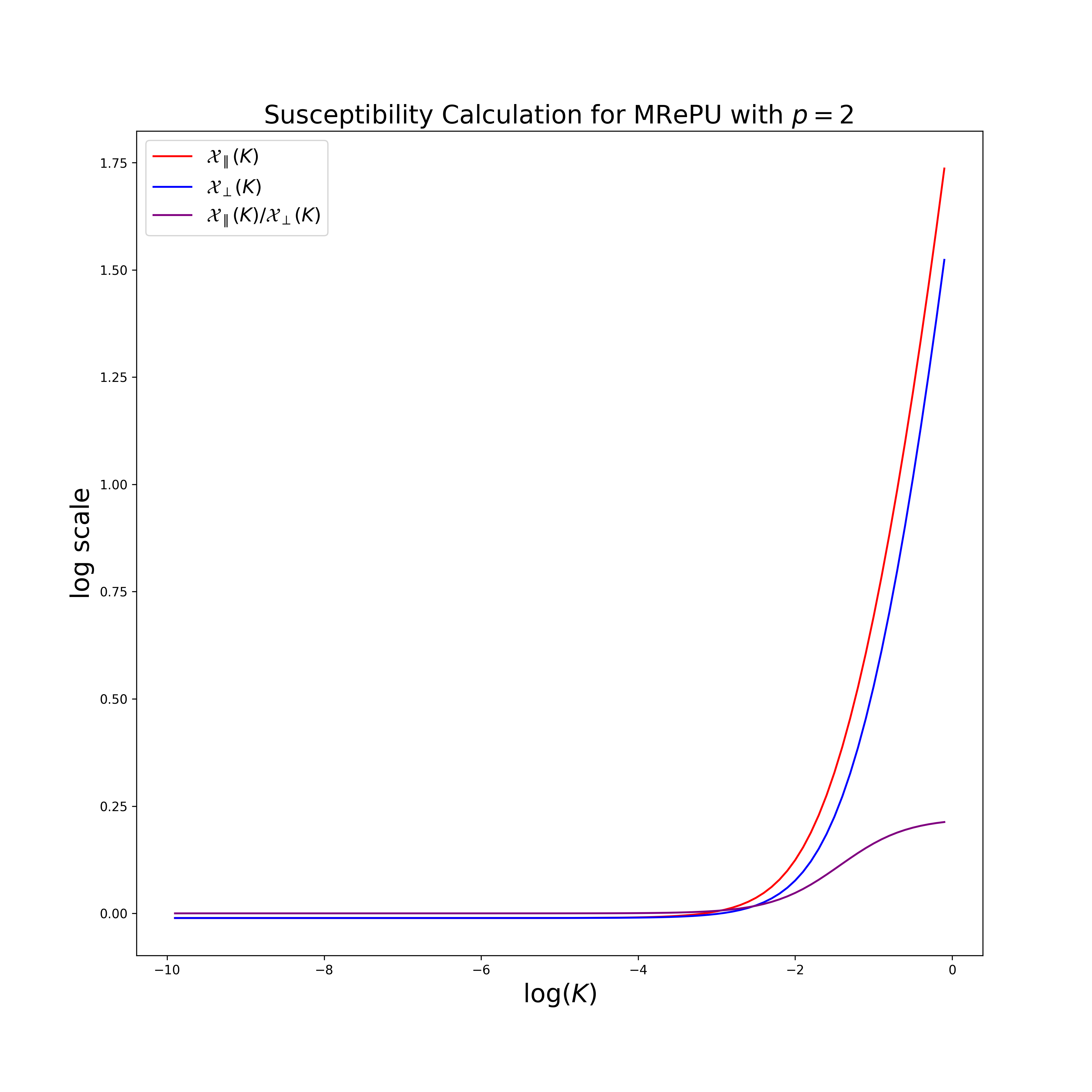}
\endminipage\hfill
\minipage{0.49\textwidth}
\includegraphics[scale=0.29]{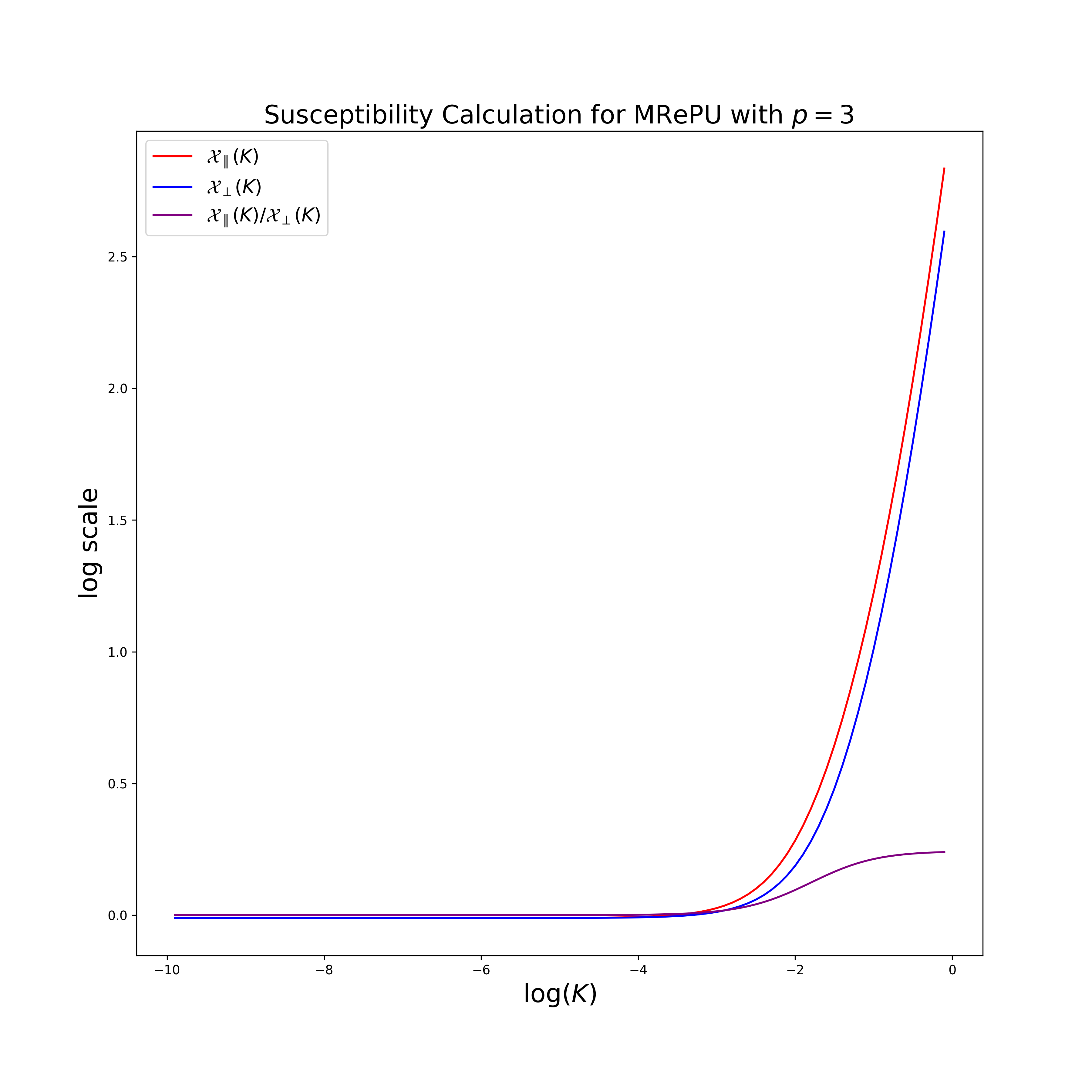}
\endminipage
\caption{\textbf{Parallel and perpendicular susceptibilities and their ratio} for MRePU with $p=2$ (left) and $p=3$ (right), with $(C_{W},C_{b})=(1,0)$. The ratio tends to $1$ as $K\to0$.}
\label{MREPUsus}
\end{figure}

\subsection{Experimental Results}

Now, as we did with RePU Neural Networks, we will investigate the statistical properties of MRePU under the same experimental settings.

\noindent
{\bf Empirical Kernel and Four-point Vertex Behavior at Initialization.}{ 
As with the RePU case, we investigate the statistical properties of the neural network ensemble given by the initialization distribution. The experimental environment remains identical to that of the RePU Networks, with the only change being the activation function, which is now MRePU with $ p=2$. The results for $ C_{W}=1.0 $ are shown in Fig.~\ref{mrepuinit}. As can be seen from the graph, the empirical kernel maintains a consistent scale across both random data and weight parameters. Additionally, we also examined the behavior of the Four-point vertex for MRePU. Eq.~\eqref{vertexrecur} can be simplified for a single input as follows:
\begin{equation}
    V^{(l+1)}=\chi_{\parallel}^{2}(K^{(l)})V^{(l)}+C_{W}^{2}\Big[\langle\sigma^{4}(z)\rangle_{K^{(l)}}-\langle\sigma^{2}(z)\rangle_{K^{(l)}}^{2}\Big]+O(\frac{1}{n_{l}}). \label{simplifiedrecur}
\end{equation}
Fig.~\ref{fourvertexatinit} illustrates how the empirical four-point vertex, calculated from an ensemble of 1000 models using Eq.~\eqref{fourvert}, changes across layers (with input $x_{0}=(1,0)$). $\chi_{\parallel}$ increases as the kernel grows, and since the kernel itself depends on $C_{W}$, the inductive change in the four-point vertex is determined by $C_{W}$ according to Eq.~\eqref{simplifiedrecur}. As evidence of this, the left figure shows that the four-point vertex increases more steeply as $C_{W}$ becomes larger. We also investigated the behavior of the four-point vertex across different widths. The results indicate that with smaller widths, perturbative effects in Eq.~\eqref{simplifiedrecur} become more pronounced, leading to a more rapid increase in the four-point vertex, and overall, higher values are observed. The four-point vertex serves as an indicator of non-Gaussianity, which is expected to increase as more layers with non-linear activations are added. Moreover, since smaller widths imply stronger interactions between neurons, these experimental results align well with our theoretical predictions.

\begin{figure}[htp]
\centering
\minipage{0.49\textwidth}
  \includegraphics[scale=0.40]{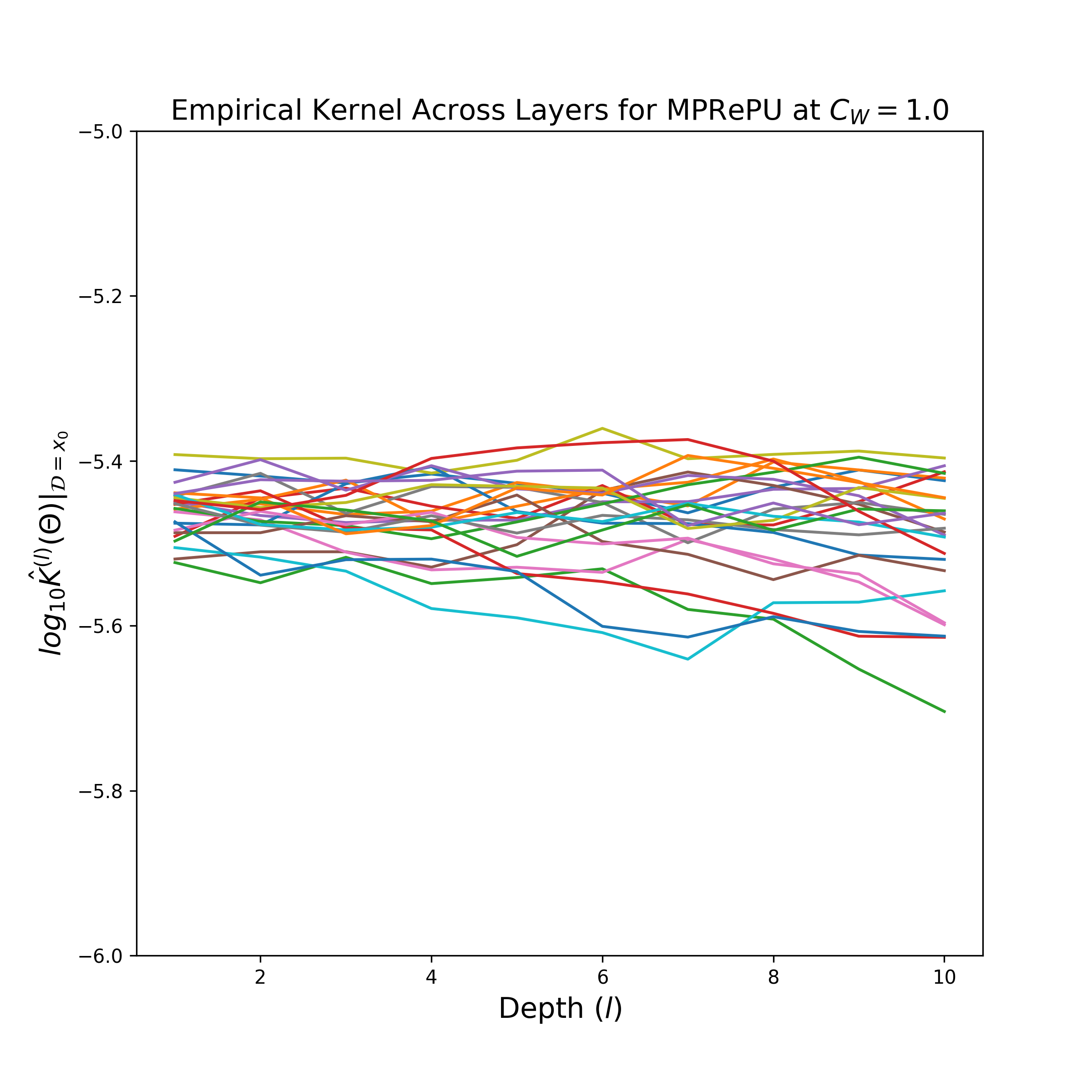}
\endminipage\hfill
\minipage{0.49\textwidth}
  \includegraphics[scale=0.40]{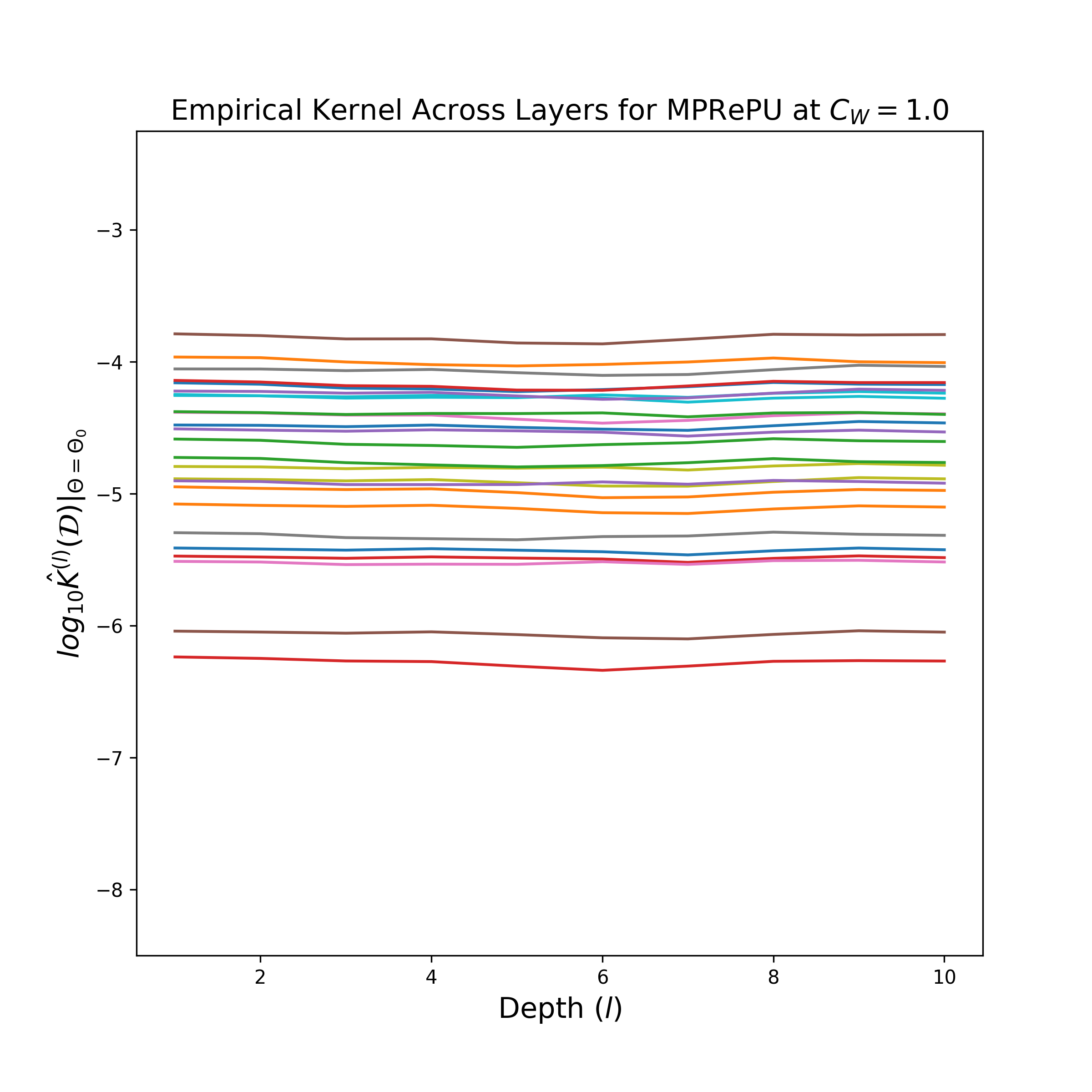}
\endminipage
\caption{Empirical Kernels at Initialization Across Layers for MRePU Activation with $p=2$. \textbf{Left}: Data is fixed and randomness is in the weight parameters. \textbf{Right}: Weight parameters are fixed and data is random. Each line represents a sample.}\label{mrepuinit}
\end{figure}

\begin{figure}[htp]
\centering
\minipage{0.49\textwidth}
  \includegraphics[scale=0.29]{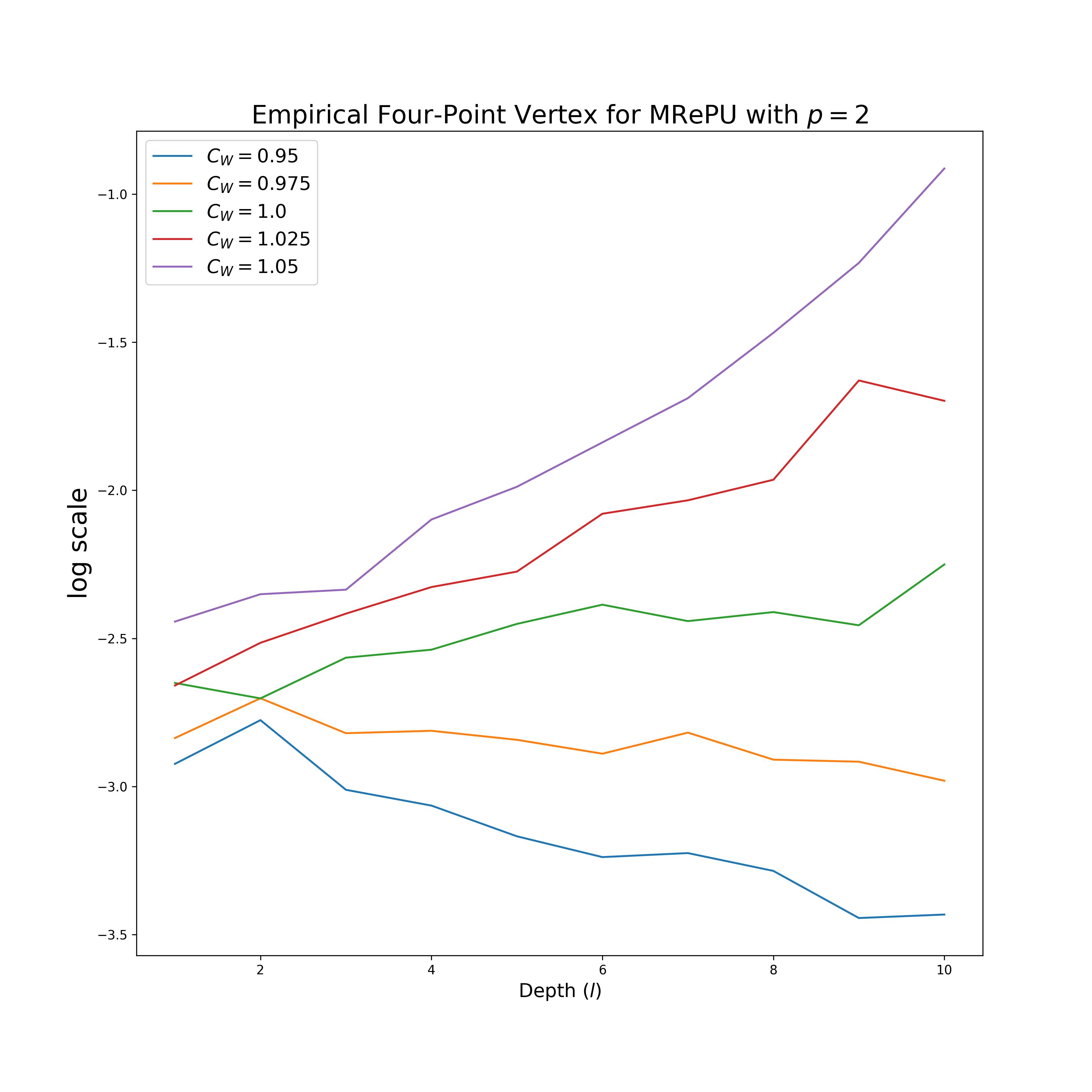}
\endminipage\hfill
\minipage{0.49\textwidth}
  \includegraphics[scale=0.29]{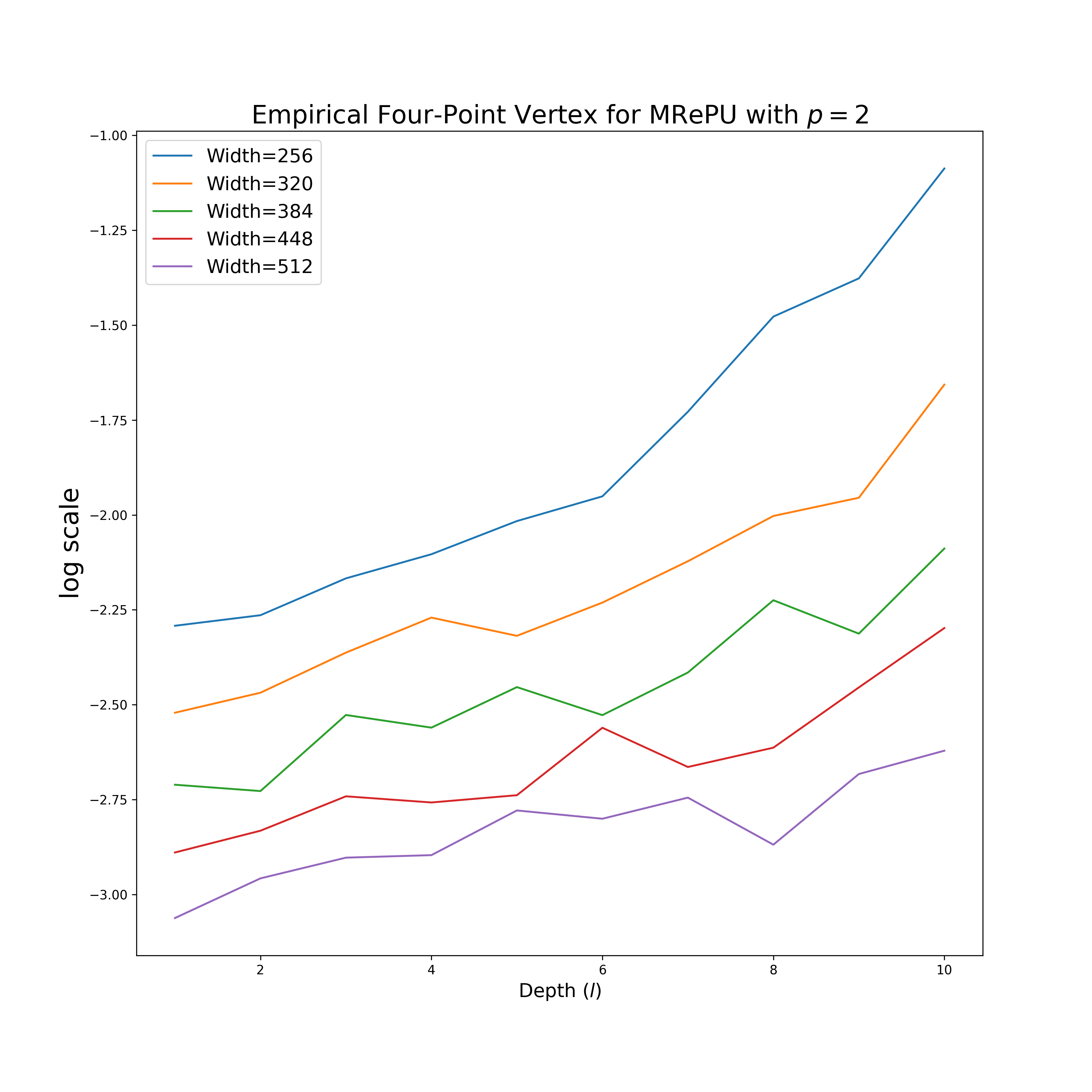}
\endminipage
\caption{Empirical Four-Point Vertex Across Layers at Initialization. Each value is calculated over an ensemble of 1000 models. \textbf{Left}: Width is fixed at 512, and $C_{W}$ is varying. \textbf{Right}: $C_{W}$ is fixed at $1.0$, and the width is varying.}\label{fourvertexatinit}
\end{figure}
}

\noindent
{\bf Dynamics of Statistics During Training.}{
Now, we will examine how the statistics of the MRePU Network ensemble change under training dynamics. The experimental setup is the same as in the case of the RePU Network, and the number of hidden layers is set to 5. In this case, we constructed the training dataset in the same way as in the RePU activation case, but the number of samples is set to 1000. Although the theoretical approximation suggests setting $C_{W} = 1.0$, meaningful learning was only achieved with a slightly lower value, possibly due to the perturbative effects in the MRePU Network. In our experiments, we set $C_{W} = 0.8$. The dynamics of the kernel and four-point vertex are depicted in Fig.~$\ref{evolofkernelandvertex}$, while the comparison of the ensemble’s mean output against the target for the test dataset is shown in Fig.~$\ref{Evolution of meanouputs for mrepu}$. Surprisingly, we observed that even with a vanilla FCN (Fully Connected Network) without considering architectures like ResNet, which are designed to prevent issues like the vanishing gradient, the deep neural network with 5 hidden layers still achieved meaningful learning using the MRePU function we designed. Additionally, as shown on the left side of Fig.~$~\ref{evolofkernelandvertex}$, we also calculated the perpendicular susceptibility, which showed a tendency to approach 1 as training progressed. On the right side of the same figure, it can be seen that the ensemble’s four-point vertex decreases to 0 during initialization but tends to increase linearly with the layers as training progresses. An increasing four-point vertex indicates growing interactions between neurons, which implies that representation learning is occurring. This suggests that the deep vanilla MRePU network is indeed engaging in meaningful representation learning.

\begin{figure}[htp]
\centering
\minipage{0.49\textwidth}
  \includegraphics[scale=0.40]{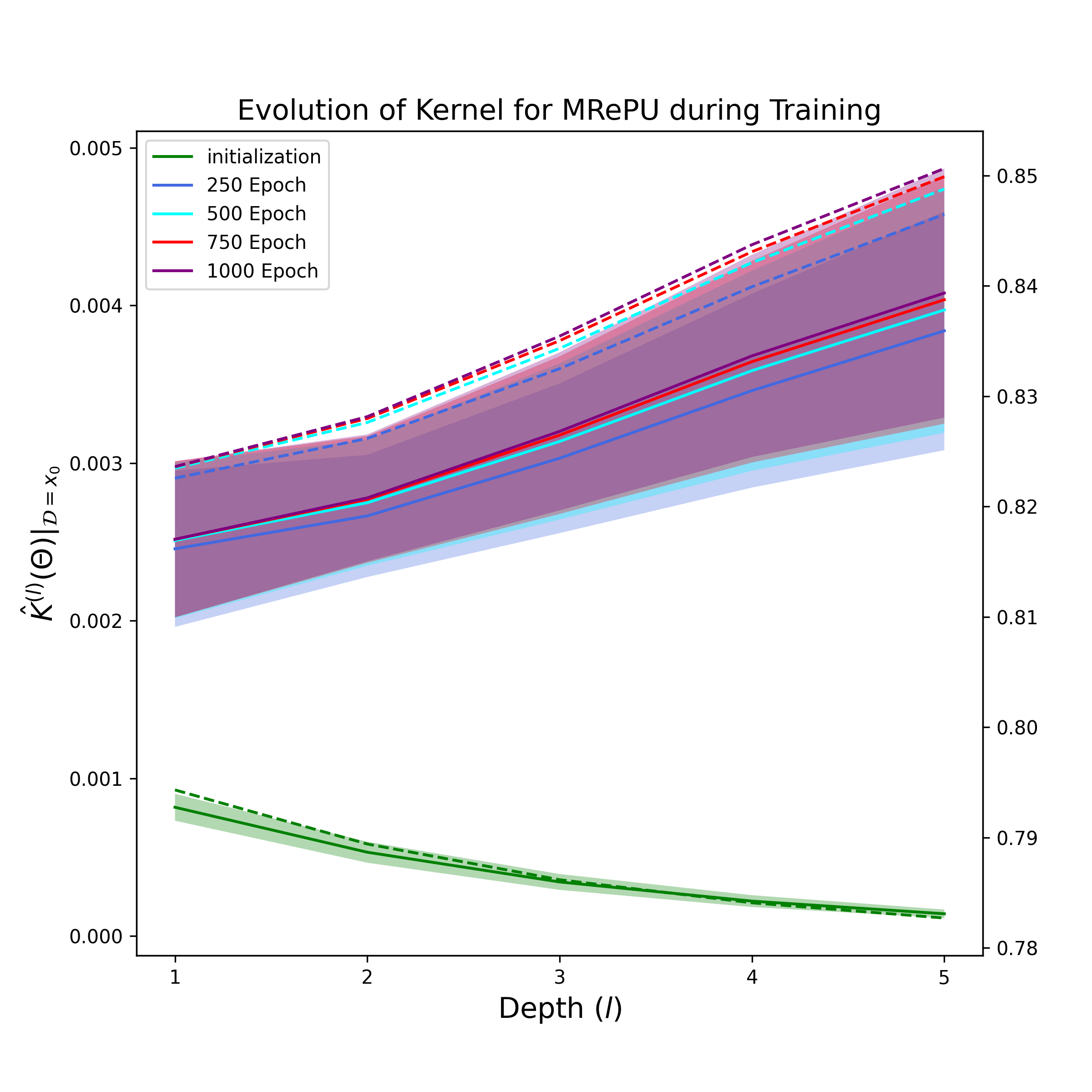}
\endminipage\hfill
\minipage{0.49\textwidth}
  \includegraphics[scale=0.40]{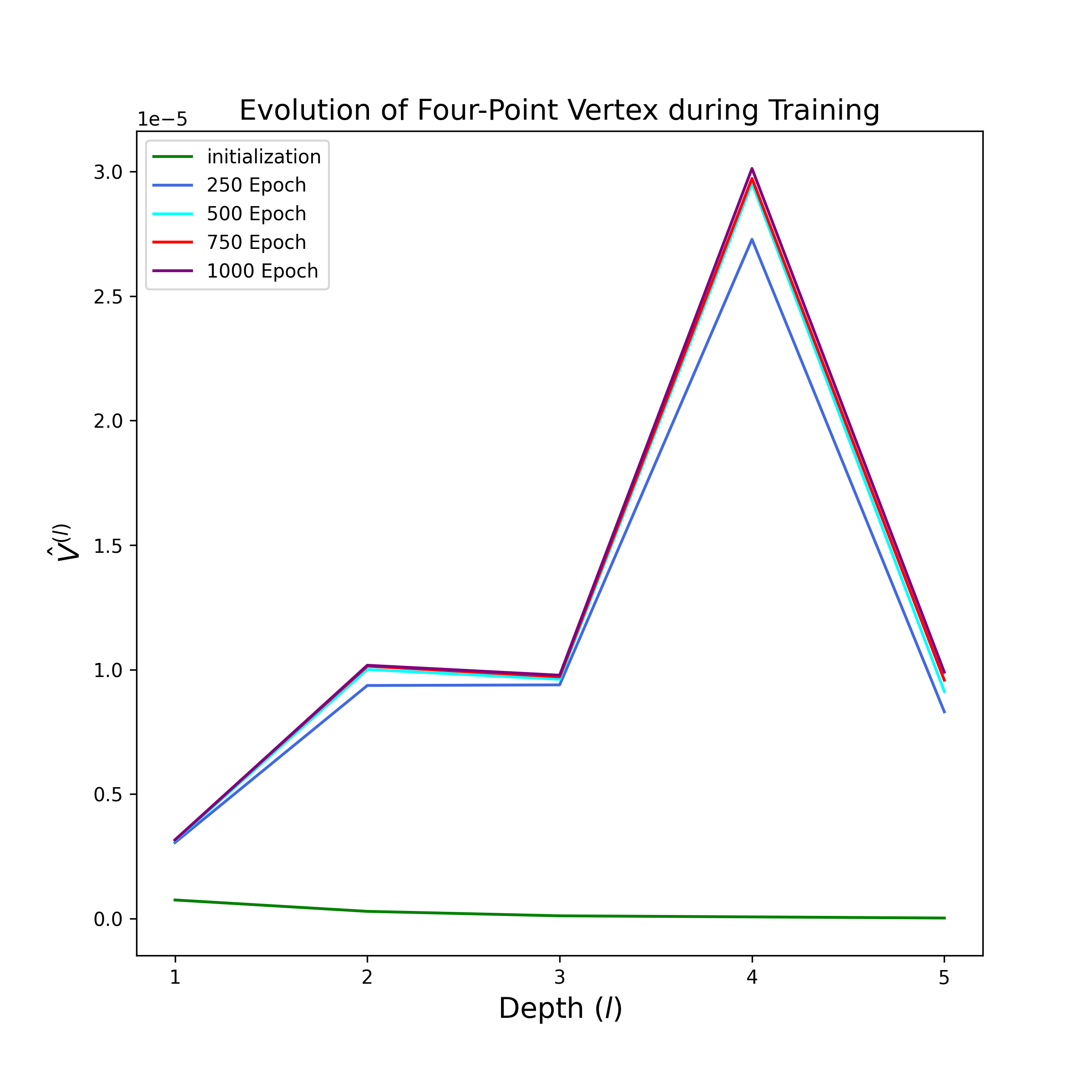}
\endminipage
\caption{The evolution of the mean of empirical kernels (solid line, left y-axis) and perpendicular susceptibility (dotted line, right y-axis) (\textbf{Left}) and the four-point vertex (\textbf{Right}) over an ensemble of 100 models for $x_{0} = (1,0)$ as training progresses for the MRePU activation with  p=2. The shaded areas represent the region between $\log_{10}(\text{mean} \pm \text{standard deviation})$ (for \textbf{Left}).}\label{evolofkernelandvertex}
\end{figure}

\begin{figure}[htp]
\centering
\includegraphics[scale=0.6]{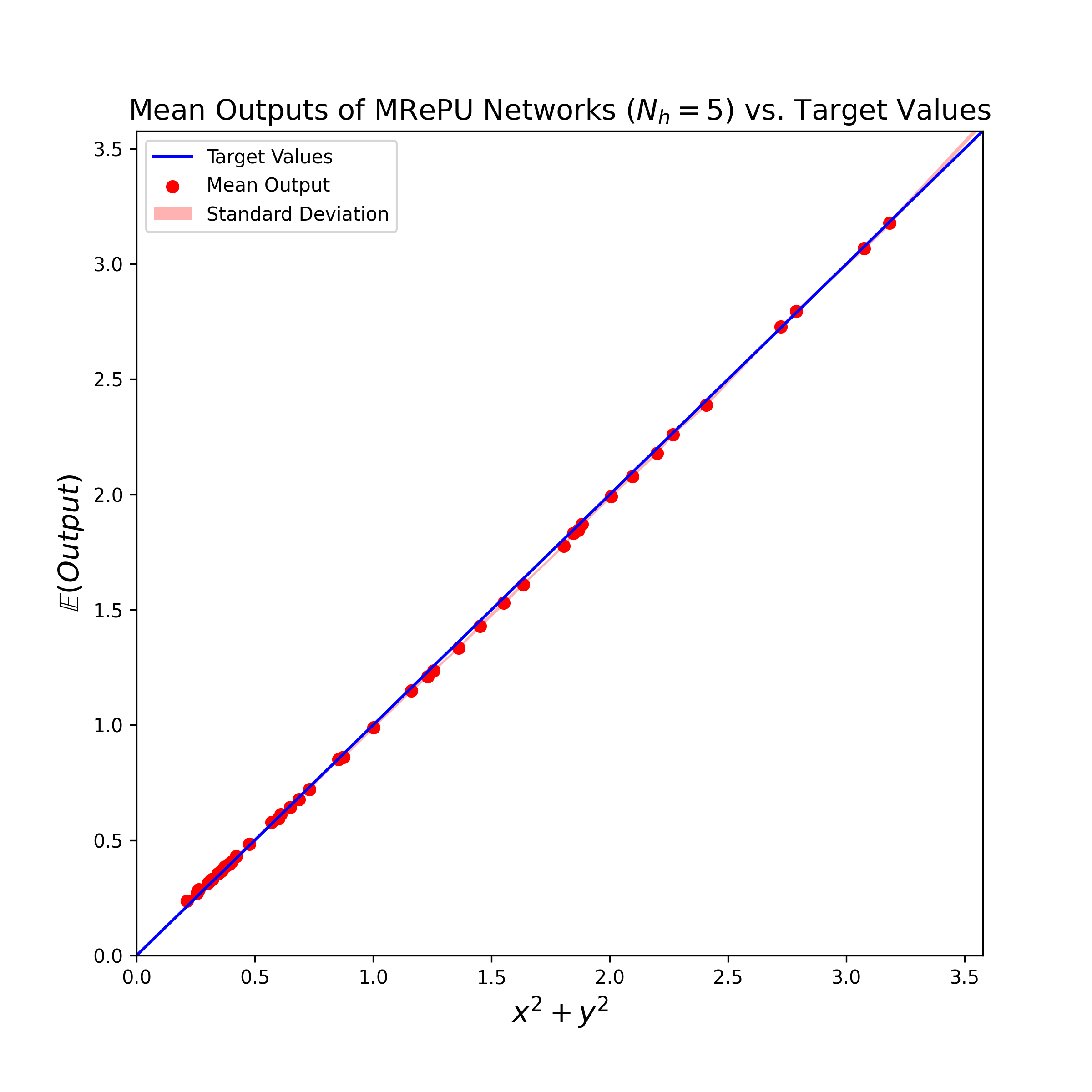}
\caption{Mean outputs over an ensemble of 100 models versus target values on random test data. The shaded areas represent the region of 1 standard deviation. The number of hidden layers is 5.}\label{Evolution of meanouputs for mrepu}
\end{figure}
}
\noindent

\subsection{Hyperparameter Criteria from Kernel Analysis}
Our kernel analysis yields guidelines for configuring MRePU networks. First, $C_W=1$ is critical only as $K\to0$ (Corollary~9), and at $C_W=1$ a small kernel grows (Section~4.1). Consequently, the input, and hence the kernel at the first layer, must be small for the kernel to stay at its initial scale over many layers. Next, at a fixed small $K$, Eq.~(9) shows that the choice
\begin{equation}
    \begin{split}
        C_{W}\simeq\frac{1}{\sum_{n=2}^{2p+2}A_{n}(p)K^{\frac{n}{2}-1}I_{n}\Big(-\frac{1}{\sqrt{K}}\Big)}\simeq\frac{1}{\sum_{n=0}^{2p}B_{n}(p)K^{\frac{n}{2}}I_{n}\Big(-\frac{1}{\sqrt{K}}\Big)}.
    \end{split} \label{criti}
\end{equation}
sets $\chi_\parallel(K)$, respectively $\chi_\perp(K)$, to one, and both tend to the critical value $C_W=1$ as $K\to0$. Corollary~9, however, concerns only the limit $K\to0$; it does not show that the kernel recursion stays stable at a positive $K$ over many layers. In the rest of this subsection we make the guideline precise.
We present three rules for setting $C_W$, together with an algorithm that implements them,
and we analyze the stability of a one-parameter family of $C_W$ by solving ODE which is continuation version of kernel recursion flow. This analysis contains all three rules as special cases.
Since Corollary~9 describes only the limit $K\rightarrow0$, we expand the kernel map and the
susceptibilities at finite $K$ and use these expansions to design concrete rules for choosing $C_W$.
Up to terms of order $e^{-\frac{1}{2K}}$, the quantities $g(K)$ and $\langle\sigma'(z)^{2}\rangle_{K}$
of the MRePU activation expand as
\begin{equation}
\begin{split}
        g(K)=K+c_pK^{2}+O(K^{3}),&\quad \langle \sigma'(z)^{2}\rangle_{K}=1+B_{2}(p)K+O(K^{2}), \\
        c_p=3p(2p-1),&\quad B_{2}(p)=7p^2-3p. \label{pertur}
\end{split}
\end{equation}
Since $c_p>0$, the choice $C_W=1$ does not make $K$ a fixed point of the kernel recursion.
Specifically, the recursion $K^{(l+1)}=C_{W}g(K^{(l)})$ yields the approximation
$K^{(l)}\simeq \frac{K^{(1)}}{1-c_{p}K^{(1)}l}$.
Motivated by this expression, we define the following rescaled depth for an MRePU network of depth $L$:
\begin{equation}
    \tau=c_pK^{(1)}L.
\end{equation}
\paragraph{Three criticality rules.}
We consider the three choices of $C_W$ that each make one quantity exact during the recursion of points $K^{(n)}$ through the layers:
\begin{align}
  C_W^{(l)} &= \frac{1}{g'(K^{(l)})}, && \chi_\parallel(K^{(1)})=1, \tag{R1}\label{eq:rule-par}\\
   C_W^{(l)}  &= \frac{1}{\langle\sigma'(z)^2\rangle_{K^{(l)}}}, && \chi_\perp(K^{(1)})=1, \tag{R2}\label{eq:rule-perp}\\
  C_W &= \frac{K^{\star}}{g(K^{\star})}, &&  \text{(fixed point)}. \tag{R3}\label{eq:rule-fp}
\end{align}
Rules \eqref{eq:rule-par} and \eqref{eq:rule-perp} are the guideline of Eq.~\eqref{criti}, now evaluated at the operating point $K^{(1)}$.
All three reduce to the critical value $C_W=1$ of Corollary~9 as $K\to0$, and by \eqref{pertur} they are members of the one-parameter family
\begin{equation}\label{eq:kappa}
  C_W ^{(l)}= 1-\kappa\, c_p K^{(l)} + O((K^{(l)})^2), \qquad
  \kappa = 2\ \text{\eqref{eq:rule-par}},\quad \kappa = r_p=B_2(p)/c_p\ \text{\eqref{eq:rule-perp}},\quad \kappa = 1\ \text{\eqref{eq:rule-fp}}.
\end{equation}
with $C_W=1$ corresponding to $\kappa=0$. Since $c_p<B_2(p)<2c_p$ ($B_2(p)/c_p = 11/9$ for $p=2$ and $6/5$ for $p=3$), rule \eqref{eq:rule-perp} lies between the other two.
The stability of this one-parameter family can now be summarized in the following proposition.

\noindent\textbf{Proposition 10.}
Let $C_b=0$, $K_{0}=K^{(1)}$ and $C_W = 1-\kappa c_pK_0+O(K_0^2)$ with $\kappa\ge0$, and consider the infinite-width kernel recursion in the limit $K_0\to0$, $L\to\infty$ with $\tau=c_pK_0L$ fixed. Then
\begin{equation}\label{eq:prop-rules}
  \frac{K^{(L)}}{K_0} \to u_\kappa(\tau) := \frac{\kappa}{1+(\kappa-1)\,e^{\kappa\tau}}, \qquad
  \prod_{\ell=1}^{L}\chi_\perp(K^{(\ell)}) \to e^{\kappa(r_p-1)\tau}\, u_\kappa(\tau)^{r_p},
\end{equation}
with $u_0(\tau) = (1-\tau)^{-1}$. Moreover:
\begin{enumerate}
  \item[(i)] for $0\le\kappa<1$, \eqref{eq:prop-rules} holds for $\tau<\tau_c(\kappa) = -\log(1-\kappa)/\kappa$ and both quantities diverge as $\tau\to\tau_c$ (so $\tau_c=1$ for $C_W=1$);
  \item[(ii)] for $\kappa\ge1$ both stay finite for every $\tau$; the kernel is constant if $\kappa=1$ and decreases monotonically if $\kappa>1$;
  \item[(iii)] for $\kappa>0$ the map $K\mapsto C_W g(K)$ has a single nonzero fixed point near the origin, at $K_\star = \kappa K_0\,(1+O(K_0))$, which is unstable; moreover
        $\chi_\parallel(K_0) = 1+(2-\kappa)c_pK_0+O(K_0^2)$ and $\chi_\perp(K_0) = 1+(B_2(p)-\kappa c_p)K_0+O(K_0^2)$.
\end{enumerate}

\noindent
{\bf Proof.}{ 
Throughout we write
\[
  \varepsilon = c_pK_0, \qquad r = B_2(p)/c_p, \qquad u_\ell = K^{(\ell)}/K_0 ,
\]
and take $\varepsilon\to0$, $L\to\infty$ with $\tau=\varepsilon L$ fixed.
As long as $u_\ell$ stays bounded, $K^{(\ell)}=O(K_0)$, so the corrections of order $e^{-1/(2K)}$ in \eqref{pertur} are smaller than any power of $\varepsilon$ and can be dropped.
 
\smallskip\noindent\textbf{Step 1: reduction to an ODE.}
Since $C_W = 1-\kappa\varepsilon+O(\varepsilon^2)$ and $g(K_0u) = K_0u\,(1+\varepsilon u+O(\varepsilon^2))$ for bounded $u$,
\[
  u_{\ell+1} = \frac{C_W\,g(K_0u_\ell)}{K_0}
  = (1-\kappa\varepsilon)\,u_\ell\,(1+\varepsilon u_\ell) + O(\varepsilon^2)
  = u_\ell + \varepsilon\,u_\ell(u_\ell-\kappa) + O(\varepsilon^2).
\]
This is the Euler scheme with step $\varepsilon$ for
\begin{equation}\label{eq:u-ode}
  \dot u = u\,(u-\kappa), \qquad u(0)=1.
\end{equation}
The local error is $O(\varepsilon^2)$ and there are $\tau/\varepsilon$ steps, so $u_\ell - u(\varepsilon\ell) = O(\varepsilon)$ uniformly on every interval $[0,\tau]$ on which the solution $u$ is bounded.
In particular $K^{(L)}/K_0 = u_L \to u(\tau)$.
 
\smallskip\noindent\textbf{Step 2: solving the ODE.}
For $\kappa=0$, \eqref{eq:u-ode} reads $\dot u=u^2$, so $u(t)=1/(1-t)$.
For $\kappa>0$, the variable $w = (\kappa-u)/u = \kappa/u-1$ satisfies
\[
  \dot w = -\frac{\kappa\,\dot u}{u^2} = \frac{\kappa\,(\kappa-u)}{u} = \kappa\,w, \qquad w(0)=\kappa-1,
\]
so $w(t) = (\kappa-1)\,e^{\kappa t}$, i.e.
\[
  u(t) = \frac{\kappa}{1+(\kappa-1)\,e^{\kappa t}} = u_\kappa(t).
\]
This is the first limit in \eqref{eq:prop-rules}.
 
\smallskip\noindent\textbf{Step 3: dependence on $\kappa$, items (i) and (ii).}
If $0<\kappa<1$, the denominator $1-(1-\kappa)e^{\kappa t}$ decreases and vanishes at $t=\tau_c(\kappa)=-\log(1-\kappa)/\kappa$, where $u$ blows up.
If $\kappa=0$, $u$ blows up at $t=1$, which is the limit of $\tau_c(\kappa)$ as $\kappa\to0$.
In both cases Step~1 applies on $[0,\tau]$ for every $\tau<\tau_c$, and both limits in \eqref{eq:prop-rules} diverge as $\tau\to\tau_c$ because $u_\kappa(\tau)\to\infty$; this is (i).
If $\kappa\ge1$, the denominator is at least $1$ for all $t$, so $u$ is bounded on $[0,\infty)$.
For $\kappa=1$, $u\equiv1$.
For $\kappa>1$, $w = \kappa/u-1 = (\kappa-1)e^{\kappa t}$ increases with $t$, so $u$ decreases from $u(0)=1$ and stays positive; this is (ii).
 
\smallskip\noindent\textbf{Step 4: the gradient factor.}
For $K=K_0u$ with $u$ bounded, \eqref{pertur} gives $\langle\sigma'(z)^2\rangle_K = 1+B_2(p)K_0u+O(\varepsilon^2) = 1+r_p\varepsilon u+O(\varepsilon^2)$, hence
\[
  \log\chi_\perp(K^{(\ell)}) = \log C_W + \log\langle\sigma'(z)^2\rangle_{K^{(\ell)}}
  = \varepsilon\,(r_p\,u_\ell-\kappa) + O(\varepsilon^2).
\]
Summing over the $L=\tau/\varepsilon$ layers, the $O(\varepsilon^2)$ terms add up to $O(\varepsilon)$, and the rest is a Riemann sum:
\[
  \log\prod_{\ell=1}^{L}\chi_\perp(K^{(\ell)}) \to \int_0^\tau \bigl(r_p\,u(t)-\kappa\bigr)\,dt.
\]
By \eqref{eq:u-ode}, $\frac{d}{dt}\log u = u-\kappa$, so $\int_0^\tau u\,dt = \kappa\tau+\log u(\tau)$ and
\[
  \int_0^\tau \bigl(r_p\,u(t)-\kappa\bigr)\,dt = \kappa(r-1)\,\tau + r_p\log u_\kappa(\tau).
\]
Exponentiating gives the second limit in \eqref{eq:prop-rules}.
 
\smallskip\noindent\textbf{Step 5: fixed point and susceptibilities at $K_0$, item (iii).}
Let $\kappa>0$. A fixed point $K\neq0$ of $K\mapsto C_Wg(K)$ satisfies $C_W\,g(K)/K=1$, which by \eqref{pertur} reads
\[
  \bigl(1-\kappa c_pK_0+O(K_0^2)\bigr)\bigl(1+c_pK+O(K^2)\bigr) = 1.
\]
Its solution near the origin is $c_pK = \kappa c_pK_0 + O(K_0^2)$, i.e.\ $K_\star = \kappa K_0\,(1+O(K_0))$.
Since $g(K)/K = 1+c_pK+O(K^2)$ is increasing near $0$, we have $C_Wg(K)>K$ just above $K_\star$ and $C_Wg(K)<K$ just below it, so $K_\star$ is unstable.
Finally, multiplying $g'(K_0) = 1+2c_pK_0+O(K_0^2)$ and $\langle\sigma'(z)^2\rangle_{K_0} = 1+B_2(p)K_0+O(K_0^2)$ by $C_W = 1-\kappa c_pK_0+O(K_0^2)$ gives
\[
  \chi_\parallel(K_0) = 1+(2-\kappa)\,c_pK_0+O(K_0^2), \qquad
  \chi_\perp(K_0) = 1+(B_2(p)-\kappa c_p)\,K_0+O(K_0^2).
\]
\hfill$\square$
}

To first order in $\tau$, the proposition gives $K^{(L)}/K_0\simeq1+(1-\kappa)\tau$ and
$\prod_\ell\chi_\perp\simeq1+(r_p-\kappa)\tau$, which for the three rules reads as follows.
\begin{center}
\begin{tabular}{lcccc}
 & $\kappa$ & $K_\star/K_0$ & $K^{(L)}/K_0$ & $\prod_\ell\chi_\perp$ \\ \hline
 $C_W=1$ & $0$ & -- & $1+\tau$ & $1+r_p\tau$ \\
 \eqref{eq:rule-fp} fixed point & $1$ & $1$ & $1$ & $1+(r_p-1)\tau$ \\
 \eqref{eq:rule-perp} $\chi_\perp=1$ & $r_p$ & $r_p$ & $1-(r_p-1)\tau$ & $1+O(\tau^2)$ \\
 \eqref{eq:rule-par} $\chi_\parallel=1$ & $2$ & $2$ & $1-\tau$ & $1-(2-r_p)\tau$
\end{tabular}
\end{center}
For $\tau\ll1$ the three rules agree up to relative corrections of order $\tau$, and all keep the kernel and the gradient factor at their initial values; this is the regime in which the guideline of Eq.~(11) is effective at initialization. For larger $\tau$ they differ: $C_W=1$ diverges at $\tau=1$, rule (R3) keeps the kernel but lets the gradient factor grow as $e^{(r_p-1)\tau}$, and rule (R1) stays finite for every $\tau$ but its kernel and gradient factor decay like $e^{-2\tau}$.
Thus $\kappa$ trades preservation against margin.
Rule~\eqref{eq:rule-fp} ($\kappa=1$) keeps the kernel exactly at $K_0$, whereas the product of $\chi_\perp$,
which sets the size of the gradients at initialization, equals $1$ to first order in $\tau$ at $\kappa=r_p$
($11/9$ for $p=2$, $6/5$ for $p=3$); since $r_p>1$, no single value of $\kappa$ preserves both.
At the same time, the unstable fixed point $K_\star=\kappa K_0$ lies above the operating point by $(\kappa-1)K_0$,
so a larger $\kappa$ leaves more room for the kernel to grow, through finite-width fluctuations or during training,
before the recursion diverges.
Rule~\eqref{eq:rule-fp} has no such margin, while rule~\eqref{eq:rule-par} has the largest one,
at the price of a kernel and a gradient factor that decrease with $\tau$
(as $1-\tau$ and $1-(2-r_p)\tau$ for small $\tau$, and like $e^{-2\tau}$ for large $\tau$).
Algorithm~\ref{alg:recipe} turns the three rules into an executable procedure.
\begin{algorithm}[htp]
\caption{Initialization of an MRePU network with a rescaled depth $\tau$}\label{alg:recipe}
\begin{algorithmic}[1]
\Require degree $p$; number $L$ of MRePU layers on the longest input--output path
\Statex \hspace{\algorithmicindent} input second moment $q=\mathbb{E}[\|x\|^2]/d$ (inputs $x\in\mathbb{R}^d$)
\Statex \hspace{\algorithmicindent} rescaled depth $\tau$ 
\Statex \hspace{\algorithmicindent} rule \eqref{eq:rule-par}, \eqref{eq:rule-perp} or \eqref{eq:rule-fp}
\State $c_p \gets 3p(2p-1)$, \quad $L_{\mathrm{eff}} \gets L$ \Comment{residual nets: Remark~\ref{rem:arch}}
\State $K_0 \gets \tau/(c_p L_{\mathrm{eff}})$ \Comment{target kernel of every MRePU input}
\State $s \gets \sqrt{K_0/q}$; first layer $W^{(1)}_{ij}\sim\mathcal N(0,1/d)$, $b^{(1)}=0$, input $s\,x$ \Comment{$K^{(1)}=K_0$}
\State $C_W \gets 1/g'(K_0)$ \eqref{eq:rule-par}, $1/\langle\sigma'(z)^2\rangle_{K_0}$ \eqref{eq:rule-perp}, $K_0/g(K_0)$ \eqref{eq:rule-fp}.
\Statex \hspace{\algorithmicindent}
\State hidden layers $\ell=2,\dots,L$: $W^{(\ell)}_{ij}\sim\mathcal N(0,C_W/n_{\ell-1})$, $b^{(\ell)}=0$
\State $K^{(L)} \gets$ $L-1$ iterations of $K\mapsto C_W\,g(K)$ starting from $K_0$
\State $C_{\mathrm{out}} \gets 1/g(K^{(L)})$; read-out $W^{(L+1)}_{ij}\sim\mathcal N(0,C_{\mathrm{out}}/n_L)$, $b^{(L+1)}=0$
\State \Return $s$, $C_W$, $C_{\mathrm{out}}$
\end{algorithmic}
\end{algorithm}

\begin{remark}[Residual and normalized architectures]\label{rem:arch}
For residual blocks $x\mapsto\gamma F(x)+\sqrt{1-\gamma^2}\,x$ the skip path preserves the kernel, and each MRePU layer on a branch contributes a drift reduced by $\gamma^2$;
we therefore use $L_{\mathrm{eff}}=\gamma^2\times$(number of MRePU layers on the branches).
If a normalization layer precedes an MRePU, its scale is initialized to $\sqrt{K_0}$ instead of $1$, so that the MRePU input again has kernel $K_0$.
\end{remark}

We validate the practical effectiveness of these initialization rules on real-world datasets MNIST in Section 5.3.

\subsection{Relation to the neural tangent kernel}\label{sec:ntk}
 
The analysis so far concerns the kernel $K^{(\ell)}$ of the forward pass.
Under gradient descent, the change of the network function is governed in the infinite-width limit by the neural tangent kernel (NTK) \citep{Jacot18},
and one may ask whether the NTK leads to the same conclusions.
In the effective theory, the NTK obeys a recursion driven by the same susceptibilities \citep[Ch.~8]{Roberts22}.
For a single input $x$ we define
\begin{equation}\label{eq:ntk-def}
  H^{(\ell)} = \mathbb{E}\biggl[\frac{1}{n_\ell}\sum_{i=1}^{n_\ell}\sum_{\ell'=1}^{\ell}\lambda^{(\ell')}\sum_{j,k}
  \Bigl(\frac{\partial z^{(\ell)}_i}{\partial W^{(\ell')}_{jk}}\Bigr)^{2}\biggr],
  \qquad \lambda^{(\ell')} = \frac{C_W^{(\ell')}}{n_{\ell'-1}},
\end{equation}
where $\lambda^{(\ell')}$ are the layer-wise learning rates for which the NTK has a finite infinite-width limit;
the biases are initialized at zero and not included.
At infinite width,
\begin{equation}\label{eq:ntk-rec}
  H^{(\ell+1)} = C_W\,g\bigl(K^{(\ell)}\bigr) + \chi_\perp\bigl(K^{(\ell)}\bigr)\,H^{(\ell)}
               = K^{(\ell+1)} + \chi_\perp\bigl(K^{(\ell)}\bigr)\,H^{(\ell)},
  \qquad H^{(1)} = K^{(1)},
\end{equation}
and therefore
\begin{equation}\label{eq:ntk-sum}
  H^{(L)} = \sum_{m=1}^{L} K^{(m)} \prod_{j=m}^{L-1}\chi_\perp\bigl(K^{(j)}\bigr).
\end{equation}
Each layer adds its kernel, and the contributions of the earlier layers are carried forward by the same products of $\chi_\perp$ that give the gradient factor in Proposition~10.
The NTK thus stays of order $LK^{(1)}$ exactly when the kernel and the gradient factor stay of order one, and it inherits the conclusions of Section~3 and Section~4.3:
\begin{itemize}
  \item For ReLU at $C_W=2$, $K^{(\ell)}=K^{(1)}$ and $\chi_\perp=1$, so $H^{(L)} = L\,K^{(1)}$.
  \item For RePU, $g(K)\propto K^p$, so at a fixed point $K_\star=C_Wg(K_\star)$ we have $\chi_\parallel=C_Wg'(K_\star)=p$,
        and by the ratio $\chi_\perp/\chi_\parallel=p/(2p-1)$ that follows from \eqref{eq:REPUsus}, $\chi_\perp = p^2/(2p-1) > 1$.
        Even a network whose kernel stayed exactly at $K_\star$ would therefore have $H^{(L)}/K_\star = \sum_{m=0}^{L-1}\bigl(p^2/(2p-1)\bigr)^m$, which grows exponentially with depth;
        and since $K_\star$ is unstable, the kernel does not stay there.
        No choice of $C_W$ avoids either effect.
  \item For MRePU, Proposition~10 extends to the NTK as follows.
\end{itemize}
 
\noindent
{\bf Corollary 11 (NTK at finite rescaled depth).}{ Under the assumptions of Proposition~10,
\begin{equation}\label{eq:ntk-limit}
  \frac{H^{(L)}}{L\,K_0} \to \eta_\kappa(\tau) := \frac{u_\kappa(\tau)^{r_p}}{\tau}\int_0^\tau u_\kappa(t)^{1-r_p}\,e^{\kappa(r_p-1)(\tau-t)}\,dt
  = 1 + \frac{1+r_p-2\kappa}{2}\,\tau + O(\tau^2).
\end{equation}
For $0\le\kappa<1$, $\eta_\kappa(\tau)\to\infty$ as $\tau\to\tau_c(\kappa)$; for $\kappa\ge1$, $\eta_\kappa(\tau)$ is finite for every $\tau$.
}
 
\noindent
{\bf Proof.}{ Let $\varepsilon=c_pK_0$ and $t_m=\varepsilon m$.
By Steps~1 and~4 of the proof of Proposition~10, $K^{(m)}/K_0 = u_\kappa(t_m)+O(\varepsilon)$ and
\[
  \prod_{j=m}^{L-1}\chi_\perp\bigl(K^{(j)}\bigr) = \exp\!\int_{t_m}^{\tau}\bigl(r_p\,u_\kappa(s)-\kappa\bigr)\,ds + O(\varepsilon)
  = e^{\kappa(r_p-1)(\tau-t_m)}\Bigl(\frac{u_\kappa(\tau)}{u_\kappa(t_m)}\Bigr)^{r_p} + O(\varepsilon)
\]
uniformly in $m$, where we used $\int_{t}^{\tau}u_\kappa\,ds = \kappa(\tau-t)+\log\bigl(u_\kappa(\tau)/u_\kappa(t)\bigr)$, which follows from \eqref{eq:u-ode}.
Dividing \eqref{eq:ntk-sum} by $LK_0 = \tau K_0/\varepsilon$ turns it into a Riemann sum with step $\varepsilon$, which converges to $\eta_\kappa(\tau)$.
For the expansion, $u_\kappa(t) = 1+(1-\kappa)t+O(t^2)$ and the exponent equals $(r_p-\kappa)(\tau-t)+O(\tau^2)$, so
$\eta_\kappa(\tau) = \tau^{-1}\int_0^\tau\bigl[1+(1-\kappa)t+(r_p-\kappa)(\tau-t)\bigr]dt+O(\tau^2) = 1+\tfrac{1}{2}(1+r_p-2\kappa)\tau+O(\tau^2)$.
For $\kappa<1$, $u_\kappa(\tau)^{r_p}\to\infty$ as $\tau\to\tau_c(\kappa)$ while the integral stays bounded below by its value on $[0,\tau_c/2]$, so $\eta_\kappa(\tau)\to\infty$;
for $\kappa\ge1$, $u_\kappa$ is bounded and positive on $[0,\infty)$ by item (ii) of Proposition~10, so $\eta_\kappa(\tau)$ is finite.
\hfill$\square$
}
 
\noindent
For $\tau\ll1$, the NTK of an MRePU network is therefore that of a critical ReLU network, $H^{(L)}\simeq LK_0$, up to a relative correction of order $\tau$;
for $p=2$ this correction is $-\tfrac{8}{9}\tau$ for rule~\eqref{eq:rule-par} and $+\tfrac{10}{9}\tau$ for $C_W=1$.
In this respect, too, MRePU at a small operating point behaves like ReLU and not like $\tanh$, whose kernel decays with depth so that its NTK grows more slowly than $L$ (Table~\ref{tab:ntk}).
 
\paragraph{Numerical check.}
Table~\ref{tab:ntk} compares \eqref{eq:ntk-rec} with the NTK of 32 networks of width $512$, measured for the preactivations $z^{(40)}$ with the normalization \eqref{eq:ntk-def}.
Wherever the networks stayed finite, the measured value agrees with the recursion within $1.4$ standard errors.
MRePU initialized with Algorithm~\ref{alg:recipe} has a predicted NTK within $1\%$ of the ReLU value, as \eqref{eq:ntk-limit} states, while the value for $\tanh$ is smaller by an order of magnitude.
None of the 32 RePU networks reached layer $40$ with a finite, nonzero NTK: $13$ exploded and $19$ vanished, as expected from the instability of the fixed point.
The standard errors are about $10\%$ of the mean for ReLU and MRePU, because the NTK of a deep network at finite width varies between networks by an amount that grows with $L/n$ \citep{HaninNica20}.
At this precision the experiment does not resolve finite-width corrections to the mean;
likewise, the measured difference between the two MRePU cases ($15\%$) has the predicted size ($13\%$) but lies within the statistical error.
 
\begin{table}[htp]
\centering
\caption{NTK $H^{(40)}$ at initialization in units of $K^{(1)}$, for networks of width $512$ (32 networks, one input; mean $\pm$ standard error).
Theory: recursion \eqref{eq:ntk-rec}; for MRePU, $g(K)$ and $\langle\sigma'(z)^2\rangle_K$ are reduced to the integrals $I_n$ as in Proposition~8.
Last column: number of networks whose NTK became non-finite (exploded) or underflowed to zero (vanished) before layer $40$. 
$^{a}$Value if the kernel stayed exactly at the (unstable) fixed point.
$^{b}$No network had a finite, nonzero NTK at layer $40$.}
\label{tab:ntk}
\small
\begin{tabular}{lccccc}
\toprule
Network & $C_W$ & $\tau$ & Theory & Measured & Exploded / vanished \\
\midrule
ReLU                                  & $2$                  & --     & $40.0$                & $41.5\pm4.1$  & $0\,/\,0$ \\
$\tanh$                               & $1$                  & --     & $4.68$                & $4.76\pm0.21$ & $0\,/\,0$ \\
RePU ($p{=}2$)                        & $K_\star/g(K_\star)$ & --     & $2.98\times10^{5\,a}$ & --$^{b}$      & $13\,/\,19$ \\
MRePU ($p{=}2$), \eqref{eq:rule-par}  & $1/g'(K_0)$          & $0.01$ & $39.65$               & $44.9\pm3.8$  & $0\,/\,0$ \\
MRePU ($p{=}2$)                       & $1$                  & $0.1$  & $44.83$               & $51.7\pm5.0$  & $0\,/\,0$ \\
\bottomrule
\end{tabular}
\end{table}
 
\noindent
The NTK therefore leads to the same conclusion as the forward kernel: whether the NTK of a deep network stays of order $LK^{(1)}$ is decided by $\chi_\perp$ along the kernel trajectory,
so RePU fails for every $C_W$ and MRePU behaves like ReLU when $\tau\ll1$.
This statement concerns the NTK at initialization and at infinite width. it is one of the reasons why the largest $\tau$ at which training succeeds is smaller than the initialization analysis suggests and decreases with depth (Section~5.3.1, Figure~\ref{fig:t3-paper}).

\section{Approximation Properties of MRePU networks}
In this section, we provide a theoretical justification that neural networks equipped with the MRePU activation can reconstruct polynomials, in close analogy to RePU networks, and consequently inherit the same universal approximation property for the class of differentiable functions established in \cite{Shen23}. Beyond this theoretical guarantee, we empirically demonstrate that MRePU networks exhibit a suitable inductive bias when approximating certain function classes. Furthermore, we show through experiments on standard benchmarks such as MNIST and CIFAR-10 that, under the proposed criticality condition, training remains stable.

\subsection{Approximation Theorems for MRePU}
In (\cite{Shen23}), it is shown that a RePU network of $p$-th order can accurately represent an expanded univariate polynomial using Horner’s method, utilizing $4p$ nodes for multiplication and $2p$ nodes to construct a $p$-th order polynomial. Furthermore, by applying these univariate polynomial results inductively, the representation of multivariate polynomials can also be achieved. Therefore, for the MRePU of order $p$ network, if we can demonstrate that it can construct any $p$-th order polynomial using $\mathcal{O}(p)$ nodes, as in the case of RePU, we obtain the same result. The following lemma formalizes this.

\noindent
{\bf Lemma 12.}{ A polynomial of degree $p$, $P(x)=\sum_{k=0}^{p}a_{k}x^{k}$ can be represented by a $p$-th order MRePU network with one hidden layer and at most $4(p+1)$ hidden nodes.
}
\\
{\bf Proof.}{ 
The proof proceeds in three steps.

\medskip
\noindent\textbf{Step 1: A 4-unit exact block for $(x-c)^p$ on $\mathbb{R}$.}
Fix any $c\in\mathbb{R}$ and write $t:=x-c$. Consider the four units
\[
\sigma_{m;p}(t-1),\qquad \sigma_{m;p}(2t-1),\qquad \sigma_{m;p}(-t-1),\qquad \sigma_{m;p}(-2t-1).
\]
We first compute them on the two half-lines.

\smallskip
\noindent\emph{Case 1: $t\ge 0$.}
Then $t-1\ge -1$ and $2t-1\ge -1$, while $-t-1<-1$ and $-2t-1<-1$.
Hence
\[
\sigma_{m;p}(t-1)=(t-1)t^p=t^{p+1}-t^p,\qquad
\sigma_{m;p}(2t-1)=(2t-1)(2t)^p=2^{p+1}t^{p+1}-2^pt^p,
\]
and $\sigma_{m;p}(-t-1)=\sigma_{m;p}(-2t-1)=0$.
Choose
\[
A:=-2,\qquad B:=2^{-p}.
\]
Then
\begin{align*}
A\sigma_{m;p}(t-1)+B\sigma_{m;p}(2t-1)
&=A(t^{p+1}-t^p)+B(2^{p+1}t^{p+1}-2^pt^p)\\
&=(A+B2^{p+1})t^{p+1}+(-A-B2^p)t^p\\
&=\bigl(-2+2^{-p}2^{p+1}\bigr)t^{p+1}+\bigl(2-2^{-p}2^p\bigr)t^p\\
&=0\cdot t^{p+1}+1\cdot t^p\\
&=t^p.
\end{align*}

\smallskip
\noindent\emph{Case 2: $t\le 0$.}
Then $t-1<-1$ and $2t-1<-1$, while $-t-1\ge -1$ and $-2t-1\ge -1$.
Thus $\sigma_{m;p}(t-1)=\sigma_{m;p}(2t-1)=0$ and
\[
\sigma_{m;p}(-t-1)=(-t-1)(-t)^p=(-t)^{p+1}-(-t)^p,
\]
\[
\sigma_{m;p}(-2t-1)=(-2t-1)(-2t)^p
=2^{p+1}(-t)^{p+1}-2^p(-t)^p.
\]
With the same $A,B$ as above,
\begin{align*}
A\sigma_{m;p}(-t-1)+B\sigma_{m;p}(-2t-1)
&=A\bigl((-t)^{p+1}-(-t)^p\bigr)+B\bigl(2^{p+1}(-t)^{p+1}-2^p(-t)^p\bigr)\\
&=(A+B2^{p+1})(-t)^{p+1}+(-A-B2^p)(-t)^p\\
&=0\cdot (-t)^{p+1}+1\cdot (-t)^p\\
&=(-t)^p.
\end{align*}
Since $(-t)^p=(-1)^p t^p$, multiplying by $(-1)^p$ yields
\[
(-1)^p\bigl(A\sigma_{m;p}(-t-1)+B\sigma_{m;p}(-2t-1)\bigr)=t^p.
\]

\smallskip
\noindent Combining the two cases, for every $t\in\mathbb{R}$ we have the identity
\begin{equation}\label{eq:block-tp}
t^p=
A\sigma_{m;p}(t-1)+B\sigma_{m;p}(2t-1)
+(-1)^p\Bigl(A\sigma_{m;p}(-t-1)+B\sigma_{m;p}(-2t-1)\Bigr),
\end{equation}
with $A=-2$ and $B=2^{-p}$. Substituting $t=x-c$ gives an exact 4-unit
one-hidden-layer $\sigma_{m;p}$-network representation of $(x-c)^p$ on $\mathbb{R}$.

\medskip
\noindent\textbf{Step 2: Spanning order $p$ polynomials by shifted $p$-th powers.}
Choose distinct real numbers $c_0,c_1,\dots,c_p$ (e.g.\ $c_i=i$).
For each $i$,
\[
(x-c_i)^p=\sum_{j=0}^{p}\binom{p}{j}x^j(-c_i)^{p-j}.
\]
We claim that for any coefficients $(a_0,\dots,a_p)$ there exist
$(\alpha_0,\dots,\alpha_p)$ such that
\begin{equation}\label{eq:span}
P(x)=\sum_{i=0}^{p}\alpha_i(x-c_i)^p.
\end{equation}
Indeed, comparing coefficients of $x^j$ in \eqref{eq:span} gives the linear system
\[
\sum_{i=0}^{p}\alpha_i(-c_i)^{p-j}=\frac{a_j}{\binom{p}{j}},\qquad j=0,1,\dots,p.
\]
The matrix $V=[(-c_i)^{p-j}]_{j,i}$ is a Vandermonde matrix up to a reversal
of rows, and since the $c_i$ are distinct it is invertible. Hence the system
admits a (unique) solution $(\alpha_0,\dots,\alpha_p)$, proving \eqref{eq:span}.

\medskip
\noindent\textbf{Step 3: Assemble the network for $P$.}
For each $i\in\{0,\dots,p\}$, represent $(x-c_i)^p$ using the 4-unit identity
\eqref{eq:block-tp} (with $t=x-c_i$), and multiply the four corresponding output
weights by $\alpha_i$. Summing these $(p+1)$ blocks yields a single one-hidden-layer
$\sigma_{m;p}$ network with $4(p+1)$ hidden units whose output equals
$\sum_{i=0}^{p}\alpha_i(x-c_i)^p=P(x)$ for all $x\in\mathbb{R}$.

Therefore any polynomial of degree at most $p$ is represented exactly on
$\mathbb{R}$ by a one-hidden-layer $p$-MRePU network with at most $4(p+1)$ units.
\hfill$\square$
}
\\
Then, by replacing Lemma 40, which is used in the proof of Theorem 3 in (\cite{Shen23}), we obtain the following theorem:

\noindent
{\bf Theorem 13.}{ If $f:\mathbb{R}^{d}\rightarrow \mathbb{R}$ is a polynomial of $d$ variables with degree $N$, then $f$ can be reconstructed via MRePU neural network with depth $\mathcal{D}$, width $\mathcal{W}$, number of neurons $\mathcal{U}$ and total number of parameters $\mathcal{P}$ with following specification:
\begin{equation*}
    \begin{split}
        &\mathcal{D}=2N-1,\quad\mathcal{W}=24(p+1)N^{d-1}+12(p+1)(N^{d-1}-N)/(N-1)\simeq\mathcal{O}(pN^{d-1}), \\
        &\mathcal{U}=(12p+14)(2N^{d}-N^{d-1}-N)+(4p+4)(2N^{d}-N^{d-1}-N)/(N-1)\simeq \mathcal{O}(pN^{d}),\\
        &\mathcal{P}=(60p+62)(2N^{d}-N^{d-1}-N)+(4p+5)(2N^{d}-N^{d-1}-N)/(N-1)\simeq\mathcal{O}(pN^{d}).
    \end{split}
\end{equation*}
}
\\
Additionally, as a corollary to the above theorem, we obtain a universal approximation result for the class of differentiable functions, which corresponds to Theorem~5 in (\cite{Shen23}). Before stating this result, we define a reference norm that will serve as our standard.

\noindent
{\bf Definition 14.}{ We define the norm of $f\in C^{s}(A)$, where $A\subset \mathbb{R}^{d}$, as following:
\begin{equation*}
    \|f\|_{C^{s}}:=\sum_{|\alpha|\leq s}\sup_{x \in A}|D^{\alpha}f(x)|,
\end{equation*}
where $\alpha$ is an index vector for partial derivative and $|\alpha|$ is total order of derivative.
}

\noindent
{\bf Corollary 15.}{ Let $f$ be a function belongs to $C^{s}(K)$ where $K\subset \mathbb{R}^{d}$ is a compact set. For any $N\in\mathbb{N}^{+}$, there exists a MRePU neural network $\psi_{N}$ with depth $\mathcal{D}$, width $\mathcal{W}$, number of neurons $\mathcal{U}$ and total number of parameters $\mathcal{P}$ with following specification:
\begin{equation*}
    \begin{split}
        &\mathcal{D}=2N-1,\quad\mathcal{W}=24(p+1)N^{d-1}+12(p+1)(N^{d-1}-N)/(N-1)\simeq\mathcal{O}(pN^{d-1}), \\
        &\mathcal{U}=(12p+14)(2N^{d}-N^{d-1}-N)+(4p+4)(2N^{d}-N^{d-1}-N)/(N-1)\simeq \mathcal{O}(pN^{d}),\\
        &\mathcal{P}=(60p+62)(2N^{d}-N^{d-1}-N)+(4p+5)(2N^{d}-N^{d-1}-N)/(N-1)\simeq\mathcal{O}(pN^{d}).
    \end{split}
\end{equation*}
where $C_{p,s,d,K}$ is a positive constant which depends only on $p, d, s$ and the diameter of $K$.
}

\subsection{Experiments on Synthetic Data}
In this section we compare MRePU networks with ReLU and GELU networks on polynomial and piecewise-defined targets, measuring the training loss, the error on a test grid and the error of the first derivative. RePU is excluded from the comparison group as stacking layers with it is entirely infeasible.

\noindent
{\bf Regression on polynomials and differentiable functions.}{
First, we validate the efficiency of MRePU on polynomial functions. The neural network architecture consists of five hidden layers, each with a width of 32, and the input dimension is set according to the variables in each polynomial. For univariate polynomial problems, the training dataset was constructed with inputs from -2.0 to 2.0, divided into 101 points at intervals of 0.04. For $x^{3} - x$, the learning rate was set to 0.005, and $C_{W}$ was set to 0.5 for MRePU and 1.0 for other cases. For $x^{5} - 5x^{3} + 6x$, the learning rate was set to 0.0001, with $C_{W}$ set to 0.8 for MRePU case and learning rate to 0.001, $C_{W}$ to 1.0 for the others. For multivariate cases, the dataset for $xy$ was created by forming a meshgrid over the range $[-2,2] \times [-2,2]$, with each axis divided into 101 points. Similarly, for $xyz$, a meshgrid was formed over the range $[-1,1] \times [-1,1] \times [-1,1]$, with each axis divided into 51 points. Training was conducted up to 10,000 epochs for all cases. The learning rate for multivariate cases was set to 0.001, and $C_{W}$ was set to 0.55 for  $xy$, 0.6 for $xyz$, and 1.0 for the other cases. Each hyperparameter was empirically determined to achieve optimal performance.
\begin{table}[htp]
\centering
\small
\begin{tabular}{l c c c c}
\toprule
 & $x^{3}-x$ & $x^{5}-5x^{3}+6x$ & $xy$ & $xyz$  \\
\midrule
MRePU &  $\mathbf{4.0\times10^{-4}}$ &  $\mathbf{2.0\times10^{-4}}$ & $\mathbf{2.7\times10^{-5}}$ &   $\mathbf{2.5\times10^{-4}}$  \\
ReLU  &  $8.0\times10^{-4}$ & $7.0\times10^{-4}$ &  $4.0\times10^{-4}$ & $4.2\times10^{-2}$ \\
GELU  & $7.2\times10^{-2}$ & $2.1\times10^{-3}$ &  $4.0\times10^{-4}$ & $4.2\times10^{-2}$ \\
\bottomrule
\end{tabular}
\caption{Mean final training losses over 25 trials for polynomial approximation for each of the MRePU (order $p = 2$), ReLU, and GELU networks. The values in bold represent the lowest values in each column.} \label{t:polyreg}
\end{table}
Table \ref{t:polyreg} presents the average final loss over 25 experiments, where neural networks were trained to regress each polynomial function. As observed in the table, MRePU-based neural networks consistently outperform the others in all cases. Notably, for approximating $xyz$, other networks fail to learn effectively, while only the MRePU network achieves meaningful learning. The loss plots for $x^{5} - 5x^{3} + 6x$ and $xy$ are depicted in Fig.~\ref{Evolution of loss 5 order}. Fig.~\ref{5 order reg} illustrates the regression results of $x^{5} - 5x^{3} + 6x$ using ReLU and MRePU. A key observation is that the trained MRePU network approximates not only the function itself but also its derivative, whereas the ReLU network produces a piecewise function with noticeable discontinuities. Similarly, Fig.~\ref{xy reg} shows the contour graphs of trained networks with different activation functions for $xy$. The MRePU network demonstrates a highly accurate approximation, while the ReLU network exhibits significant instability in the contours, and GELU shows reduced accuracy.

\begin{figure}[htp]
\centering
\includegraphics[width=0.49\textwidth]{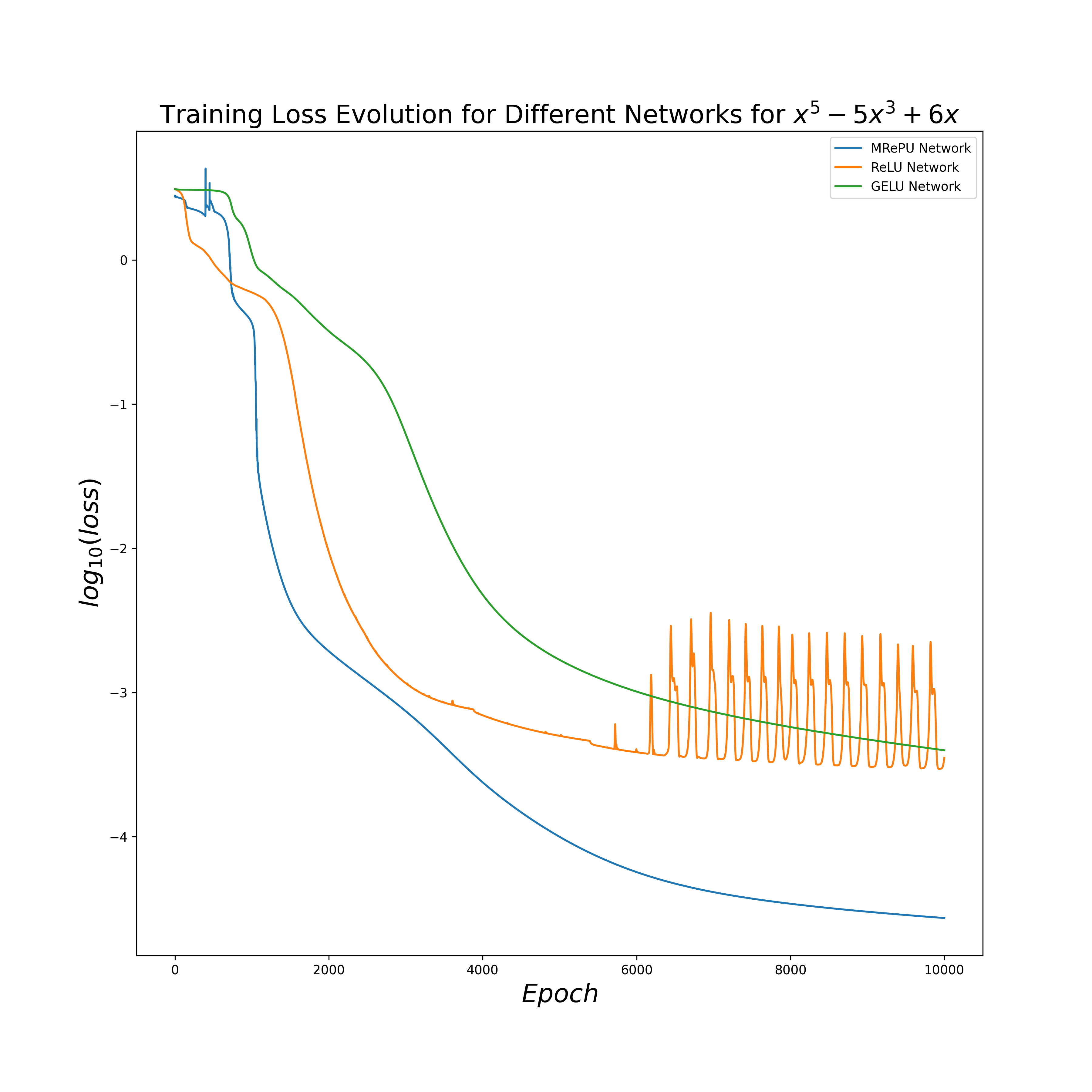}
\includegraphics[width=0.49\textwidth]{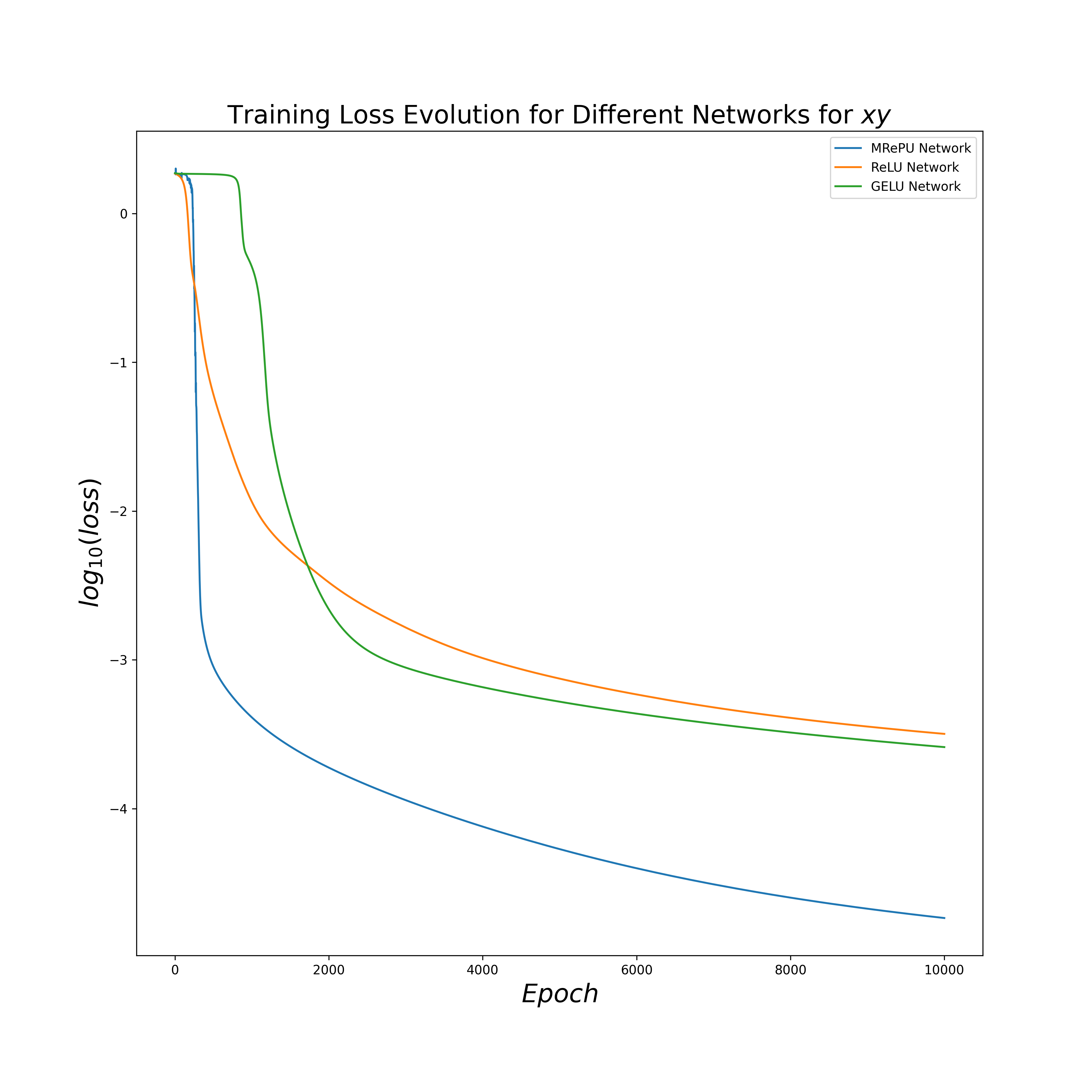}
\caption{The training loss plots for each neural network over epochs for the cases of $x^{5}-5x^{3}+6x$ (\textbf{Left}) and $xy$ (\textbf{Right}).}\label{Evolution of loss 5 order}
\end{figure}

\begin{figure}[htp]
    \begin{center}
        \includegraphics[width=0.49\textwidth]{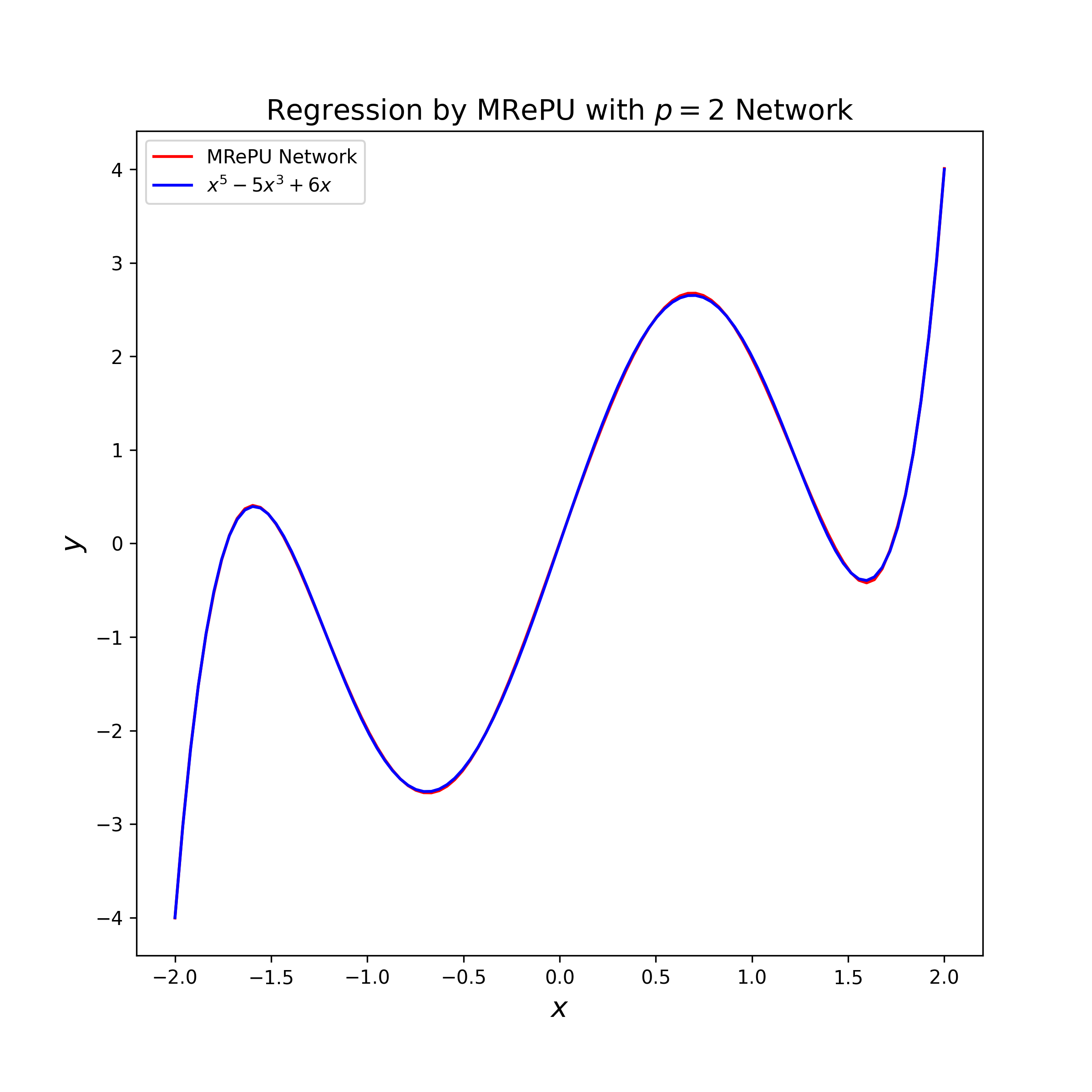}
        \includegraphics[width=0.49\textwidth]{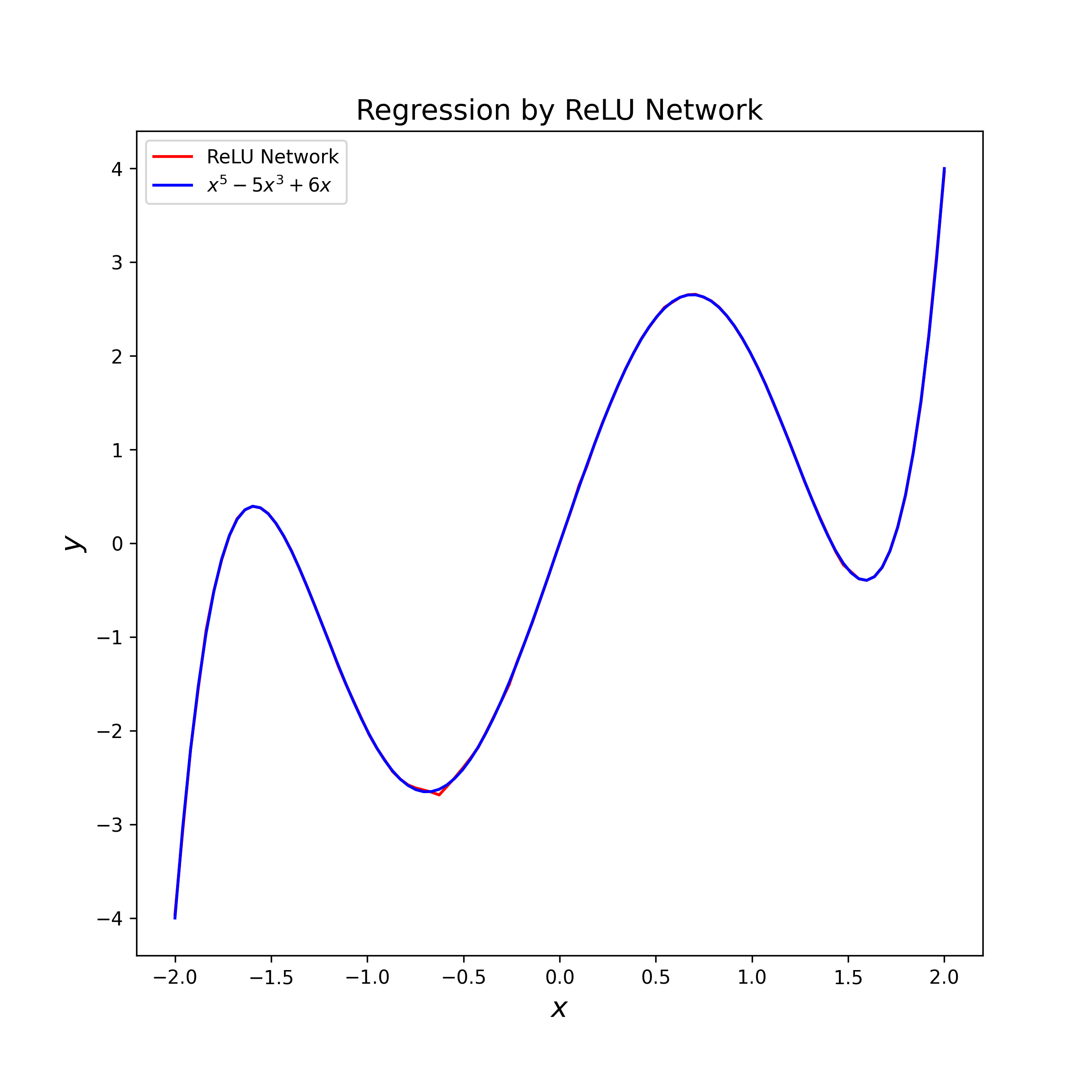}
        \\[\smallskipamount]
        \includegraphics[width=0.49\textwidth]{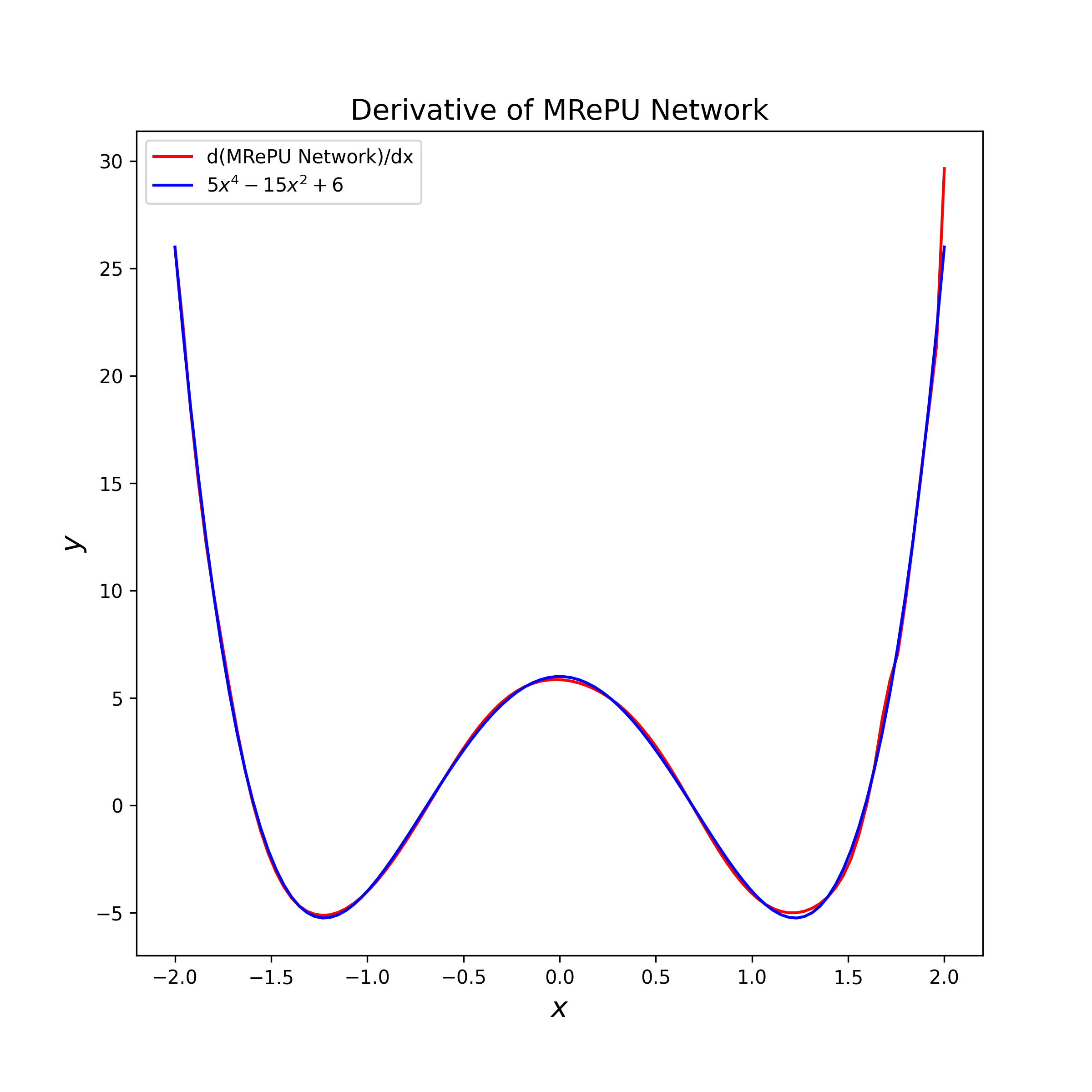}
        \includegraphics[width=0.49\textwidth]{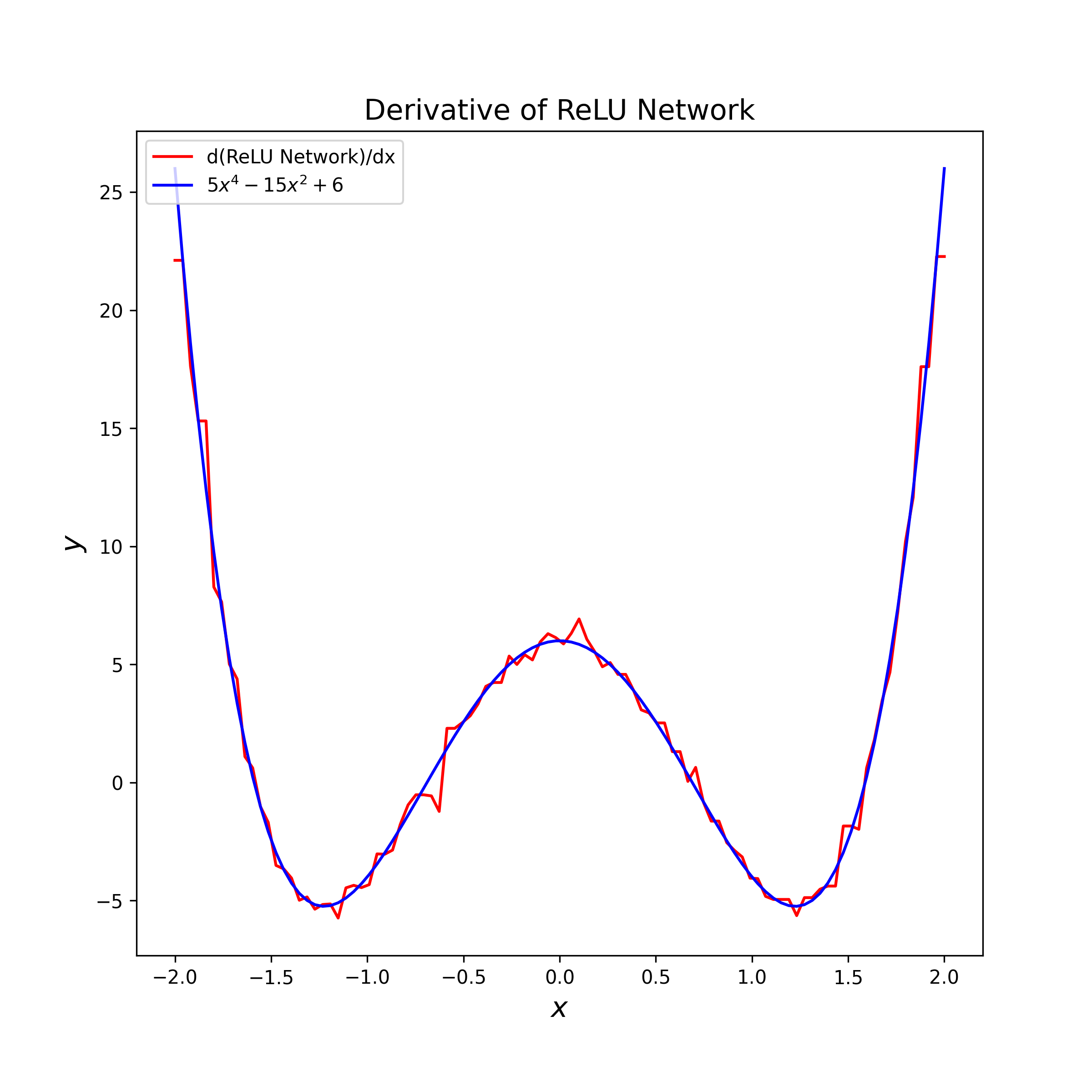}

    \end{center}
    \caption{\textbf{Top}: The results of training for regressing $x^{5}-5x^{3}+6x$, where the left plot shows the output of the MRePU with $p=2$ network, and the right plot shows the output of the ReLU network.
\textbf{Bottom}: The comparison between the derivative of the trained networks and the actual derivative function, $5x^{4}-15x^{2}+6$.
     }
     
   \label{5 order reg}
   
\end{figure}

\begin{figure}[htp]
    \begin{center}
        \includegraphics[width=0.49\textwidth]{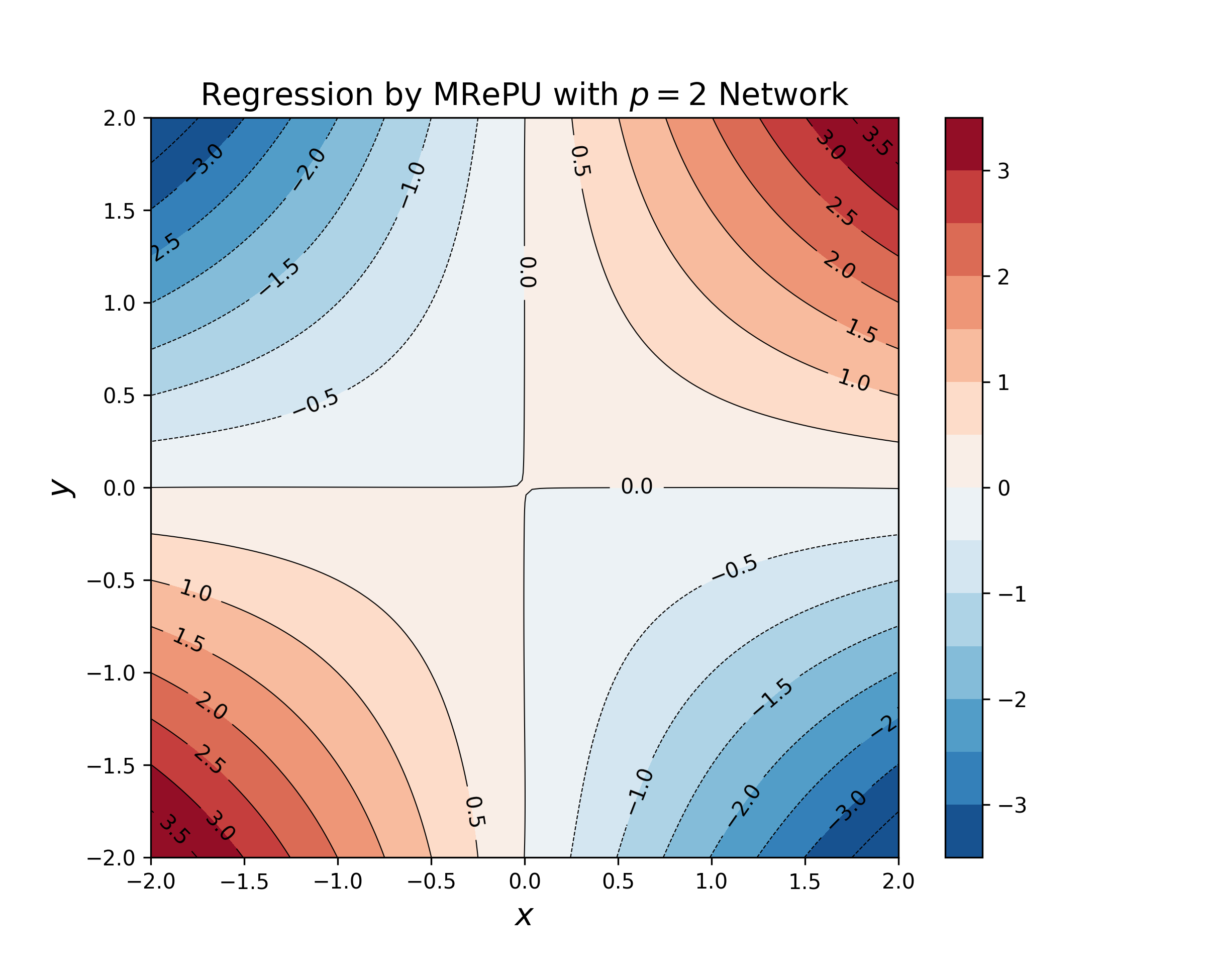}
        \includegraphics[width=0.49\textwidth]{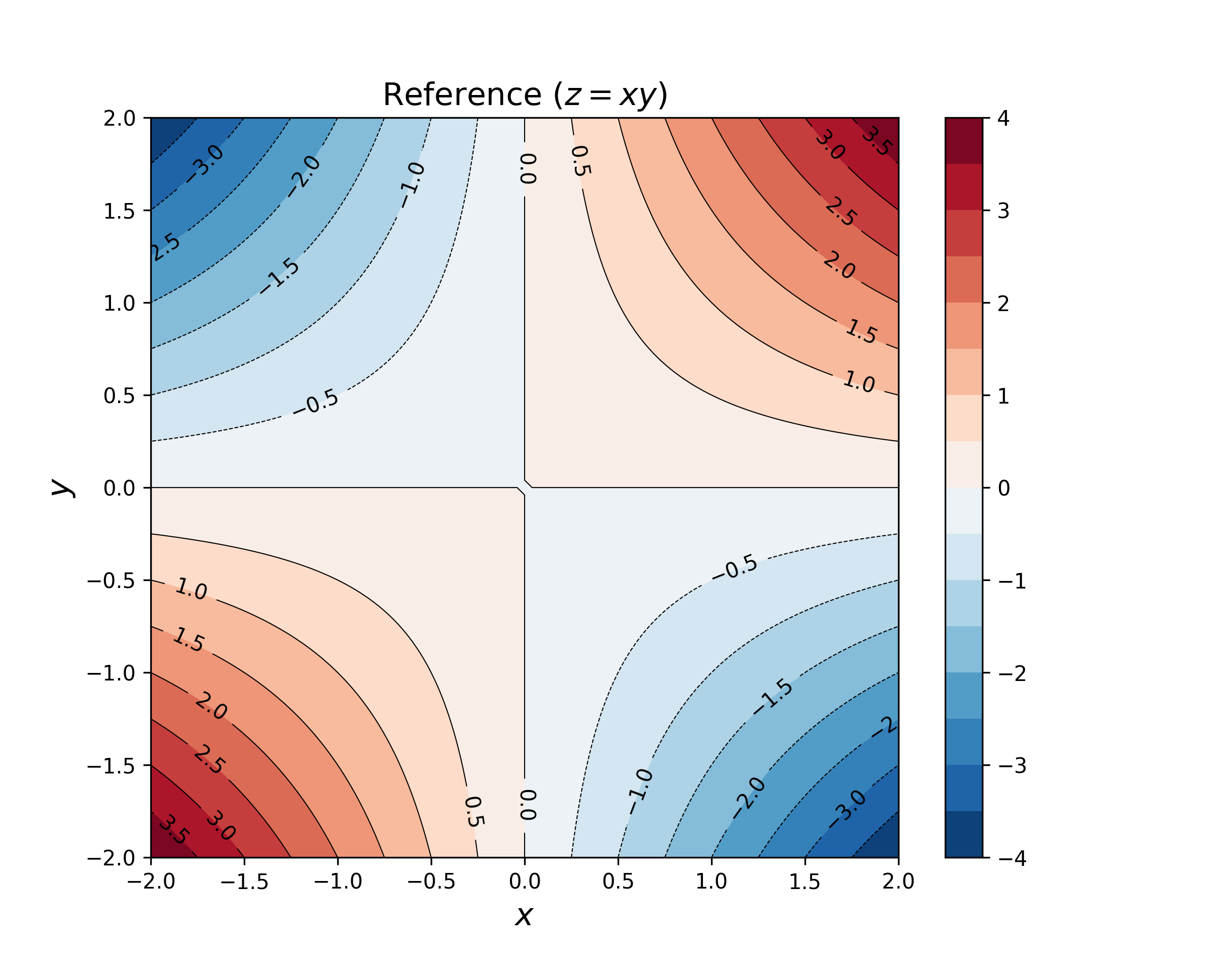}
        \\[\smallskipamount]
        \includegraphics[width=0.49\textwidth]{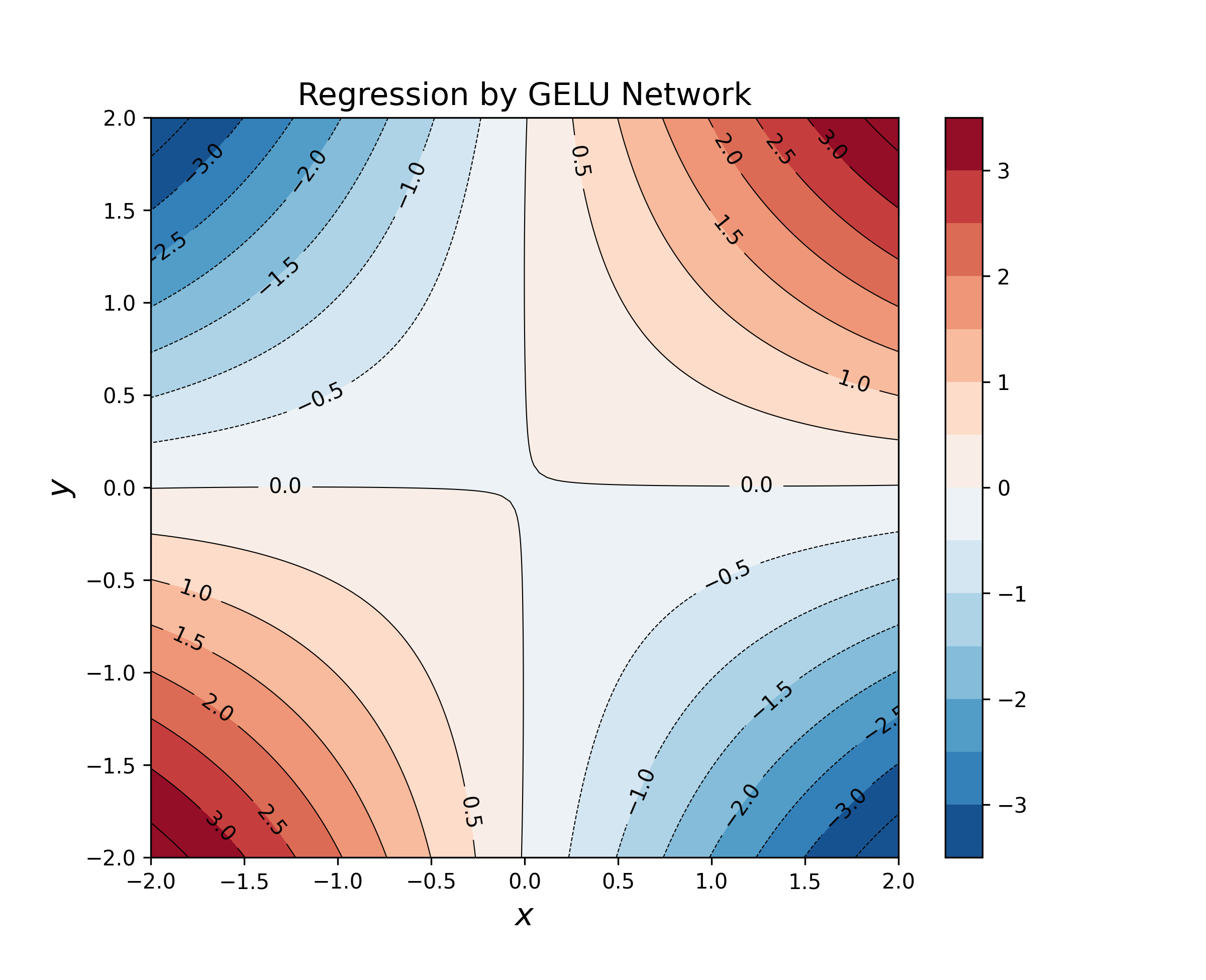}
        \includegraphics[width=0.49\textwidth]{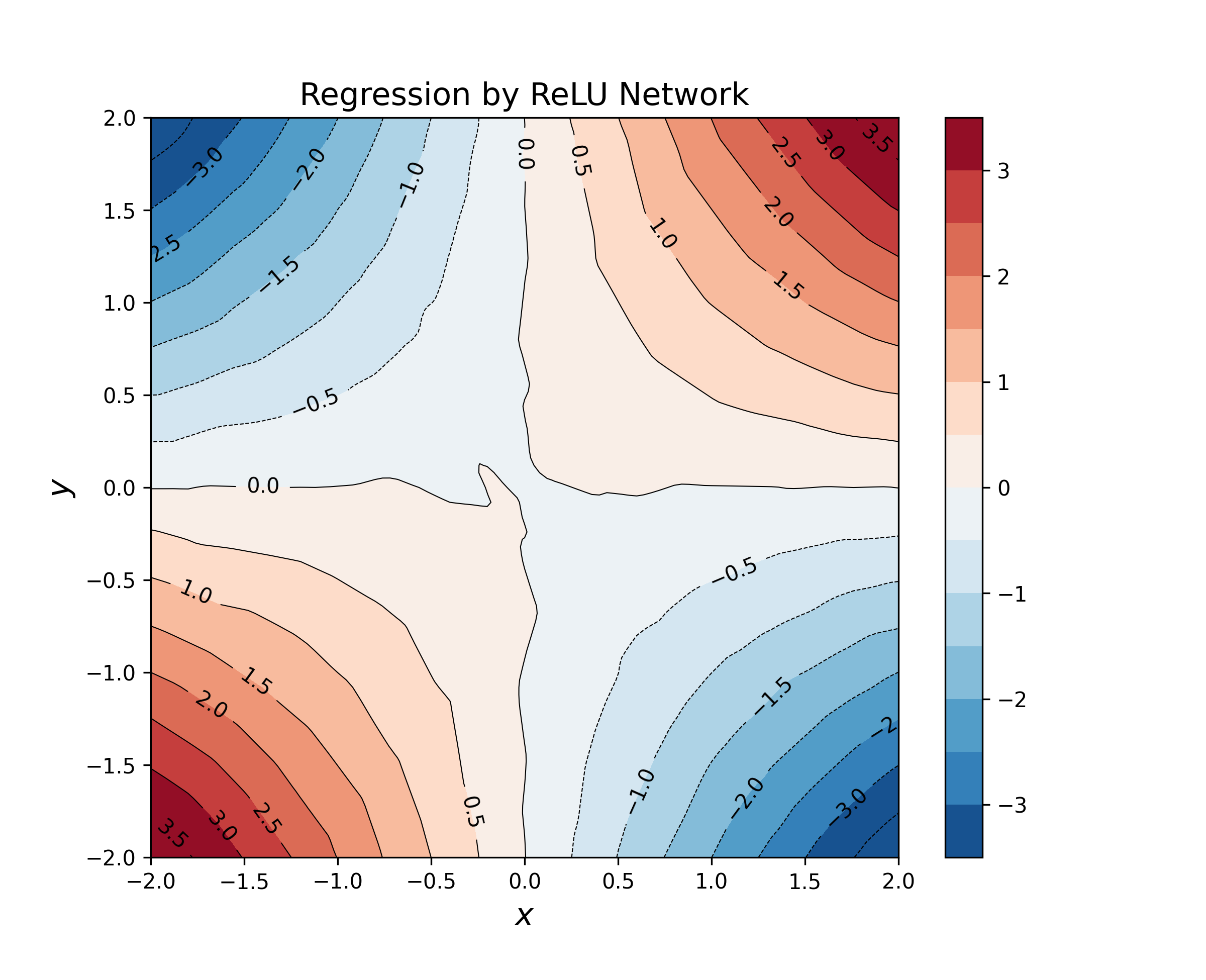}

    \end{center}
    \caption{Contour plots for the reference function and the trained neural networks. \textbf{Top left}: MRePU with $p=2$ network. \textbf{Top right}: Reference ($xy$). \textbf{Bottom left}: GELU network. \textbf{Bottom right}: ReLU network.
     }
     
   \label{xy reg}
   
\end{figure}

Next, we conducted experiments on differentiable functions. The neural network architecture remained identical to that used for polynomial regression. The experiments involved regression on $\lfloor x \rfloor_{+}^{3}$ and the functions $f(x)$ and $g(x,y)$, defined as follows. 
\begin{equation*}
    f(x) = \begin{cases} 
		x^{4}-3x^{2}+1 & \text{if }x\leq -1 \\ 
         -x^{2} & \text{if }-1\leq x\leq 0 \\
         x^{3} & \text{if } 0\leq x \leq 1 \\
         3x-2 & \text{if } 1 \leq x
     \end{cases}
\end{equation*}
\begin{equation*}
\begin{split}
    g(x,y) &=\begin{cases} 
		10(r^{2}-\frac{1}{2})^{2} & \text{if } r\leq \frac{1}{\sqrt{2}} \\ 
         (r^{2}-\frac{1}{2})^{3} & \text{otherwise} 
     \end{cases}  \\
     r&=\sqrt{x^{2}+y^{2}}
\end{split}
\end{equation*}
The training dataset for univariate functions was constructed in the same manner as for the polynomials. For multivariate functions, a meshgrid was generated by dividing the domain $[-1, 1] \times [-1, 1]$ into 51 intervals along each axis. The learning rate and $C_{W}$ values for these experiments were as follows: For $\lfloor x \rfloor_{+}^{3}$, the MRePU network used a learning rate of 0.0004 and $C_{W} = 0.6$, while the other networks used 0.001 and $C_{W} = 1.0$. For $f(x)$, The MRePU network employed a learning rate of 0.001 and $C_{W} = 0.6$, while the other networks used 0.001 and $C_{W} = 1.0$. For $g(x,y)$, the MRePU network used a learning rate of 0.002 and $C_{W} = 0.8$, whereas the remaining networks used 0.003 and $C_{W} = 1.0$. Each hyperparameter was empirically determined to achieve optimal performance.

\begin{table}[htp]
\centering
\small
\begin{tabular}{l c c c}
\toprule
 & $\lfloor x\rfloor_{+}^{3}$ & $f(x)$ & $g(x,y)$ \\
\midrule
MRePU & $8.1\times10^{-5}$ & $7.9\times10^{-5}$ & $4.0\times10^{-4}$ \\
ReLU  & $\mathbf{3.2\times10^{-5}}$ & $\mathbf{5.6\times10^{-5}}$ & $\mathbf{2.0\times10^{-4}}$ \\
GELU  & $2.4\times10^{-3}$ & $8.0\times10^{-4}$ & $4.4\times10^{-1}$ \\
\bottomrule
\end{tabular}
\caption{Mean final training losses over 25 trials for the approximation of differentiable functions for each of the MRePU (order $p = 2$), ReLU, and GELU networks. The values in bold represent the lowest values in each column.} \label{t:diffreg}
\end{table}

As shown in Table \ref{t:diffreg} and Fig.~\ref{Evolution of loss f and g}, for differentiable functions with discontinuities in higher-order derivatives, the ReLU network generally demonstrates the best performance. However, as observed in Figs.~\ref{f reg} and \ref{g reg}, even though the training loss of the ReLU network is lower (i.e., its $L^{2}$ norm is smaller), the MRePU network exhibits more stable function approximation in terms of contour smoothness and the behavior of the differentiated function. For the GELU network, it is evident that learning fails entirely for $g(x, y)$.

\begin{figure}[htp]
\centering
\includegraphics[width=0.49\textwidth]{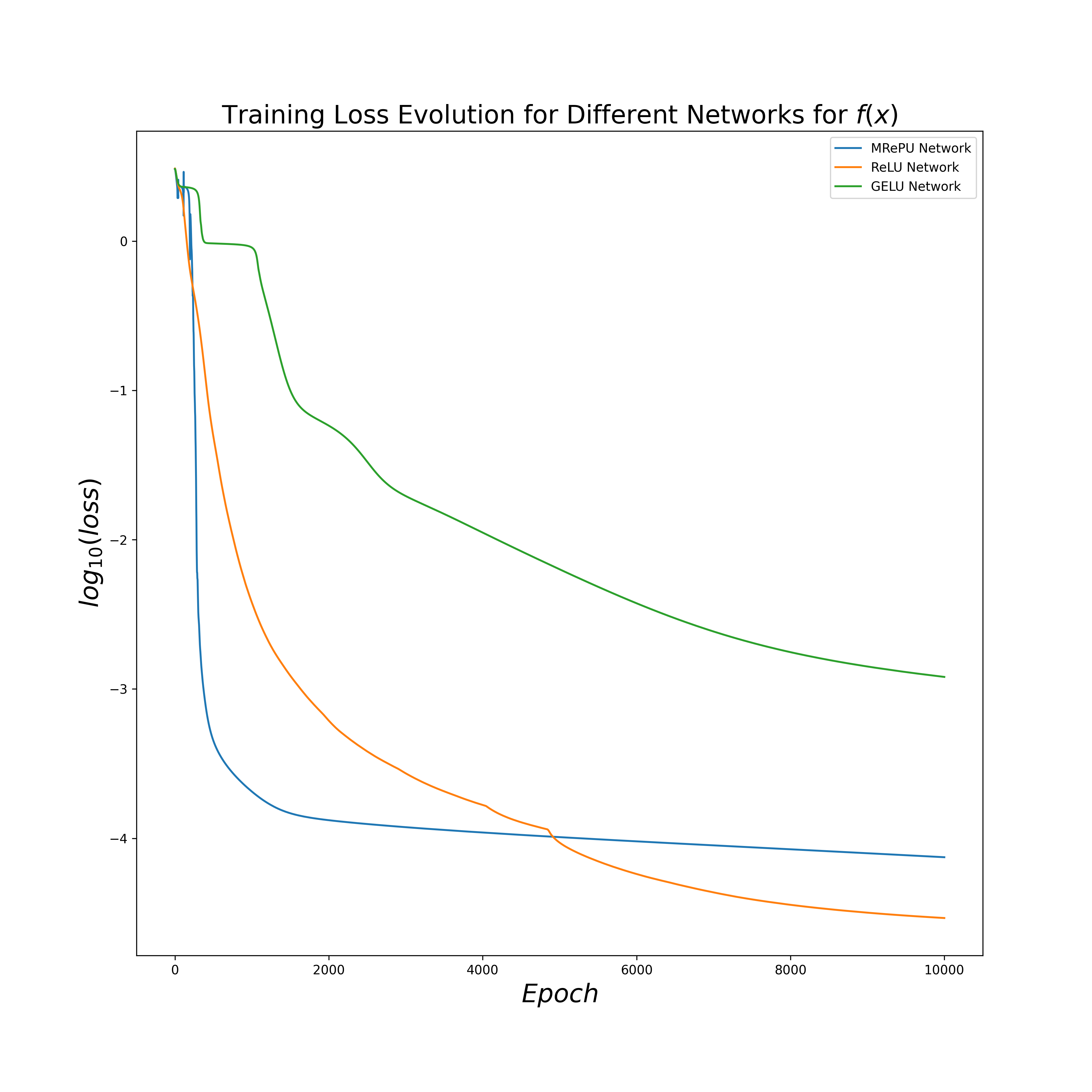}
\includegraphics[width=0.49\textwidth]{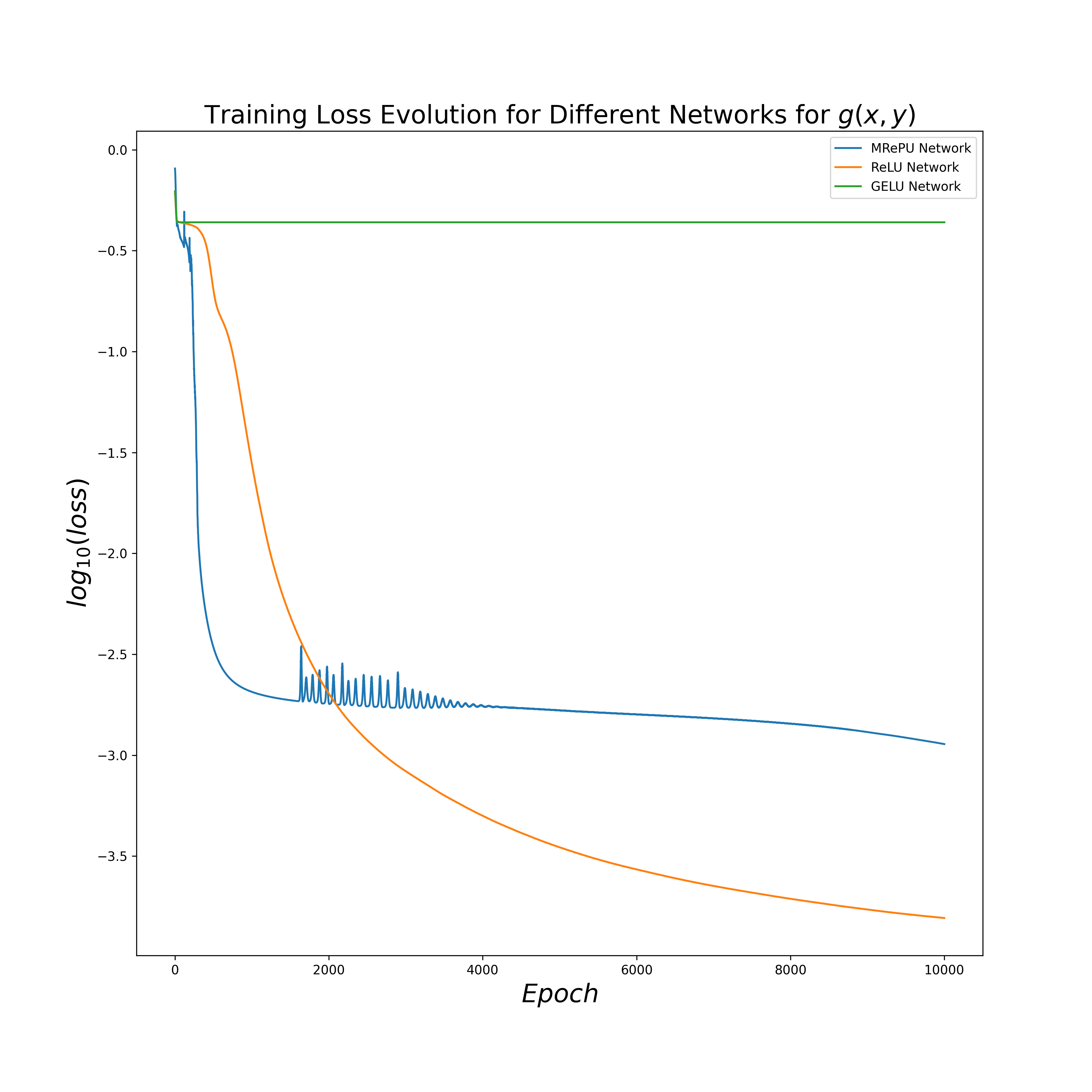}
\caption{The training loss plots for each neural network over epochs for the cases of $f(x)$ (\textbf{Left}) and $g(x,y)$ (\textbf{Right}).}\label{Evolution of loss f and g}
\end{figure}

\begin{figure}[htp]
    \begin{center}
        \includegraphics[width=0.49\textwidth]{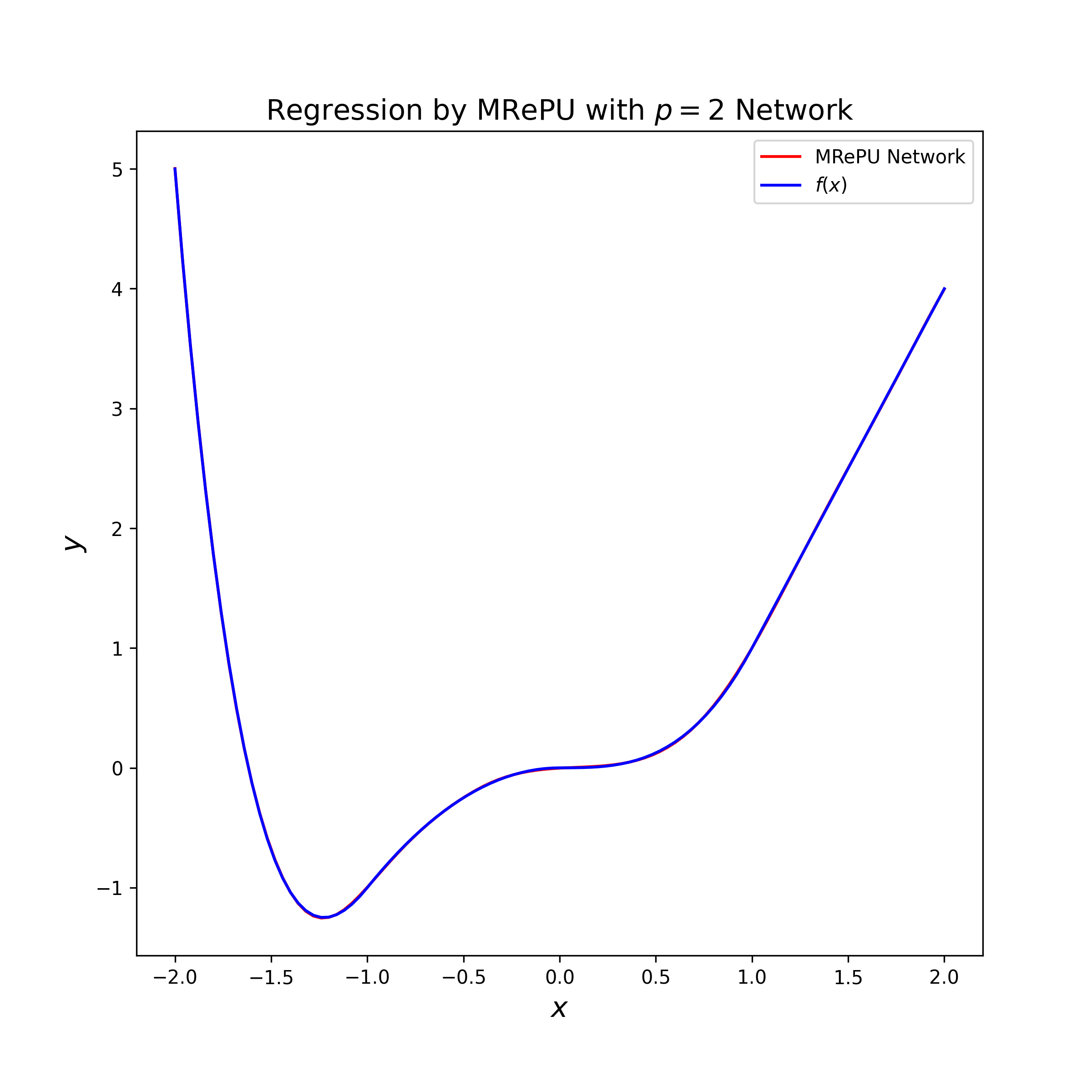}
        \includegraphics[width=0.49\textwidth]{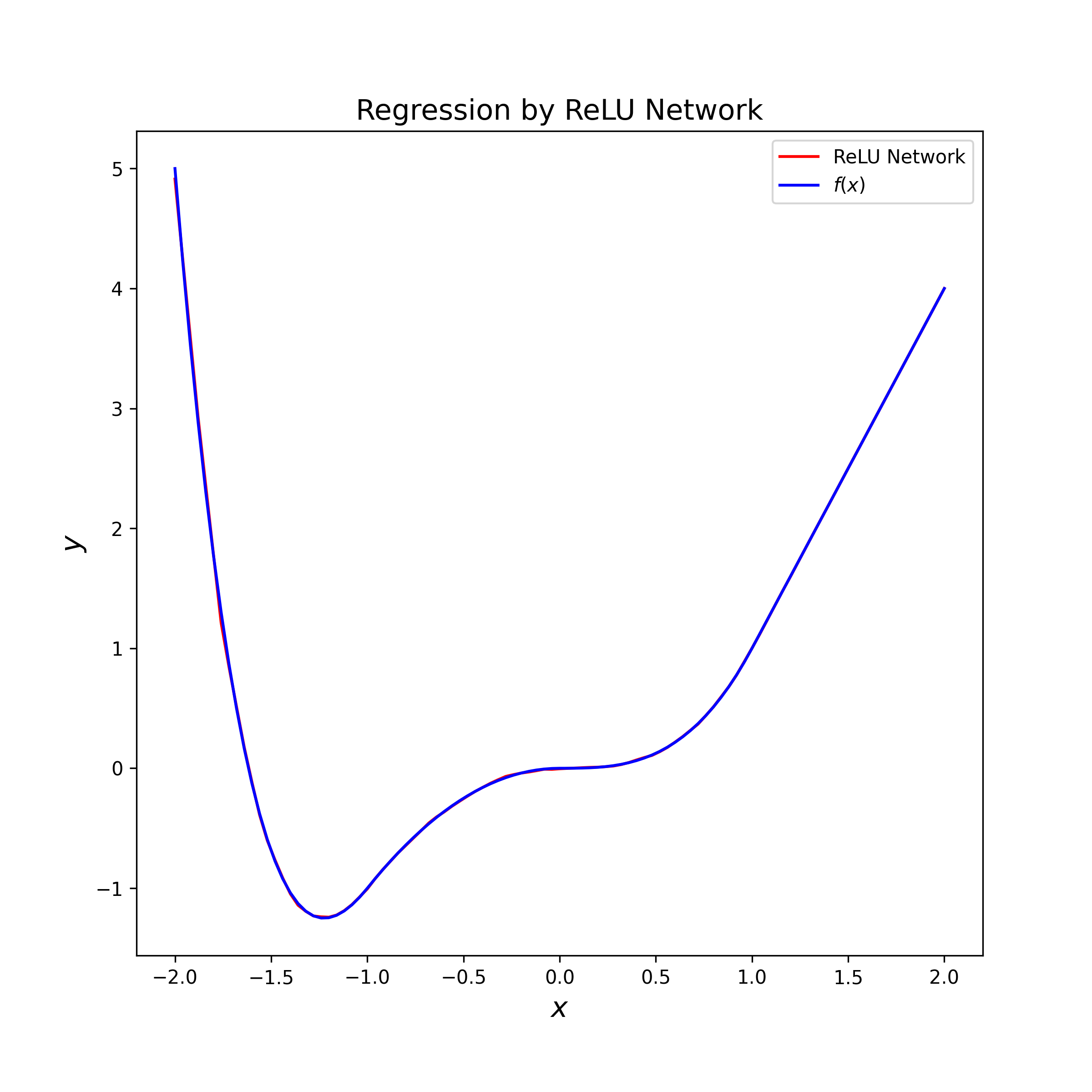}
        \\[\smallskipamount]
        \includegraphics[width=0.49\textwidth]{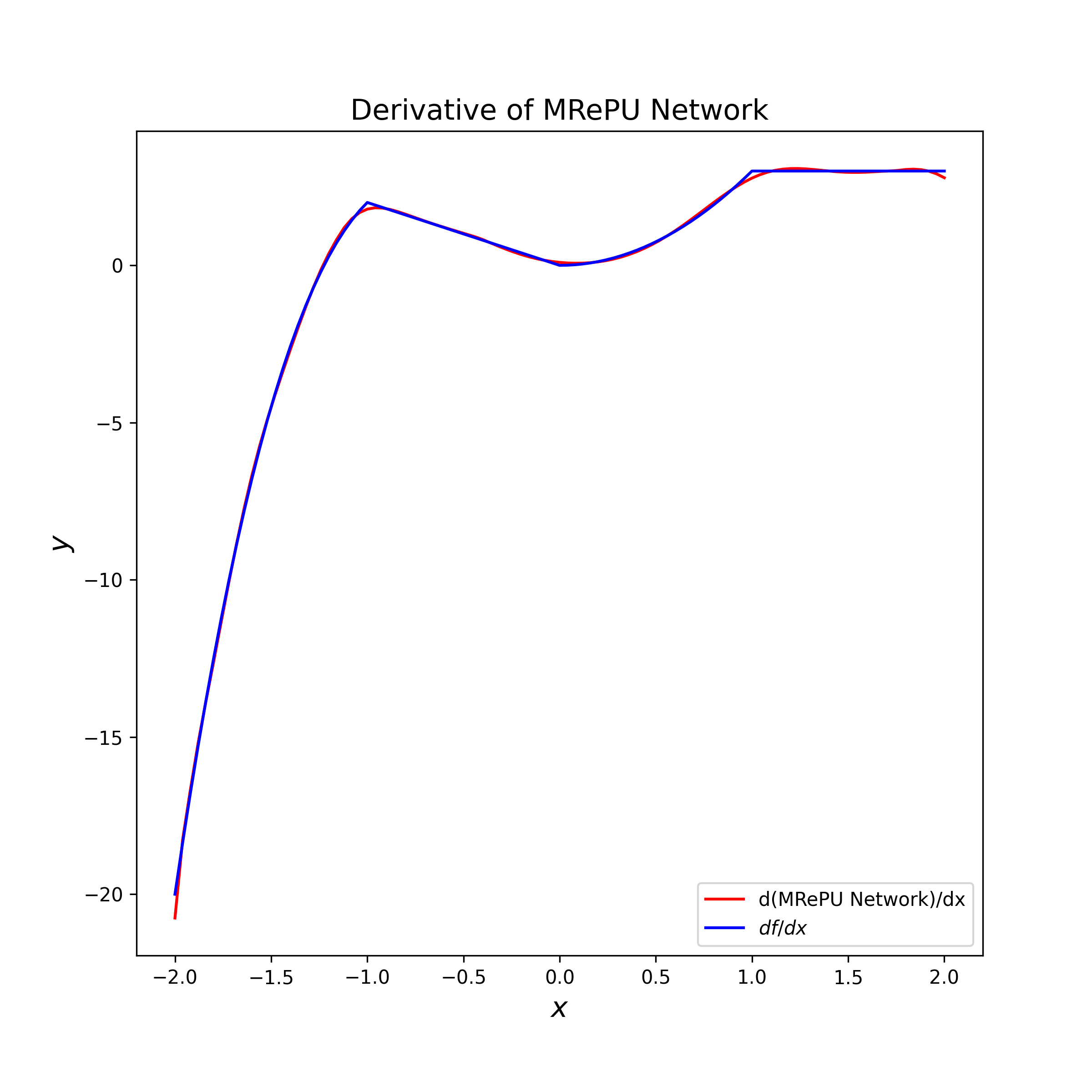}
        \includegraphics[width=0.49\textwidth]{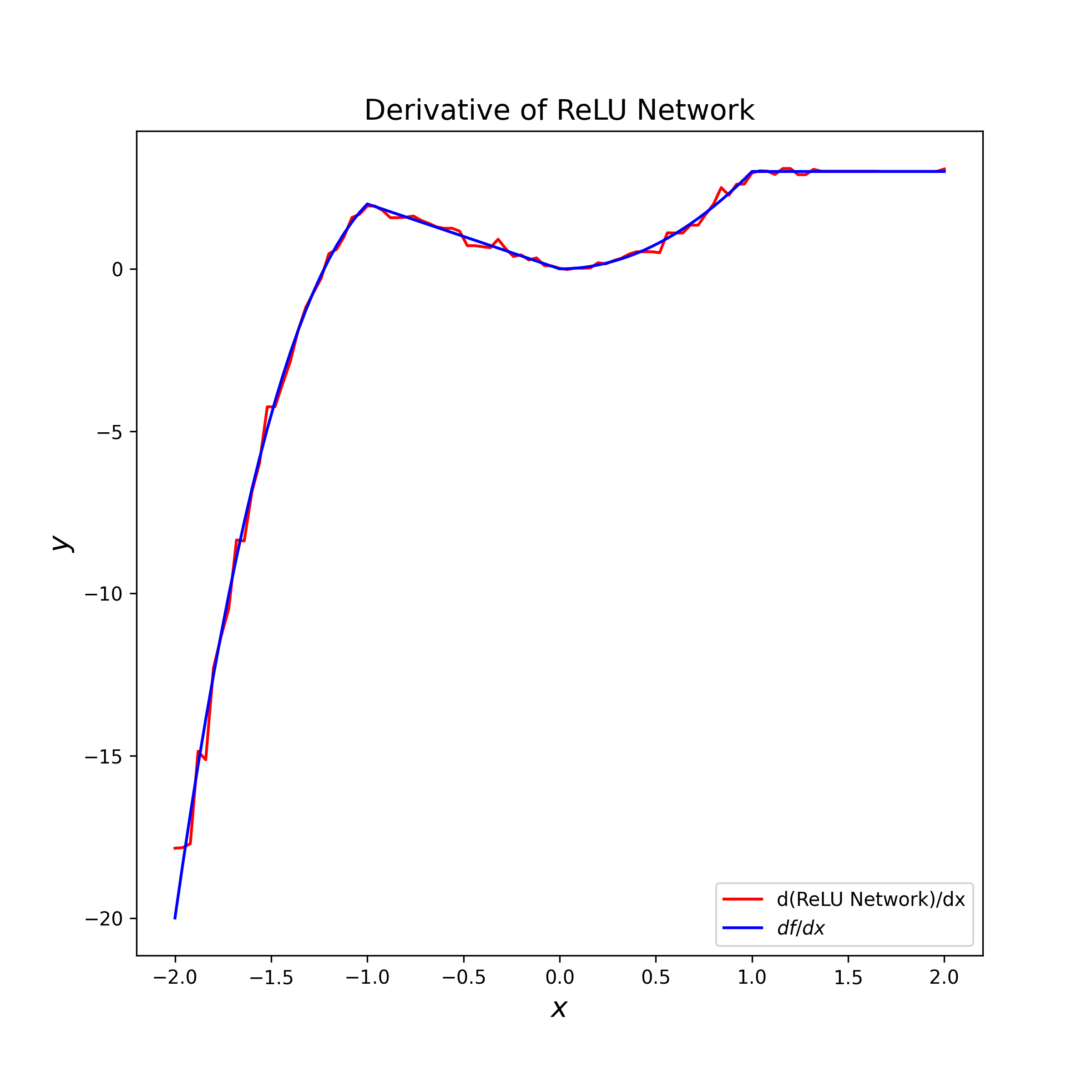}

    \end{center}
    \caption{\textbf{Top}: The results of training for regressing $f(x)$, where the left plot shows the output of the MRePU with $p=2$ network, and the right plot shows the output of the ReLU network.
\textbf{Bottom}: The comparison between the derivative of the trained networks and the actual derivative function.
     }
     
   \label{f reg}
   
\end{figure}

\begin{figure}[htp]
    \begin{center}
        \includegraphics[width=0.49\textwidth]{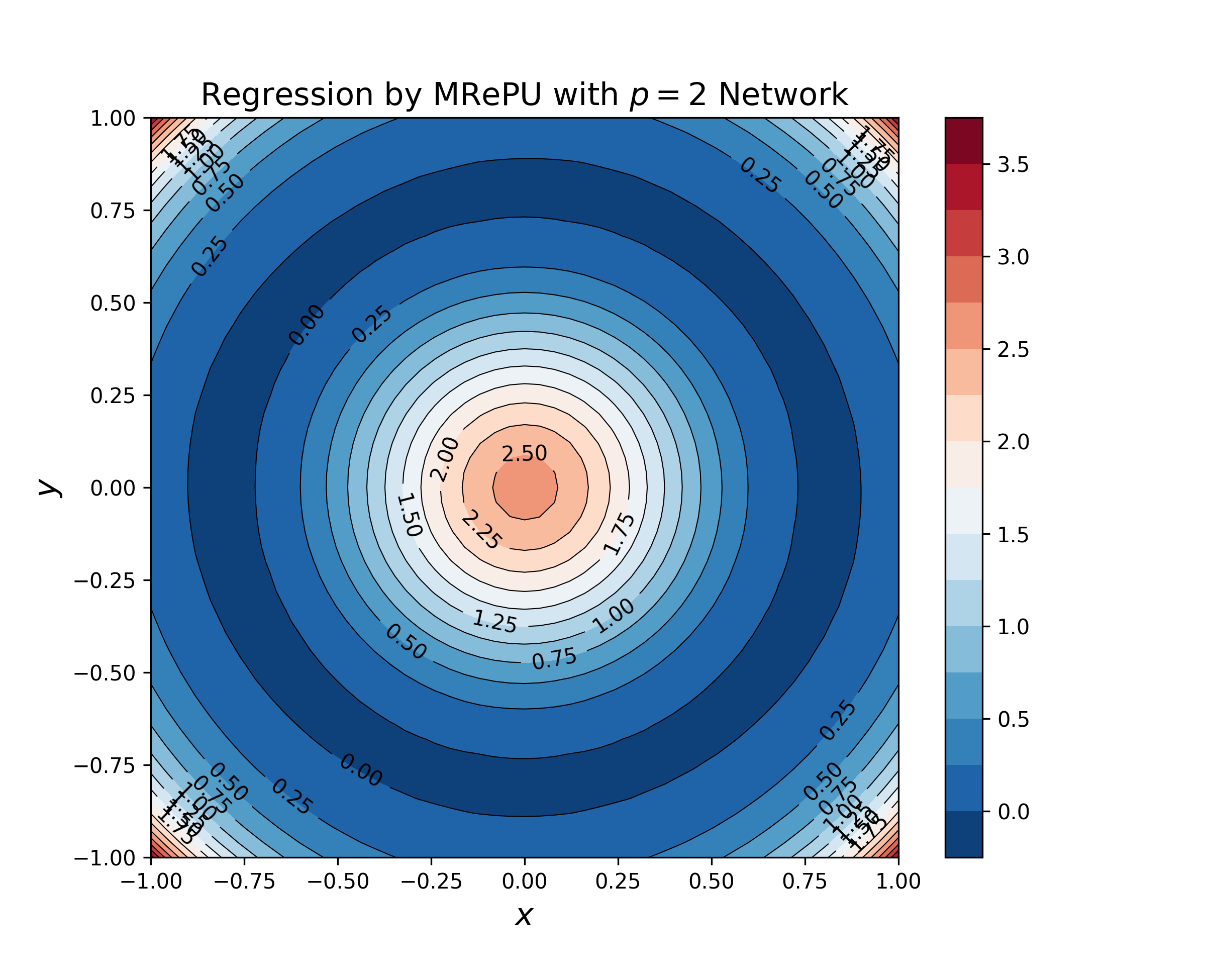}
        \includegraphics[width=0.49\textwidth]{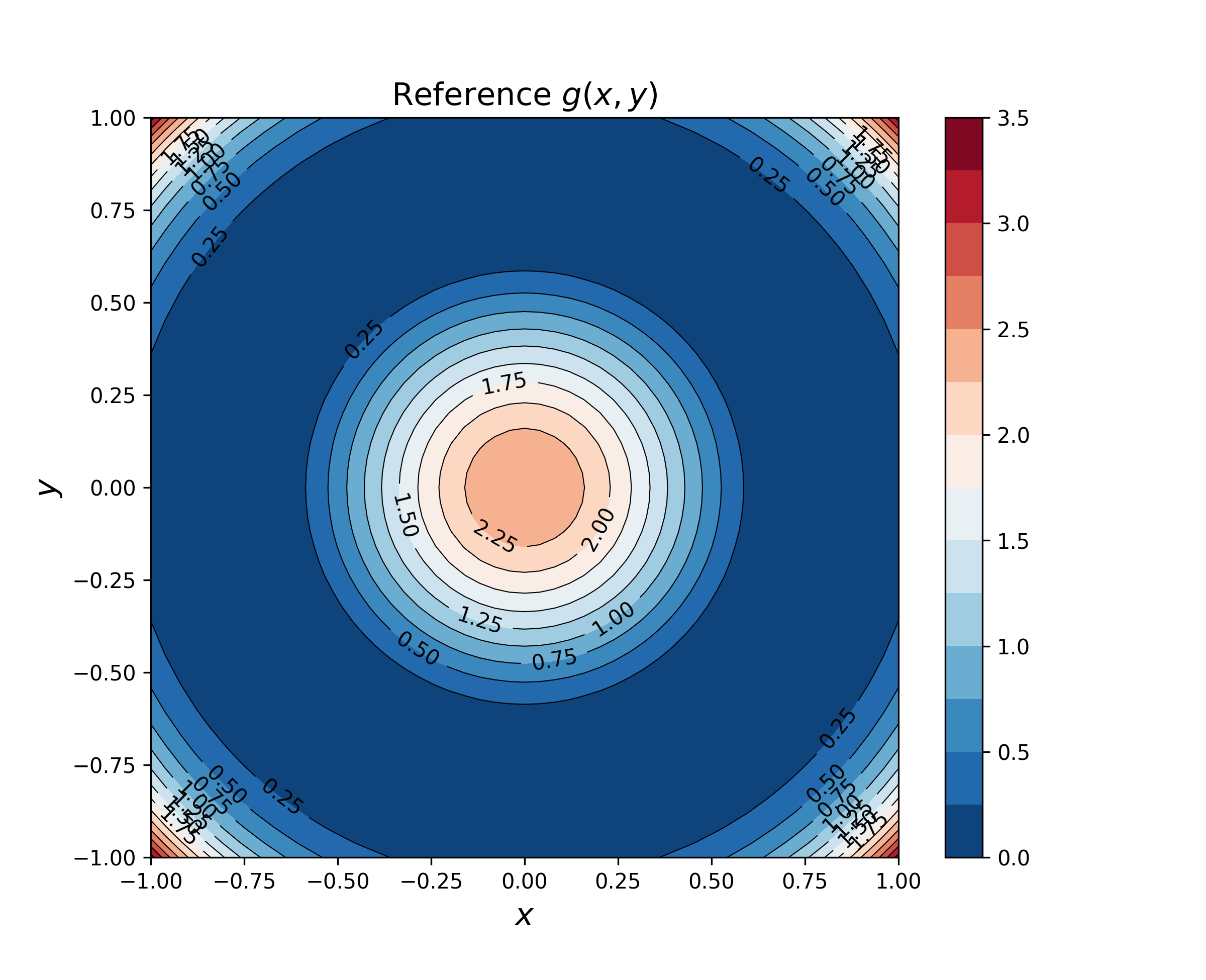}
        \\[\smallskipamount]
        \includegraphics[width=0.49\textwidth]{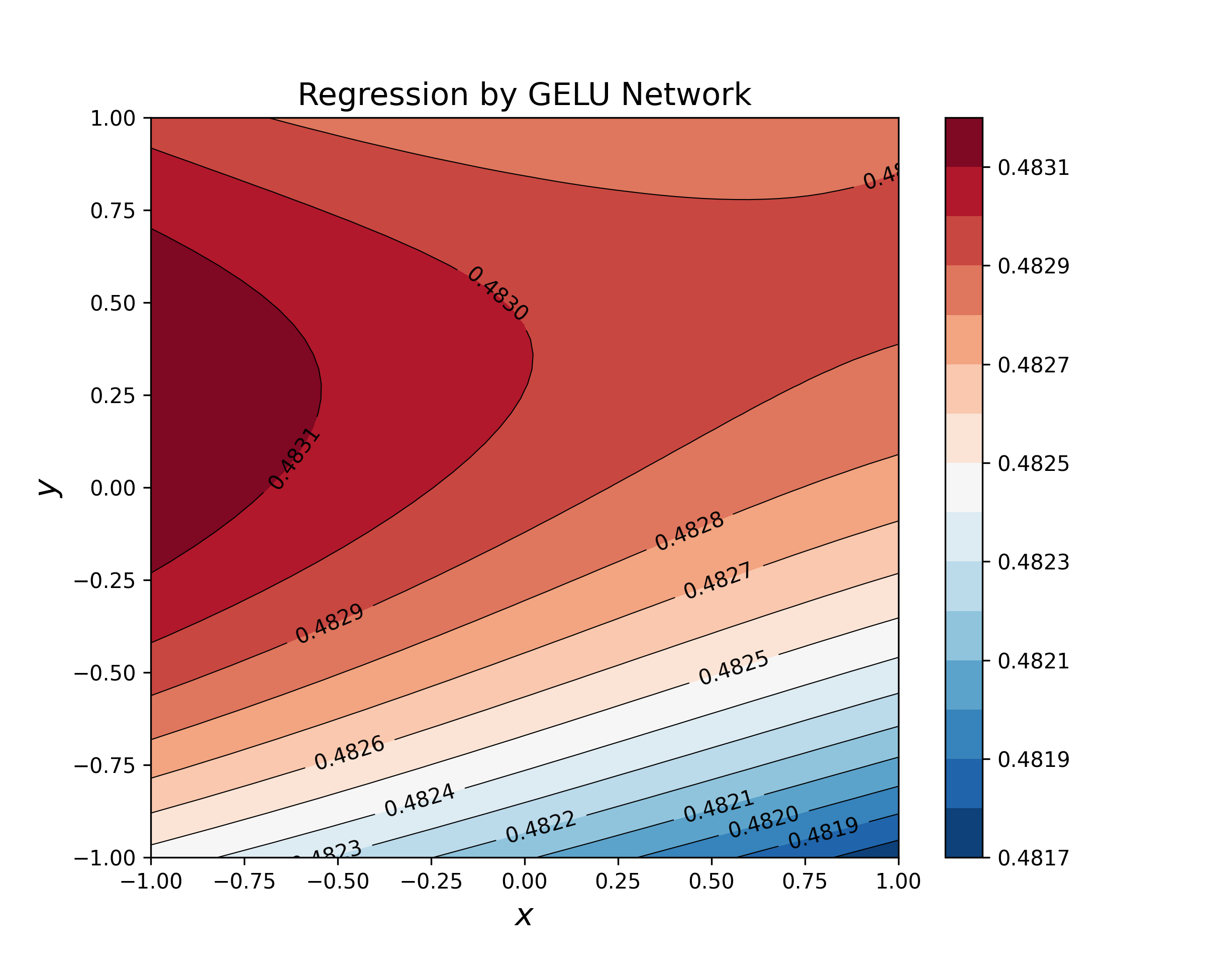}
        \includegraphics[width=0.49\textwidth]{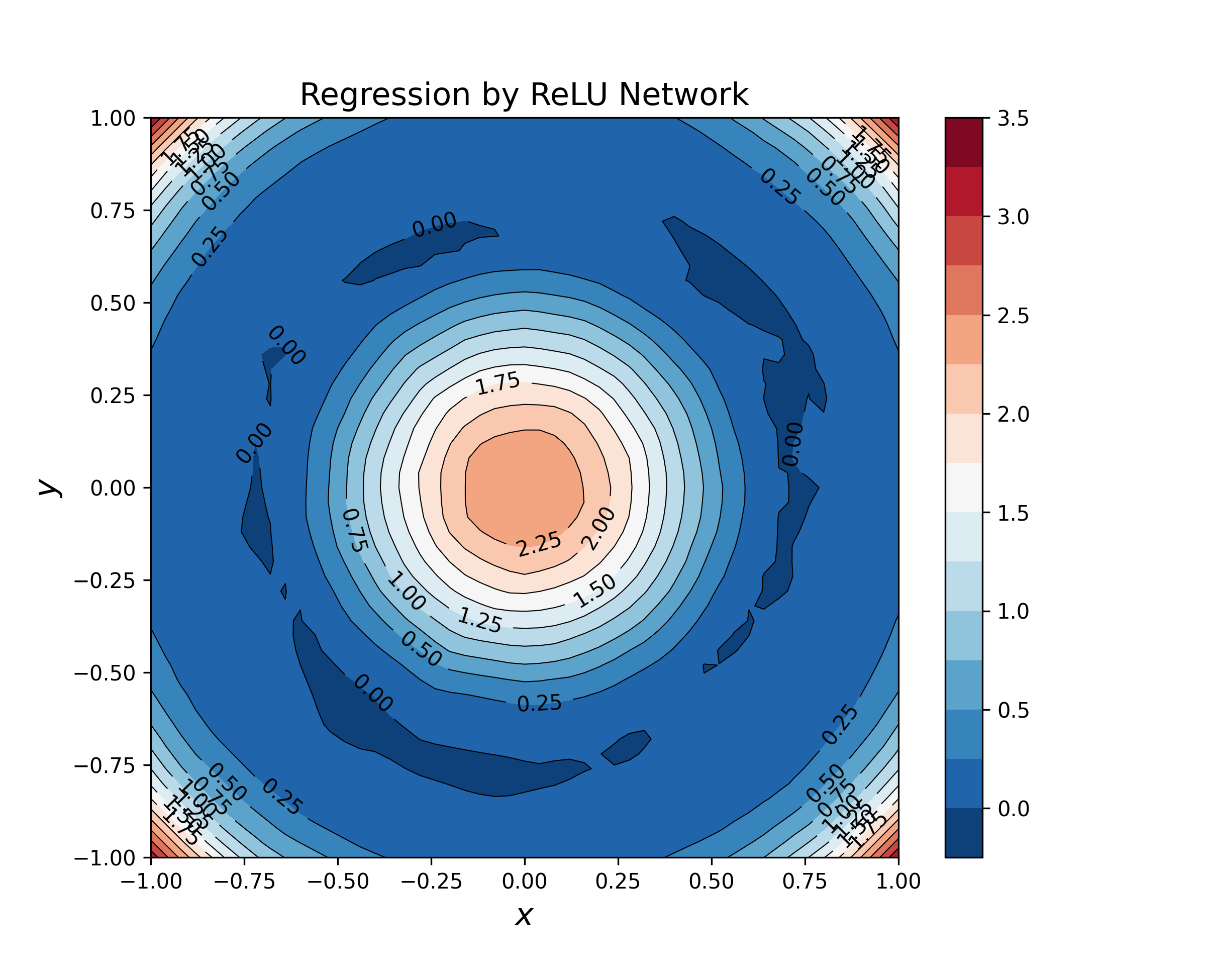}

    \end{center}
    \caption{Contour plots for the reference function and the trained neural networks. \textbf{Top left}: MRePU with $p=2$ network. \textbf{Top right}: Reference ($g(x,y)$). \textbf{Bottom left}: GELU network. \textbf{Bottom right}: ReLU network.
     }
     
   \label{g reg}
   
\end{figure}

}
\noindent
{\bf Case of Physics-Informed Neural Networks (PINNs).}{
Next, we compare neural networks with different activation functions in the framework of
physics-informed neural networks (PINNs)~\citep{Raissi19}, which solve a PDE by incorporating
the equation itself into the loss function. We consider four PDEs: the viscous Burgers equation,
the two-dimensional Poisson equation, the two-dimensional Helmholtz equation, and the
Allen--Cahn equation.

\paragraph{Burgers equation.}
\begin{equation*}
\begin{aligned}
  &\frac{\partial u}{\partial t} + u\,\frac{\partial u}{\partial x} = \nu\,\frac{\partial^{2} u}{\partial x^{2}},
  && x \in [-1,1],\ t \in [0,1],\\
  &u(x,0) = -\sin(\pi x), \qquad u(\pm 1, t) = 0,
\end{aligned}
\end{equation*}
with $\nu =\frac{0.1}{\pi} $ 
.
The reference solution is obtained from the Cole--Hopf transformation.

\paragraph{Poisson equation.}
\begin{equation*}
  \Delta u + 2\pi^{2}\sin(\pi x)\sin(\pi y) = 0 \quad \text{in } \Omega = [-1,1]^{2},
  \qquad u = 0 \quad \text{on } \partial\Omega,
\end{equation*}
whose exact solution is $u^{*}(x,y) = \sin(\pi x)\sin(\pi y)$.

\paragraph{Helmholtz equation.}
\begin{equation*}
  \Delta u + k^{2} u = q \quad \text{in } \Omega = [-1,1]^{2},
  \qquad u = 0 \quad \text{on } \partial\Omega,
\end{equation*}
with $k = 1$ and $q(x,y) = (k^{2} - 2\pi^{2})\sin(\pi x)\sin(\pi y)$, so that the exact solution is
again $u^{*}(x,y) = \sin(\pi x)\sin(\pi y)$.

\paragraph{Allen--Cahn equation.}
\begin{equation*}
\begin{aligned}
  &\frac{\partial u}{\partial t} - \epsilon\,\frac{\partial^{2} u}{\partial x^{2}} + \gamma\,(u^{3} - u) = 0,
  && x \in [-1,1],\ t \in [0,1],\\
  &u(x,0) = x^{2}\cos(\pi x), \qquad u(-1,t) = u(1,t), \qquad u_{x}(-1,t) = u_{x}(1,t),
\end{aligned}
\end{equation*}
with $\epsilon = 10^{-2}$ and $\gamma = 1$.
The reference solution is computed with a Fourier spectral method and ETDRK4 time stepping~\citep{Kassam05}.

The PINN loss is the sum of the mean squared PDE residual over $N_r = 10{,}000$ collocation points
sampled uniformly in the domain and the mean squared error of the initial and boundary conditions
over $N_b = 100$ points.
Every network has input dimension 2 (coordinates rescaled to $[-1,1]$), width 128, and
$L \in \{3, 5, 7\}$ hidden layers, and is trained with Adam (learning rate $5\times10^{-4}$,
$\beta = (0.9, 0.999)$, no weight decay) for $10^{4}$ iterations in single precision, with the same
training rule for every activation function. Biases are initialized to zero. For tanh, GELU and SiLU
we use $C_W = 1$ (He initialization, $C_W = 2$, for ReLU); RePU is initialized at the fixed point
$K_0 = 0.1$, and MRePU follows the initialization recipe of Algorithm~\ref{alg:recipe} with rule \eqref{eq:rule-par} and $\tau = 0.01$. We report the relative $L^2$ error of the parameters with the lowest training loss
(evaluated every 50 iterations; test data are not used), as the median over 3 seeds.
Table~\ref{tab:pinn} reports the relative $L^2$ errors (seed ranges in Appendix, Table~\ref{tab:pinn-range}).
RePU trains only at $L = 3$: it diverges at $L = 5$ and its training loss is already non-finite at
initialization for $L = 7$. MRePU with the initialization recipe
trains without divergence for every PDE and depth. Its accuracy is comparable to GELU and SiLU on the
Poisson and Helmholtz equations at $L = 5$ and better than tanh there, but about an order of magnitude
worse than GELU on the Burgers and Allen--Cahn equations. ReLU fails on all second-order problems. Training loss curves are given in Figure~\ref{fig:pinn-loss}. The solutions learned for each PDE are shown in Figures~\ref{fig:pinn-sol-burgers}--\ref{fig:pinn-sol-ac} in Appendix. We also tested Sobolev-type regularization in the form of gradient-enhanced PINNs \citep{Yu22}
($w = 10^{-2}$, Poisson and Helmholtz, $L = 5, 7$): the penalty did not systematically reduce the error
for any activation, and the ordering of the activations was unchanged (Appendix~\ref{app:gpinn}).
\begin{table}[htp]
\scriptsize
\setlength{\tabcolsep}{4pt}
\begin{tabular}{llccccccc}
\toprule
PDE & Depth & tanh & GELU & SiLU & ReLU & RePU & MRePU $(p{=}2)$ & MRePU $(p{=}3)$ \\
\midrule
\multirow{3}{*}{Burgers} & 3 & {\boldmath\bfseries \sci{1.6}{-3}} & \sci{1.7}{-3} & \sci{4.0}{-3} & \sci{4.6}{-1} & \sci{8.0}{-2} & \sci{3.0}{-1} & \sci{9.0}{-2} \\
 & 5 & \sci{1.3}{-3} & {\boldmath\bfseries \sci{2.5}{-4}} & \sci{4.6}{-4} & \sci{4.6}{-1} & div.$^{b}$ (3/3) & \sci{3.2}{-3} & \sci{1.7}{-1} \\
 & 7 & \sci{9.8}{-4} & \sci{3.2}{-4} & {\boldmath\bfseries \sci{3.2}{-4}} & \sci{4.7}{-1} & NaN$^{a}$ & \sci{2.3}{-3} & \sci{1.9}{-1} \\
\midrule
\multirow{3}{*}{Poisson} & 3 & \sci{1.2}{-2} & \sci{2.9}{-3} & {\boldmath\bfseries \sci{1.7}{-3}} & \sci{1.0}{0} & \sci{4.8}{-2} & \sci{7.7}{-2} & \sci{1.5}{-2} \\
 & 5 & \sci{1.3}{-2} & \sci{1.9}{-3} & {\boldmath\bfseries \sci{1.8}{-3}} & \sci{1.0}{0} & div.$^{b}$ (2/3) & \sci{2.5}{-3} & \sci{2.8}{-3} \\
 & 7 & \sci{1.6}{-2} & {\boldmath\bfseries \sci{1.2}{-3}} & \sci{2.3}{-3} & \sci{1.0}{0} & NaN$^{a}$ & \sci{6.2}{-3} & \sci{7.1}{-3} \\
\midrule
\multirow{3}{*}{Helmholtz} & 3 & \sci{9.9}{-3} & \sci{3.0}{-3} & {\boldmath\bfseries \sci{1.4}{-3}} & \sci{2.0}{1} & \sci{6.4}{-2} & \sci{4.0}{-2} & \sci{1.3}{-2} \\
 & 5 & \sci{1.1}{-2} & {\boldmath\bfseries \sci{1.7}{-3}} & \sci{2.1}{-3} & \sci{2.0}{1} & div.$^{b}$ (2/3) & \sci{3.1}{-3} & \sci{2.9}{-3} \\
 & 7 & \sci{1.7}{-2} & {\boldmath\bfseries \sci{1.4}{-3}} & \sci{2.1}{-3} & \sci{2.0}{1} & NaN$^{a}$ & \sci{5.7}{-3} & \sci{1.0}{-2} \\
\midrule
\multirow{3}{*}{Allen--Cahn} & 3 & {\boldmath\bfseries \sci{1.5}{-2}} & \sci{1.8}{-2} & \sci{2.7}{-2} & \sci{2.1}{-1} & \sci{4.1}{-2} & \sci{1.5}{-1} & \sci{1.5}{-1} \\
 & 5 & {\boldmath\bfseries \sci{2.1}{-2}} & \sci{2.3}{-2} & \sci{2.4}{-2} & \sci{1.8}{-1} & NaN$^{a}$ & \sci{5.0}{-2} & \sci{6.2}{-2} \\
 & 7 & \sci{2.8}{-2} & {\boldmath\bfseries \sci{1.2}{-2}} & \sci{1.8}{-2} & \sci{1.4}{-1} & NaN$^{a}$ & \sci{1.3}{-1} & \sci{5.0}{-2} \\
\bottomrule
\end{tabular}
\centering
\caption{Relative $L^2$ error of PINNs (median over 3 seeds; checkpoint with the lowest training loss).
Bold: lowest median in each row.
$^{a}$Non-finite training loss already at initialization (all seeds). $^{b}$Diverged: non-finite loss during training or relative $L^2$ error $>10^2$ (number of seeds in parentheses).}
\label{tab:pinn}
\end{table}

\begin{figure}[t]
\centering
\includegraphics[width=\textwidth]{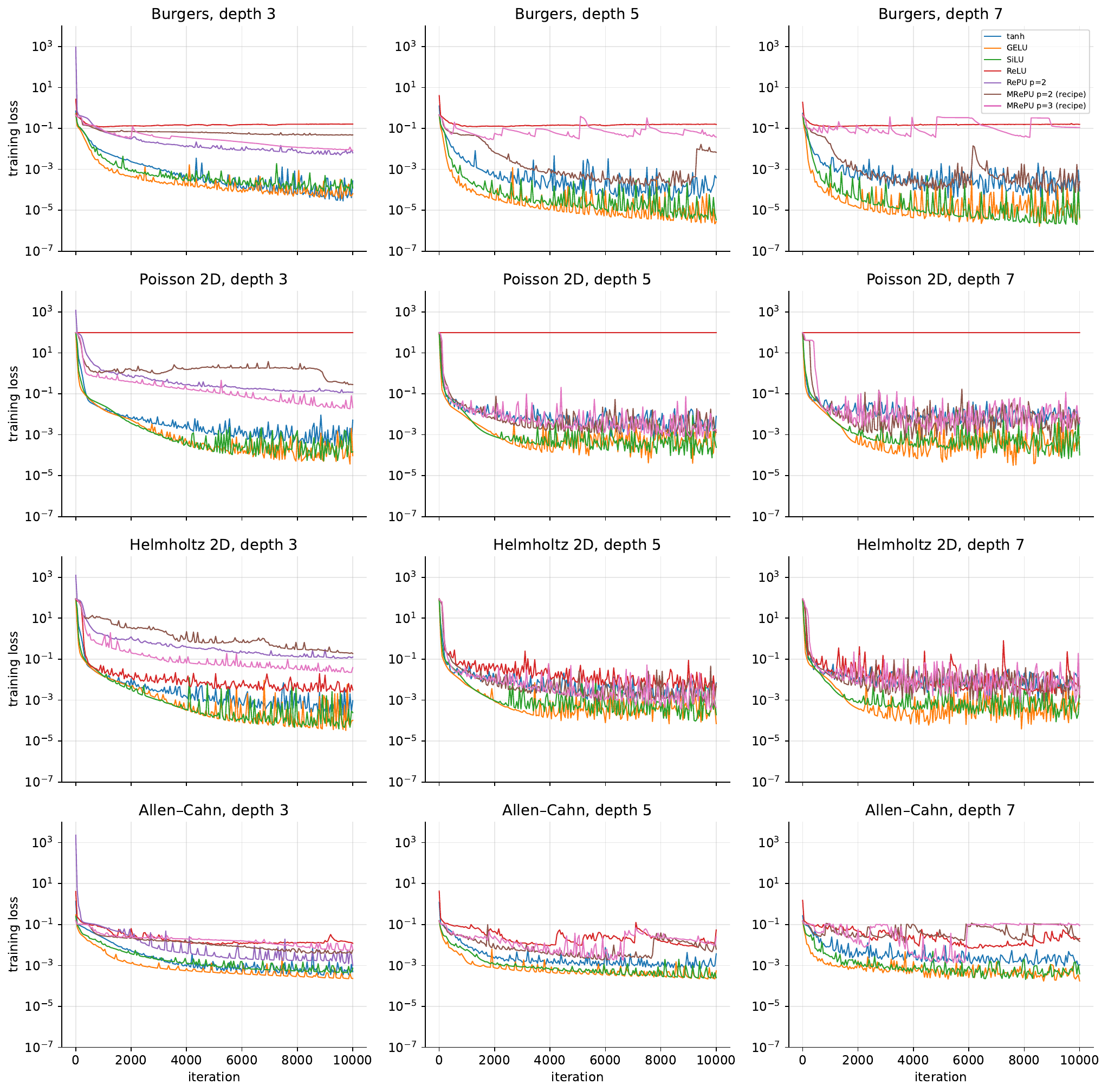}
\caption{Training loss of the PINNs in Table~\ref{tab:pinn} for each PDE (rows) and number of hidden
layers $L$ (columns). Each curve is the median over 3 seeds.}
\label{fig:pinn-loss}
\end{figure}

}
\paragraph{Summary.}
The experiments in this section address two distinct questions: whether a network can be
trained at all (trainability), and how accurate it is once trained (accuracy). Our results
support improved trainability of RePU-type networks with MRePU, but not a general improvement
in predictive accuracy.

\emph{Trainability.}
Section~3 showed that no choice of $C_W$ makes a RePU network critical, and accordingly RePU
networks became untrainable as depth increased. In the regression experiments of
Section~3.2 they trained only with a single hidden layer. In the PINN experiments, at least
two of the three seeds diverged on every PDE at $L=5$, and at $L=7$ the training loss was not
finite even at initialization. In contrast, all 72 MRePU runs in the PINN experiments (four
PDEs, three depths, $p\in\{2,3\}$, three seeds each), initialized by
Algorithm~\ref{alg:recipe}, trained without divergence (Table~\ref{tab:pinn}).

\emph{Accuracy.}
No single activation was the most accurate across tasks; the ranking depended on the target.
For the polynomial targets, MRePU attained the lowest mean training loss in all four cases
(Table~\ref{t:polyreg}), consistent with the polynomial approximation results of
Section~5.1. For the piecewise-defined targets, whose higher-order derivatives are
discontinuous, ReLU reached a training loss 1.4--2.5 times lower than MRePU
(Table~\ref{t:diffreg}), although the derivatives of the MRePU networks were visibly smoother
(Figs.~\ref{f reg} and~\ref{g reg}); we did not quantify this difference. In the PINN
experiments, GELU or SiLU gave the lowest relative $L^2$ error in nine of the twelve settings
(four PDEs at three depths), and tanh in the remaining three. On the Poisson and Helmholtz
equations, MRePU ($p=2$) came within a factor of two of GELU and SiLU at $L=5$ and within a
factor of five at $L=7$, and at both depths it was about three to five times more accurate
than tanh. On the Burgers and Allen--Cahn equations at the same depths, it was 2--13 times
less accurate than GELU. At $L=3$, MRePU was one to two orders of magnitude less accurate than
GELU on all four PDEs, a gap that our initialization analysis does not explain. ReLU failed on
every PDE, as expected, since its second derivative vanishes almost everywhere.

\subsection{Experiments on Real-World Data}
 In this section, we extend our experiments beyond synthetic data to real-world datasets, specifically MNIST and CIFAR-10. Consistent with the setup in Section 5.2, we trained neural network architectures employing MRePU and various activation functions. On CIFAR-10, we also examine how the trainability of MRePU networks depends on the residual architecture.

\noindent
\subsubsection{\bf MNIST Classification.}
{
First, we conducted experiments on the MNIST dataset, which is fundamental and widely used for verification. The $28 \times 28$ input images were flattened into 784-dimensional vectors, and a FCN architecture with width 256 was employed. 

\paragraph{Trainability across input scale and weight variance.}
In the first experiment, we test whether the fixed point of the kernel recursion marks the boundary of trainability.
We train MLPs of width $256$ and depth $L=3$ and $L=5$ with MRePU activations ($p=2$) on MNIST, using batch size $512$.
The inputs are centered and divided by $N_K$, which ranges from $0.1$ to $2.0$ in steps of $0.05$.
All weights are drawn with variance $C_W/n_{\ell-1}$ and all biases are set to zero, where $C_W=s^2$ and $s$ ranges from $0.1$ to $2.0$ in steps of $0.05$.
Since our analysis concerns the kernel at initialization, we assess trainability with short runs of one epoch.
Each configuration is trained from $3$ independent initializations, and Fig.~\ref{pdmnist} shows the mean test accuracy and the fraction of runs that diverged.
The dashed line in Fig.~\ref{pdmnist} is $C_W^*=g^{-1}(q)/q$ with $q=\sigma_x^2/N_K^2$, and follows from the fixed-point condition~\eqref{eq:rule-fp}.
Since the first layer uses the same $C_W$, the input kernel is $K^{(1)}=C_Wq$; requiring it to be the fixed point of $K\mapsto C_Wg(K)$ gives $g(C_Wq)=q$.
 As shown in the figure, the success of training is highly sensitive to $C_W$.
For $L=3$, networks train over a wider range than the fixed-point curve predicts, presumably because a shallow network has too few layers for the growth of the kernel to accumulate.
For the deeper network ($L=5$), the boundary of successful training follows the fixed-point curve.
As for the input scale, training is more likely to diverge for small $N_K$ ranges, and for $L=5$ it diverges for every $C_W$ in this range ($N_K<0.5$), even though the kernel at initialization decreases there.
Since standard normalization (dotted line) lies in this range, we conclude empirically that the inputs of an MRePU network must be scaled down substantially relative to standard normalization.
\paragraph{Rescaled depth and the choice of $C_W$.}
The second experiment assesses how well Algorithm~\ref{alg:recipe} works in practice.
We evaluate each rule across depths, rescaled depths and activations on MNIST.
We use MLPs of width $256$ and depth $L\in\{4,8,16,32\}$, MRePU with $p\in\{2,3\}$, rescaled depths $\tau\in\{0.01,0.03,\dots,10\}$,
and the rules~\eqref{eq:rule-par}, \eqref{eq:rule-fp} and $C_W=1$ ($\kappa=2,1,0$).
Following Algorithm~\ref{alg:recipe}, the inputs are scaled by $s=\sqrt{K_0/q}$ and the first-layer weights have variance $1/d$;
for all activations, the read-out is rescaled to unit initial logit standard deviation.
Each setting is trained for $3$ epochs with $3$ seeds using Adam, with the better of two learning rates $\{10^{-3},2\times 10^{-4}\}$ chosen for each setting.
In Figure~\ref{fig:t3-paper}a, a run counts as successful if its test accuracy is at least $0.5$; diverged runs are counted as chance accuracy.
As baselines we use ReLU, GELU and SiLU with standard initialization, and RePU with the best of the initializations we tested.
The value of $\tau$ at which training fails decreases with depth (Figure~\ref{fig:t3-paper}b).
Unlike the kernel at initialization, which depends on depth only through $\tau$, the outcome of training thus depends on depth beyond $\tau$,
and the largest safe $\tau$ is an empirical quantity.
Training with rule~\eqref{eq:rule-par} succeeds over the widest range of $\tau$, for both $p=2$ and $p=3$.
The ranges are ordered as \eqref{eq:rule-par} $>$ \eqref{eq:rule-fp} $>$ $C_W=1$.
This ordering is consistent with the margin $K_\star/K_0=\kappa$ in Proposition~10, which measures how much the kernel can grow before the recursion diverges.
At $\tau=10^{-2}$, MRePU trains at every depth (Figure~\ref{fig:t3-paper}c), whereas RePU fails for $L\geq8$ even with its best initialization.
The other baselines reach test accuracies of $0.95$--$0.97$, higher than MRePU ($0.92$--$0.94$).
The point of this experiment is therefore the trainability of MRePU networks, not an advantage in accuracy.
Together, Algorithm~\ref{alg:recipe}, Proposition~10 and the empirical findings of this section allow one to choose a concrete $C_W$ and input scale
for a given depth, activation and input range.

\begin{figure}[htp]
\centering
\includegraphics[width=0.9\textwidth]{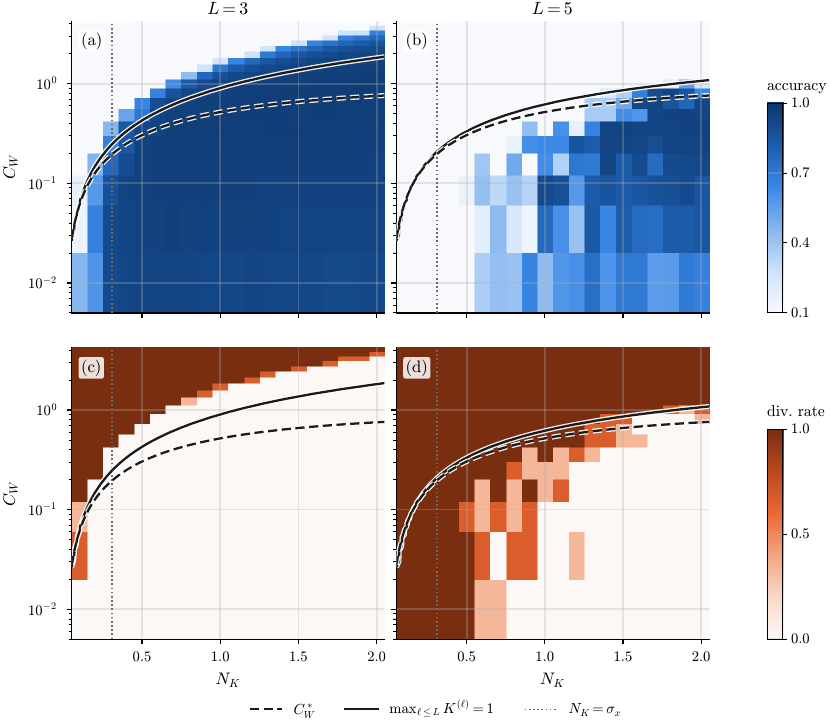}
\caption{Phase diagram of MRePU ($p=2$) networks on MNIST as a function of the input scale $N_K$ and the weight variance $C_W$.
(a,b) Test accuracy; (c,d) fraction of runs that diverged during training.
The dashed line, $C_W^* = g^{-1}(q)/q$, is the fixed point of the kernel recursion and gives the boundary in the limit of infinite depth.
The solid line marks where the kernel at initialization reaches $1$ in at least one of the $L$ layers.
The dotted line marks standard normalization ($N_K = \sigma_x$).}\label{pdmnist}
\end{figure}

\begin{figure}[htp]
\centering
\includegraphics[width=\textwidth]{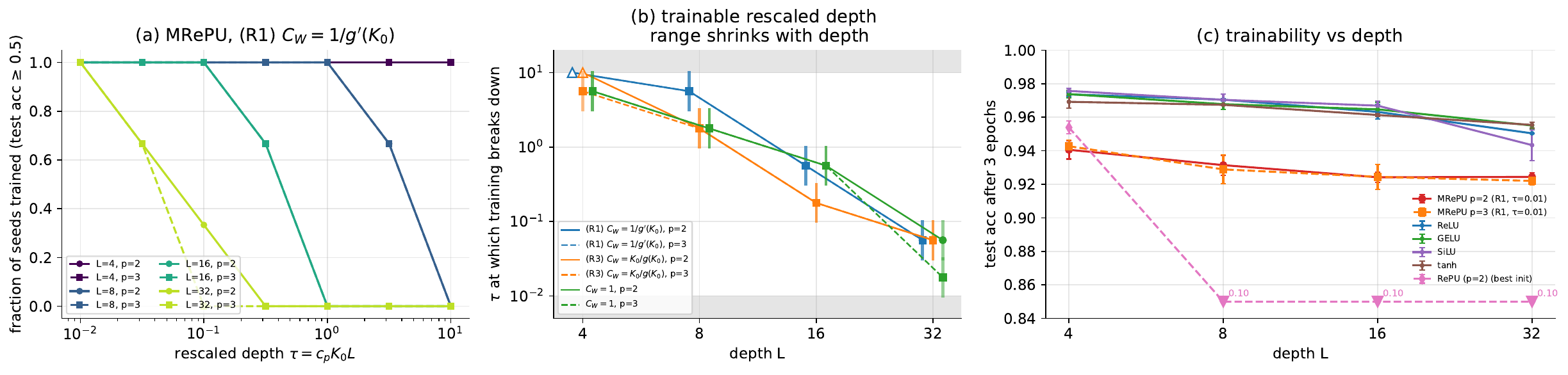}
\caption{Rescaled depth of MRePU networks on MNIST.
(a) Fraction of seeds reaching test accuracy $\geq 0.5$ as a function of the rescaled depth $\tau = c_p K_0 L$ under rule~\eqref{eq:rule-par};
colors indicate depth, solid and dashed lines $p=2$ and $p=3$.
(b) Value of $\tau$ at which training breaks down for rules~\eqref{eq:rule-par}, \eqref{eq:rule-fp} and $C_W=1$.
(c) Test accuracy versus depth (mean $\pm$ s.d.\ over seeds) for MRePU initialized with Algorithm~\ref{alg:recipe}
(rule~\eqref{eq:rule-par}, $\tau = 10^{-2}$), RePU with the best of the tested initializations, and standard activations.}
\label{fig:t3-paper}
\end{figure}
}

\noindent
\subsubsection{\bf CIFAR-10 Classification.}
Next, we consider CIFAR-10 with residual networks.
If the MRePU network uses a different residual block than the baselines, the differences in the results cannot be attributed to the activation function alone.
We therefore cross six activation functions with six residual architectures and train every combination with the same procedure.
As in the MNIST experiments, we distinguish whether a network can be trained at all (trainability) from how accurate it is once trained;
the analysis of Sections~3 and~4 concerns only the former.
 
\paragraph{Architectures.}
All networks share a ResNet backbone for $32\times32$ images \citep{He15b}.
A $3\times3$ convolution with 128 output channels (the stem) is followed by four stages of $2,2,1,1$ residual blocks with $128,256,512$ and $1024$ channels,
and the first block of the second, third and fourth stages halves the spatial resolution with stride $2$.
Each block contains two $3\times3$ convolutions $W_1$ and $W_2$.
When the number of channels changes, the shortcut $S$ is a $1\times1$ convolution with stride $2$; otherwise it is the identity.
Global average pooling and a linear read-out follow the last stage.
The network thus has 14 weight layers (the stem, 12 convolutions in the blocks and the read-out), 13 activations on its longest input--output path, and about 21 million parameters.
The architectures differ in the position of batch normalization (BN), in how the residual branch and the shortcut are combined, and in the initial value of the BN scales.
A block maps its input $z$ to
\begin{align}
  \text{post-BN:}\quad & z' = \sigma\Bigl(\alpha\,\mathrm{BN}\bigl(W_2\,\sigma(\mathrm{BN}(W_1 z))\bigr) + \beta\,S(z)\Bigr), \label{eq:block-post}\\
  \text{pre-BN:}\quad & z' = W_2\,\sigma\bigl(\mathrm{BN}(W_1 u)\bigr) + S(z), \qquad u = \sigma(\mathrm{BN}(z)), \label{eq:block-pre}\\
  \text{normalizer-free:}\quad & z' = \gamma\,W_2\,\sigma\bigl(W_1\,\sigma(z)\bigr) + \sqrt{1-\gamma^2}\,S(z). \label{eq:block-nf}
\end{align}
In \eqref{eq:block-post} a projection shortcut is followed by BN, and in \eqref{eq:block-pre} it acts on $u$ instead of $z$, as in \citet{He16identity}.
The post-BN networks apply BN and $\sigma$ after the stem, the pre-BN networks apply BN and $\sigma$ after the last block, and the normalizer-free network applies $\sigma$ after the last block.
Table~\ref{tab:cifar-arch} lists the six architectures; we use $\gamma=1/2$ throughout.
 
\begin{table}[htp]
\centering
\caption{Residual architectures for CIFAR-10. All share the backbone described in the text.
BN scale: initial value of the scale parameter of every BN layer.
$c_2K_0$: value of $c_2K$ at the MRePU ($p=2$) inputs at initialization, set by the BN scale or, for NF-$\gamma$, by Algorithm~\ref{alg:recipe}.
$^{\dagger}$Larger at the activation that follows the residual sum.}
\label{tab:cifar-arch}
\small
\begin{tabular}{lllcc}
\toprule
Name & Block & Combination & BN scale & $c_2K_0$ \\
\midrule
Post-BN               & \eqref{eq:block-post} & $\alpha=\beta=1$                          & $1$                       & $18^{\dagger}$ \\
Post-BN-$\gamma$      & \eqref{eq:block-post} & $\alpha=\gamma,\ \beta=\sqrt{1-\gamma^2}$ & $1$                       & $18^{\dagger}$ \\
Post-BN-$\gamma$-crit & \eqref{eq:block-post} & $\alpha=\gamma,\ \beta=\sqrt{1-\gamma^2}$ & $\sqrt{K_0}\approx0.041$  & $0.03$ \\
Pre-BN                & \eqref{eq:block-pre}  & unweighted sum                            & $1$                       & $18$ \\
Pre-BN-crit           & \eqref{eq:block-pre}  & unweighted sum                            & $\sqrt{K_0}\approx0.041$  & $0.03$ \\
NF-$\gamma$           & \eqref{eq:block-nf}   & $\gamma,\ \sqrt{1-\gamma^2}$              & -- (no BN)                & $0.0033$ \\
\bottomrule
\end{tabular}
\end{table}
 
Post-BN is the standard ResNet \citep{He15b} and Pre-BN is the pre-activation ResNet \citep{He16identity}.
In Post-BN-$\gamma$ and NF-$\gamma$, the weights $\gamma$ and $\sqrt{1-\gamma^2}$ make the shortcut preserve the kernel \citep{Roberts22}.
The crit variants follow Remark~\ref{rem:arch}: all BN scales, including those after the stem and on the shortcuts of Post-BN-$\gamma$-crit and the final one of Pre-BN-crit, are initialized to $\sqrt{K_0}$,
so that every activation that follows BN receives the kernel $K_0$ at initialization.
We choose $c_2K_0=0.03$ ($K_0=1.7\times10^{-3}$), a small value in view of Section~5.3.1, where the largest $\tau$ at which training succeeded decreased with depth.
To keep the architecture and the activation separate, the crit variants use this scale for every activation, computed with $p=2$; for MRePU with $p=3$ it gives $c_3K_0=0.075$.
At this scale GELU and SiLU start close to their linear regime, whereas the nonlinearity of the positively homogeneous ReLU does not depend on the scale.
An architecture in this section is therefore a block together with the initialization of its normalization layers.
 
\paragraph{Activations and initialization.}
We compare ReLU, GELU, SiLU, RePU with $p=2$ and MRePU with $p\in\{2,3\}$.
In the five architectures with BN, every activation that follows a BN layer receives the kernel set by the BN scale.
There, the stem and the $3\times3$ convolutions use $C_W=2$ for every activation and the $1\times1$ shortcut convolutions use $C_W=1$; the convolutions have no biases, and BN shifts start at zero.
In NF-$\gamma$ the initialization depends on the activation.
The stem and the $1\times1$ shortcuts are linear maps and use $C_W=1$, which preserves the kernel.
ReLU, GELU and SiLU use $C_W=2$ for the $3\times3$ convolutions.
RePU is initialized at the fixed point $K_0=0.1$, i.e.\ $C_W=K_0/g(K_0)\approx6.7$ with inputs scaled by $\sqrt{K_0}$; by Section~3 this fixed point is unstable.
MRePU follows Algorithm~\ref{alg:recipe} with rule~\eqref{eq:rule-par}, $\tau=10^{-2}$ and $L=12$ MRePU layers on the residual branches, so that $L_{\mathrm{eff}}=\gamma^2L=3$ (Remark~\ref{rem:arch}).
This gives $c_pK_0=\tau/L_{\mathrm{eff}}=3.3\times10^{-3}$ and $C_W=0.993$ for both degrees,
with $K_0=1.9\times10^{-4}$ and input scale $s=0.014$ for $p=2$, and $K_0=7.4\times10^{-5}$ and $s=0.0086$ for $p=3$.
The kernels measured at initialization agree with these values: they lie between $0.6K_0$ and $1.15K_0$ at every MRePU input of NF-$\gamma$,
and between $1.66\times10^{-3}$ and $1.67\times10^{-3}$ at every MRePU input of Pre-BN-crit.
The read-out uses the default PyTorch initialization for all activations, instead of step~7 of Algorithm~\ref{alg:recipe}.
 
\paragraph{Training and evaluation.}
All combinations are trained with the same procedure: AdamW \citep{Loshchilov19} with decoupled weight decay $5\times10^{-4}$, batch size 128 and 5 epochs (1{,}955 steps),
a linear warm-up over the first epoch followed by cosine decay to zero, and the gradient norm clipped at $1$ for every activation.
Training images are augmented by random $32\times32$ crops from images padded by 4 pixels (reflection) and by random horizontal flips;
inputs are standardized per channel, and training uses bfloat16 mixed precision with the loss computed in single precision.
For each combination, the learning rate is selected from \{$5\times10^{-4}$, $10^{-3}$, $2\times10^{-3}$\} by the validation accuracy after 8 epochs of training on 45{,}000 training images,
with the remaining 5{,}000 held out (seed $0$); diverged runs are ranked last.
With the selected rate we train three networks on all 50{,}000 training images (seeds $0,1,2$; the seed fixes the initialization, the data order and the augmentation)
and report the test accuracy after the last epoch; the test set is not used for any selection.
A run is \emph{diverged} if its training loss becomes non-finite, and \emph{stalled} if the loss stays finite but the training accuracy in the last epoch is below $0.2$ (chance level is $0.1$); all other runs are \emph{trained}.
We also record the kernel at every activation input at initialization, estimated by the mean squared pre-activation on 256 training images with BN in training mode.
 
\begin{table}[htp]
\footnotesize
\setlength{\tabcolsep}{3pt}
\begin{tabular}{lcccccc}
\toprule
 & ReLU & GELU & SiLU & RePU & MRePU & MRePU \\
Architecture & & & & $(p{=}2)$ & $(p{=}2)$ & $(p{=}3)$ \\
\midrule
Post-BN               & $0.873\pm0.003$ & {\boldmath$0.881\pm0.004$} & $0.876\pm0.003$ & stall$^{b}$ (3/3) & div.$^{a}$ (3/3) & div.$^{a}$ (3/3) \\
Post-BN-$\gamma$      & $0.869\pm0.004$ & {\boldmath$0.876\pm0.003$} & $0.865\pm0.003$ & stall$^{b}$ (3/3) & div.$^{a}$ (3/3) & div.$^{a}$ (3/3) \\
Post-BN-$\gamma$-crit & {\boldmath$0.874\pm0.002$} & $0.682\pm0.005$ & $0.615\pm0.007$ & stall$^{b}$ (3/3) & $0.727$ (1/3)$^{c}$ & stall$^{b}$ (3/3) \\
Pre-BN                & $0.875\pm0.001$ & {\boldmath$0.881\pm0.001$} & $0.869\pm0.001$ & stall$^{b}$ (3/3) & stall$^{b}$ (3/3) & stall$^{b}$ (3/3) \\
Pre-BN-crit           & {\boldmath$0.879\pm0.004$} & $0.679\pm0.006$ & $0.634\pm0.017$ & stall$^{b}$ (3/3) & $0.725\pm0.001$ & $0.756$ (1/3)$^{d}$ \\
NF-$\gamma$           & $0.833\pm0.001$ & {\boldmath$0.876\pm0.002$} & $0.873\pm0.002$ & div.$^{a}$ (3/3) & $0.733\pm0.008$ & $0.706\pm0.003$ \\
\bottomrule
\end{tabular}
\centering
\caption{Test accuracy on CIFAR-10 after 5 epochs (mean $\pm$ standard deviation over 3 seeds).
Bold: highest mean in each row.
$(1/3)$: only one of the three seeds trained, and its test accuracy is shown.
$^{a}$Diverged: activations already non-finite at initialization (all seeds).
$^{b}$Stalled: finite loss but training accuracy below $0.2$ in the last epoch.
$^{c}$The other two seeds reached a training accuracy of $0.70$ but a test accuracy at chance level.
$^{d}$In the other two seeds, the training accuracy rose to $0.52$ and $0.35$ and then fell to chance level.}
\label{tab:cifar}
\end{table}
 
\paragraph{Results.}
Table~\ref{tab:cifar} shows the results.
ReLU, GELU and SiLU train for every seed in all six architectures.
In the three architectures with unit BN scales their test accuracies lie between $0.865$ and $0.881$, and the accuracy of ReLU varies by at most $0.046$ across all six architectures.
GELU and SiLU, in contrast, reach only $0.615$--$0.682$ in the crit variants, with training accuracies below $0.67$, consistent with their start close to the linear regime.
 
RePU trains in none of the 18 runs.
It stalls at chance level in the five architectures with BN, and in NF-$\gamma$ its forward pass overflows already at initialization for every seed.
For seed $0$, the kernel at successive activation inputs of NF-$\gamma$ is $0.10$, $0.21$, $0.51$, $35$, \sci{1.1}{5}, \sci{2.0}{13} and \sci{5.0}{29}, and it is not finite at the eighth.
Since the fixed point $K_0$ is unstable (Section~3), inputs whose kernel exceeds it are amplified from layer to layer.
 
For MRePU, the outcome is decided by the kernel at its inputs.
In Post-BN and Post-BN-$\gamma$, the activation after the residual sum receives, in addition to a BN output, the shortcut that carries the output of the previous MRePU.
At initialization its kernel is already $32$--$46$ for $p=2$ and $590$--$1{,}030$ for $p=3$ in the first block, reaches $10^{10}$--$10^{25}$ in the second block, and is not finite thereafter, for every seed.
Gradient clipping cannot prevent this, because the overflow occurs in the forward pass.
In Pre-BN every MRePU input has kernel $1$ ($c_2K=18$).
These networks do not diverge, but they stall at chance level for both degrees and all seeds, although the training loss decreases from about $22$ to $3.0$ ($p=2$) and from about $900$ to $14$--$16$ ($p=3$).
When this kernel is small, MRePU with $p=2$ trains for every seed in Pre-BN-crit ($c_2K_0=0.03$; $0.725\pm0.001$) and in NF-$\gamma$ ($c_2K_0=0.0033$; $0.733\pm0.008$),
and MRePU with $p=3$ trains for every seed in NF-$\gamma$ ($0.706\pm0.003$).
At $c_3K_0=0.075$ in the crit variants, MRePU with $p=3$ stalls in Post-BN-$\gamma$-crit and trains for only one of three seeds in Pre-BN-crit.
In Post-BN-$\gamma$-crit, MRePU with $p=2$ reaches a training accuracy of $0.70$--$0.71$ for every seed, but its test accuracy after the last epoch is at chance level for two of them; we return to this below.
A comparison of ReLU in Post-BN with MRePU in Post-BN-$\gamma$ would thus attribute to the activation what is an interaction between activation and architecture:
ReLU is nearly insensitive to the architecture, whereas MRePU trains only where the architecture keeps the kernel at its inputs small.
The learning rate selection runs show the same pattern for width 64 and for a deeper backbone with $2,2,2,2$ blocks (Appendix~\ref{app:cifar}).
 
\paragraph{Trainability and accuracy.}
Where MRePU trains, its test accuracy ($0.70$--$0.76$) is $0.10$--$0.15$ below that of ReLU in the same architecture.
Its training accuracy ($0.70$--$0.74$, against $0.83$--$0.89$ for ReLU) is lower by a similar amount, so the gap reflects slower optimization within five epochs rather than overfitting.
GELU and SiLU, started from the same small kernel in the crit variants, reach even lower accuracies ($0.615$--$0.682$).
This suggests that part of the gap is due to the small kernel required for stability, with which the network starts close to linear; we did not test this further.
MRePU also required small learning rates: in every architecture in which it trained at smaller rates, it failed at $2\times10^{-3}$, and the selected rate was $5\times10^{-4}$ in all but one combination (Appendix~\ref{app:cifar}).
This is consistent with Section~5.3.1, where the largest safe $\tau$ decreased because training moves the kernel away from its initial value.
We therefore do not claim an accuracy advantage for MRePU on CIFAR-10.
What this experiment supports is the statement on trainability: a RePU-type network becomes trainable once the kernel at the inputs of its activations is small,
and the standard post- and pre-activation ResNets do not keep this kernel small.
The accuracies are those reached within five epochs and are not converged values.
 
\paragraph{Normalization does not remove the instability.}
BN fixes the variance of each channel, but not the tails of its distribution.
In Post-BN-$\gamma$-crit, the activation after the residual sum also receives the identity shortcut, i.e.\ the unnormalized output of the previous block.
At initialization, the kernel at one such input (the second block of the second stage) was $0.39$, $90$ and $1.06$ for MRePU with $p=3$ in the three seeds, $230$ to \sci{5}{4} times its target $1.7\times10^{-3}$,
while for $p=2$ it stayed within $25\%$ of the target.
In Pre-BN-crit, where every activation follows BN, it stayed between $1.66\times10^{-3}$ and $1.67\times10^{-3}$ for both degrees.
Moreover, the test accuracy of MRePU networks with BN was unstable even while their training accuracy increased steadily.
In Post-BN-$\gamma$-crit with $p=2$, it fell to chance level at intermediate or final epochs for two seeds and varied between $0.33$ and $0.73$ for the third,
and the test loss was non-finite in at least one epoch in every MRePU run of the two crit variants, including all trained runs.
Since evaluation uses the running BN statistics instead of batch statistics, this points to an interaction between the BN statistics and rare large pre-activations, but we have not isolated the cause.
In NF-$\gamma$, which has no BN, the test loss was finite and the test accuracy increased in every epoch for every MRePU run,
although the training loss averaged over an epoch showed transient spikes (up to $56$ for $p=2$ and \sci{2.3}{9} for $p=3$) from which training recovered.
For RePU, BN and clipping make the failure dependent on the run: in the learning-rate selection runs for Pre-BN-crit, RePU reached validation accuracies of $0.646$ and $0.706$ at two of the three learning rates,
but it stalled in all three final runs, although no choice of $C_W$ makes RePU critical (Section~3).
Finally, in a first round of selection runs (width 64, $2,2,2,2$ blocks) with $c_2K_0=0.3$ in Pre-BN-crit, $\tau=0.3$ in NF-$\gamma$ and no gradient clipping,
MRePU did not exceed a validation accuracy of $0.098$ in Pre-BN-crit and diverged at every learning rate in NF-$\gamma$; with the settings above, the same backbone reached $0.780$ in Pre-BN-crit.
Because $K_0$ (or $\tau$) and the gradient clipping were changed together, these runs do not show which of the two was responsible.
The instability analyzed in Sections~3 and~4 is therefore seen most directly without normalization:
in NF-$\gamma$ the outcome is decided at initialization, where RePU overflows within eight activation layers and MRePU initialized with Algorithm~\ref{alg:recipe} trains for every seed.
In normalized networks, the outcome depends in addition on the BN scale, on rare large pre-activations and on the BN statistics used at evaluation.

\section{Conclusion}

We analyzed deep networks with Rectified Power Unit (RePU) activations in the effective field theory of deep neural networks. For RePU of order $p\ge2$, the ratio of the parallel and perpendicular susceptibilities equals $(2p-1)/p$ for every kernel and every $C_W$, so no initialization makes a RePU network critical, and its kernel explodes or vanishes double-exponentially with depth. Experiments at initialization and during training confirmed this failure.

To remove it, we proposed the Modified Rectified Power Unit (MRePU), which keeps the differentiability of RePU and, like RePU, represents polynomials exactly with a single hidden layer, so that the approximation results for RePU networks carry over. Both susceptibilities of MRePU tend to $C_W$ as the kernel tends to zero, but at a finite kernel $K_0$ the kernel drifts over about $1/(c_pK_0)$ layers. We showed that the rescaled depth $\tau=c_pK_0L$ determines the kernel and the gradients at initialization (Proposition~10), that the neural tangent kernel leads to the same conclusions, and derived from this an initialization rule (Algorithm~1) that sets the input scale and $C_W$ from the depth, the degree and the input range. These results hold for the kernel recursion at initialization in the limit of infinite width.

The experiments separate trainability from accuracy. In regression, MRePU reached the lowest error on polynomial targets but not on piecewise-defined targets. On MNIST, the fixed point of the kernel recursion marked the boundary of divergence for networks with five layers, and MRePU networks initialized by Algorithm~1 with $\tau=10^{-2}$ trained at every depth up to 32 layers, whereas RePU networks failed from eight layers on. In physics-informed neural networks for four PDEs, all MRePU networks trained without divergence, while RePU networks diverged from five hidden layers on. On CIFAR-10 with six residual architectures, RePU trained in none, and MRePU trained in the architectures that keep the kernel at its inputs small but not in the standard post- and pre-activation ResNets. Where MRePU trained, it was less accurate than the best standard activation: slightly on MNIST, by up to about a factor of five relative to GELU and SiLU on the Poisson and Helmholtz equations at five and seven hidden layers, by up to an order of magnitude on the Burgers and Allen--Cahn equations, and by $0.10$--$0.15$ relative to ReLU in the same architecture on CIFAR-10 after five epochs.


\newpage

\appendix
\section{Additional Results for the PINN Experiments}\label{app:pinn}
\subsection{Seed Variability and Learned Solutions}\label{app:pinn-detail}
\begin{figure}[htp]
\centering
\includegraphics[width=\textwidth]{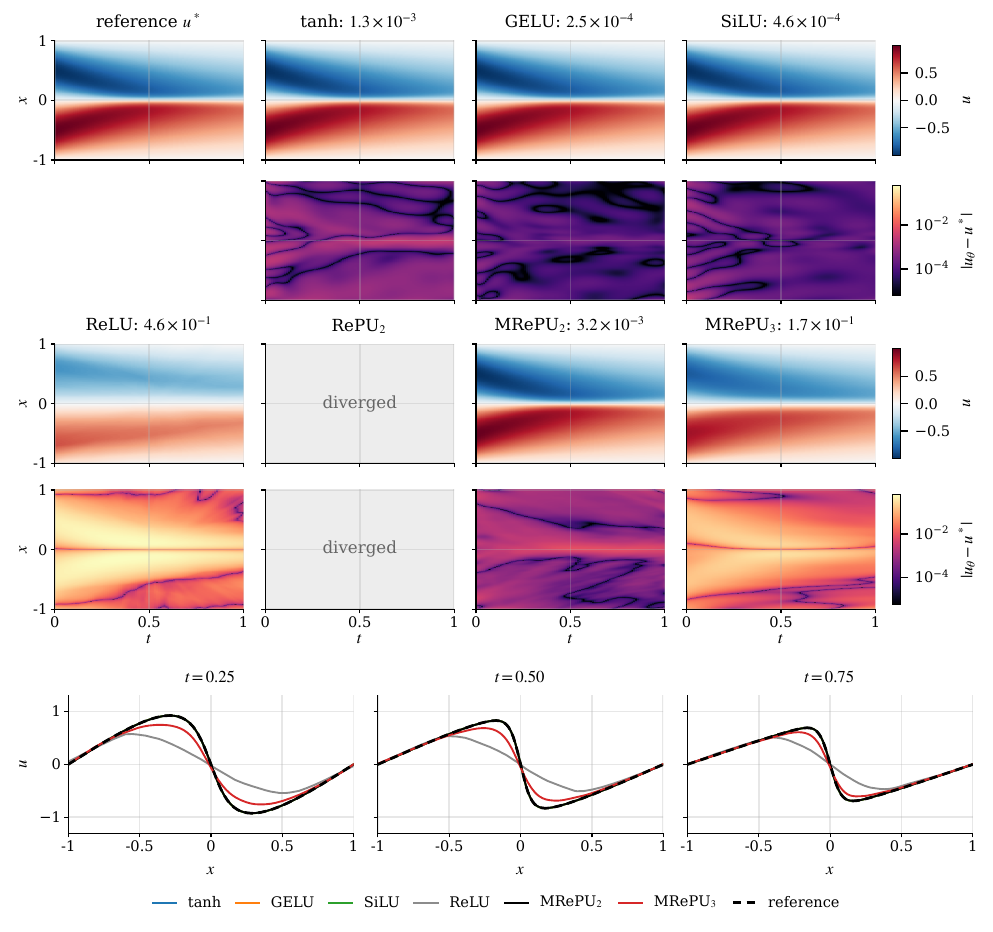}
\caption{PINN solutions of the Burgers equation ($L=5$). For each activation, the seed with the median
relative $L^2$ error (shown in the panel title) is plotted at the lowest-training-loss checkpoint.
Rows alternate between the prediction $u_\theta$ and the pointwise error $|u_\theta-u^*|$ (common log
scale); the bottom row shows cross-sections at $t=0.25, 0.5, 0.75$ with the reference dashed.
``Diverged'' as in Table~\ref{tab:pinn}.}
\label{fig:pinn-sol-burgers}
\end{figure}

\begin{figure}[htp]
\centering
\includegraphics[width=\textwidth]{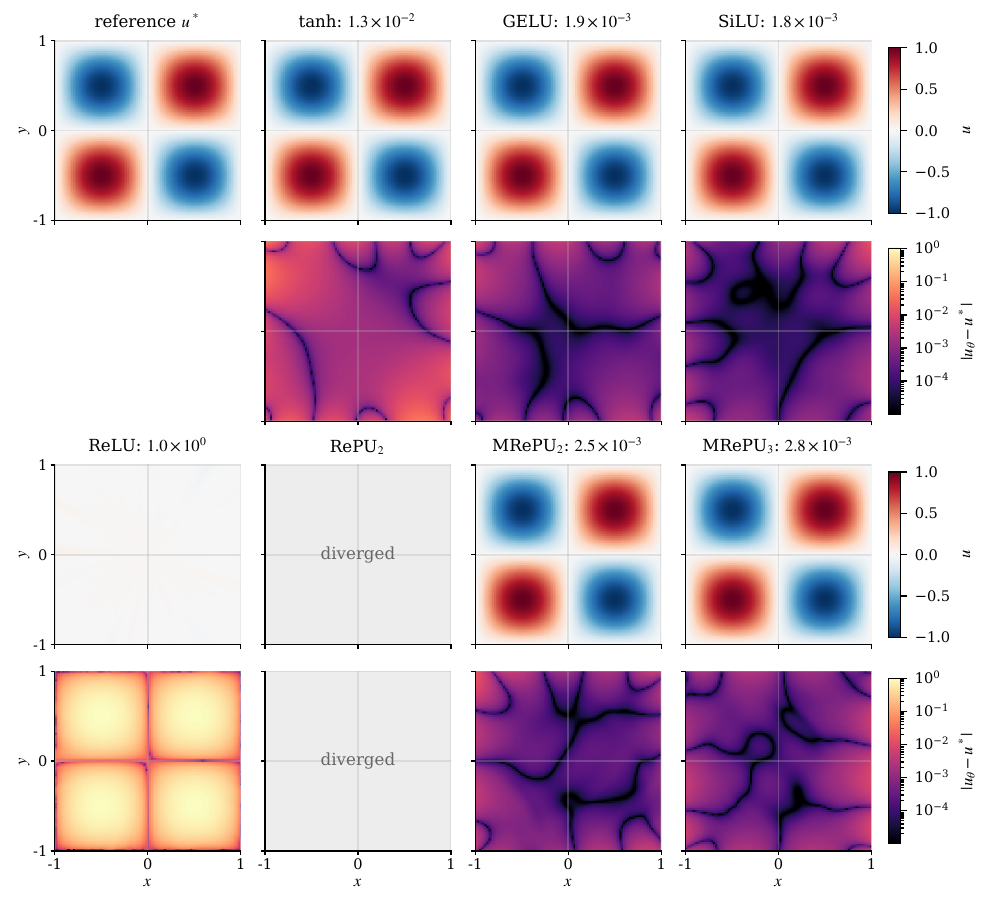}
\caption{As Figure~\ref{fig:pinn-sol-burgers}, for the Poisson equation.}
\label{fig:pinn-sol-poisson}
\end{figure}

\begin{figure}[htp]
\centering
\includegraphics[width=\textwidth]{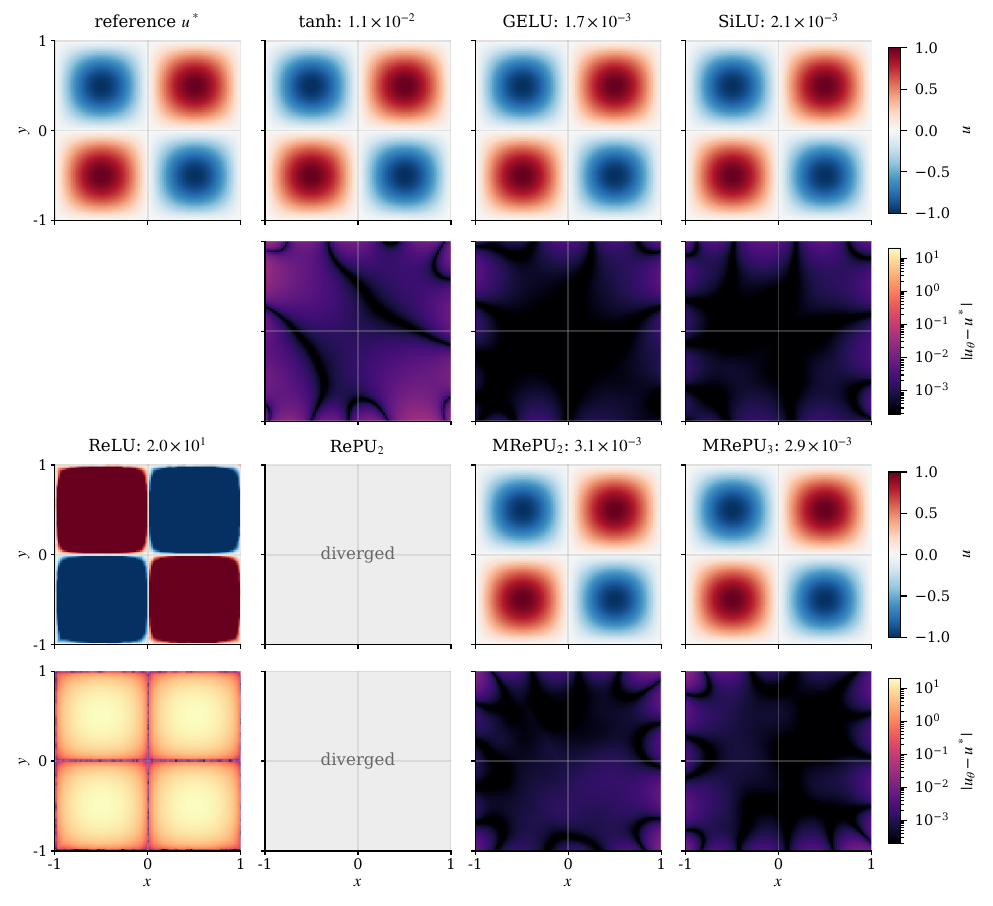}
\caption{As Figure~\ref{fig:pinn-sol-burgers}, for the Helmholtz equation. The ReLU network
converges to the trivial solution $u=q/k^2$, whose values lie outside the colour range.}
\label{fig:pinn-sol-helmholtz}
\end{figure}

\begin{figure}[htp]
\centering
\includegraphics[width=\textwidth]{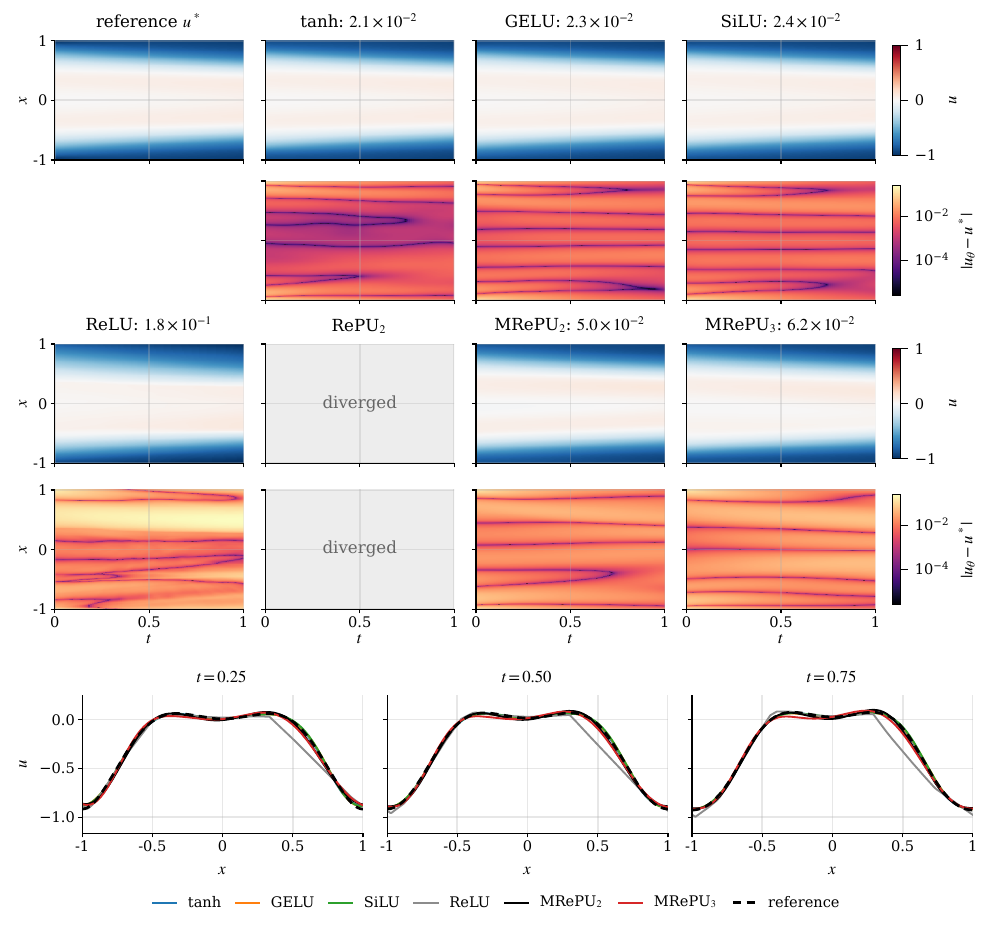}
\caption{As Figure~\ref{fig:pinn-sol-burgers}, for the Allen--Cahn equation.}
\label{fig:pinn-sol-ac}
\end{figure}

Table~\ref{tab:pinn-range} gives the range over seeds of the errors in Table~\ref{tab:pinn}, and
Figures~\ref{fig:pinn-sol-burgers}--\ref{fig:pinn-sol-ac} show the learned solutions and pointwise errors
for $L = 5$.
\begin{sidewaystable}
\centering
\caption{As Table~\ref{tab:pinn}, with the minimum and maximum over the 3 seeds in brackets.}\label{tab:pinn-range}
\footnotesize
\setlength{\tabcolsep}{3pt}
\begin{tabular}{llccccccc}
\toprule
PDE & Depth & tanh & GELU & SiLU & ReLU & RePU $p{=}2$ & MRePU $p{=}2$ & MRePU $p{=}3$ \\
\midrule
\multirow{3}{*}{Burgers} & 3 & {\boldmath\bfseries \shortstack{\sci{1.6}{-3}\\[-1pt] \tiny\mdseries\unboldmath[\sci{1.4}{-3}, \sci{1.6}{-3}]}} & \shortstack{\sci{1.7}{-3}\\[-1pt] \tiny\mdseries\unboldmath[\sci{1.1}{-3}, \sci{2.8}{-3}]} & \shortstack{\sci{4.0}{-3}\\[-1pt] \tiny\mdseries\unboldmath[\sci{2.0}{-3}, \sci{4.2}{-3}]} & \shortstack{\sci{4.6}{-1}\\[-1pt] \tiny\mdseries\unboldmath[\sci{4.4}{-1}, \sci{4.7}{-1}]} & \shortstack{\sci{8.0}{-2}\\[-1pt] \tiny\mdseries\unboldmath[\sci{3.8}{-2}, \sci{1.1}{-1}]} & \shortstack{\sci{3.0}{-1}\\[-1pt] \tiny\mdseries\unboldmath[\sci{2.8}{-1}, \sci{3.5}{-1}]} & \shortstack{\sci{9.0}{-2}\\[-1pt] \tiny\mdseries\unboldmath[\sci{6.7}{-2}, \sci{2.0}{-1}]} \\
 & 5 & \shortstack{\sci{1.3}{-3}\\[-1pt] \tiny\mdseries\unboldmath[\sci{6.1}{-4}, \sci{1.8}{-3}]} & {\boldmath\bfseries \shortstack{\sci{2.5}{-4}\\[-1pt] \tiny\mdseries\unboldmath[\sci{1.7}{-4}, \sci{3.0}{-4}]}} & \shortstack{\sci{4.6}{-4}\\[-1pt] \tiny\mdseries\unboldmath[\sci{4.3}{-4}, \sci{8.6}{-4}]} & \shortstack{\sci{4.6}{-1}\\[-1pt] \tiny\mdseries\unboldmath[\sci{4.5}{-1}, \sci{4.8}{-1}]} & div.$^{b}$ (3/3) & \shortstack{\sci{3.2}{-3}\\[-1pt] \tiny\mdseries\unboldmath[\sci{3.1}{-3}, \sci{3.9}{-3}]} & \shortstack{\sci{1.7}{-1}\\[-1pt] \tiny\mdseries\unboldmath[\sci{1.6}{-1}, \sci{2.0}{-1}]} \\
 & 7 & \shortstack{\sci{9.8}{-4}\\[-1pt] \tiny\mdseries\unboldmath[\sci{5.8}{-4}, \sci{1.1}{-3}]} & \shortstack{\sci{3.2}{-4}\\[-1pt] \tiny\mdseries\unboldmath[\sci{1.4}{-4}, \sci{4.7}{-4}]} & {\boldmath\bfseries \shortstack{\sci{3.2}{-4}\\[-1pt] \tiny\mdseries\unboldmath[\sci{2.3}{-4}, \sci{3.7}{-4}]}} & \shortstack{\sci{4.7}{-1}\\[-1pt] \tiny\mdseries\unboldmath[\sci{4.6}{-1}, \sci{4.7}{-1}]} & NaN$^{a}$ & \shortstack{\sci{2.3}{-3}\\[-1pt] \tiny\mdseries\unboldmath[\sci{1.3}{-3}, \sci{4.1}{-3}]} & \shortstack{\sci{1.9}{-1}\\[-1pt] \tiny\mdseries\unboldmath[\sci{7.0}{-3}, \sci{2.1}{-1}]} \\
\midrule
\multirow{3}{*}{Poisson} & 3 & \shortstack{\sci{1.2}{-2}\\[-1pt] \tiny\mdseries\unboldmath[\sci{9.4}{-3}, \sci{1.9}{-2}]} & \shortstack{\sci{2.9}{-3}\\[-1pt] \tiny\mdseries\unboldmath[\sci{1.5}{-3}, \sci{3.0}{-3}]} & {\boldmath\bfseries \shortstack{\sci{1.7}{-3}\\[-1pt] \tiny\mdseries\unboldmath[\sci{1.6}{-3}, \sci{2.1}{-3}]}} & \shortstack{\sci{1.0}{0}\\[-1pt] \tiny\mdseries\unboldmath[\sci{1.0}{0}, \sci{1.0}{0}]} & \shortstack{\sci{4.8}{-2}\\[-1pt] \tiny\mdseries\unboldmath[\sci{4.1}{-2}, \sci{1.2}{-1}]} & \shortstack{\sci{7.7}{-2}\\[-1pt] \tiny\mdseries\unboldmath[\sci{2.1}{-2}, \sci{1.1}{-1}]} & \shortstack{\sci{1.5}{-2}\\[-1pt] \tiny\mdseries\unboldmath[\sci{1.0}{-2}, \sci{1.8}{-2}]} \\
 & 5 & \shortstack{\sci{1.3}{-2}\\[-1pt] \tiny\mdseries\unboldmath[\sci{1.3}{-2}, \sci{2.1}{-2}]} & \shortstack{\sci{1.9}{-3}\\[-1pt] \tiny\mdseries\unboldmath[\sci{1.6}{-3}, \sci{2.0}{-3}]} & {\boldmath\bfseries \shortstack{\sci{1.8}{-3}\\[-1pt] \tiny\mdseries\unboldmath[\sci{1.7}{-3}, \sci{2.8}{-3}]}} & \shortstack{\sci{1.0}{0}\\[-1pt] \tiny\mdseries\unboldmath[\sci{1.0}{0}, \sci{1.0}{0}]} & div.$^{b}$ (2/3) & \shortstack{\sci{2.5}{-3}\\[-1pt] \tiny\mdseries\unboldmath[\sci{2.1}{-3}, \sci{2.5}{-3}]} & \shortstack{\sci{2.8}{-3}\\[-1pt] \tiny\mdseries\unboldmath[\sci{2.2}{-3}, \sci{4.8}{-3}]} \\
 & 7 & \shortstack{\sci{1.6}{-2}\\[-1pt] \tiny\mdseries\unboldmath[\sci{1.6}{-2}, \sci{2.9}{-2}]} & {\boldmath\bfseries \shortstack{\sci{1.2}{-3}\\[-1pt] \tiny\mdseries\unboldmath[\sci{6.8}{-4}, \sci{1.4}{-3}]}} & \shortstack{\sci{2.3}{-3}\\[-1pt] \tiny\mdseries\unboldmath[\sci{1.9}{-3}, \sci{2.7}{-3}]} & \shortstack{\sci{1.0}{0}\\[-1pt] \tiny\mdseries\unboldmath[\sci{1.0}{0}, \sci{1.0}{0}]} & NaN$^{a}$ & \shortstack{\sci{6.2}{-3}\\[-1pt] \tiny\mdseries\unboldmath[\sci{5.1}{-3}, \sci{6.9}{-3}]} & \shortstack{\sci{7.1}{-3}\\[-1pt] \tiny\mdseries\unboldmath[\sci{6.1}{-3}, \sci{8.3}{-3}]} \\
\midrule
\multirow{3}{*}{Helmholtz} & 3 & \shortstack{\sci{9.9}{-3}\\[-1pt] \tiny\mdseries\unboldmath[\sci{8.9}{-3}, \sci{1.6}{-2}]} & \shortstack{\sci{3.0}{-3}\\[-1pt] \tiny\mdseries\unboldmath[\sci{1.7}{-3}, \sci{3.2}{-3}]} & {\boldmath\bfseries \shortstack{\sci{1.4}{-3}\\[-1pt] \tiny\mdseries\unboldmath[\sci{1.4}{-3}, \sci{1.6}{-3}]}} & \shortstack{\sci{2.0}{1}\\[-1pt] \tiny\mdseries\unboldmath[\sci{2.0}{1}, \sci{2.0}{1}]} & \shortstack{\sci{6.4}{-2}\\[-1pt] \tiny\mdseries\unboldmath[\sci{5.9}{-2}, \sci{1.1}{-1}]} & \shortstack{\sci{4.0}{-2}\\[-1pt] \tiny\mdseries\unboldmath[\sci{3.7}{-2}, \sci{8.0}{-2}]} & \shortstack{\sci{1.3}{-2}\\[-1pt] \tiny\mdseries\unboldmath[\sci{7.3}{-3}, \sci{3.1}{-2}]} \\
 & 5 & \shortstack{\sci{1.1}{-2}\\[-1pt] \tiny\mdseries\unboldmath[\sci{8.8}{-3}, \sci{1.9}{-2}]} & {\boldmath\bfseries \shortstack{\sci{1.7}{-3}\\[-1pt] \tiny\mdseries\unboldmath[\sci{1.4}{-3}, \sci{3.9}{-3}]}} & \shortstack{\sci{2.1}{-3}\\[-1pt] \tiny\mdseries\unboldmath[\sci{1.8}{-3}, \sci{2.1}{-3}]} & \shortstack{\sci{2.0}{1}\\[-1pt] \tiny\mdseries\unboldmath[\sci{2.0}{1}, \sci{2.0}{1}]} & div.$^{b}$ (2/3) & \shortstack{\sci{3.1}{-3}\\[-1pt] \tiny\mdseries\unboldmath[\sci{2.8}{-3}, \sci{4.0}{-3}]} & \shortstack{\sci{2.9}{-3}\\[-1pt] \tiny\mdseries\unboldmath[\sci{2.9}{-3}, \sci{3.5}{-3}]} \\
 & 7 & \shortstack{\sci{1.7}{-2}\\[-1pt] \tiny\mdseries\unboldmath[\sci{1.5}{-2}, \sci{3.1}{-2}]} & {\boldmath\bfseries \shortstack{\sci{1.4}{-3}\\[-1pt] \tiny\mdseries\unboldmath[\sci{1.1}{-3}, \sci{2.5}{-3}]}} & \shortstack{\sci{2.1}{-3}\\[-1pt] \tiny\mdseries\unboldmath[\sci{1.5}{-3}, \sci{2.8}{-3}]} & \shortstack{\sci{2.0}{1}\\[-1pt] \tiny\mdseries\unboldmath[\sci{2.0}{1}, \sci{2.0}{1}]} & NaN$^{a}$ & \shortstack{\sci{5.7}{-3}\\[-1pt] \tiny\mdseries\unboldmath[\sci{4.2}{-3}, \sci{7.9}{-3}]} & \shortstack{\sci{1.0}{-2}\\[-1pt] \tiny\mdseries\unboldmath[\sci{6.2}{-3}, \sci{1.2}{-2}]} \\
\midrule
\multirow{3}{*}{Allen--Cahn} & 3 & {\boldmath\bfseries \shortstack{\sci{1.5}{-2}\\[-1pt] \tiny\mdseries\unboldmath[\sci{1.4}{-2}, \sci{1.9}{-2}]}} & \shortstack{\sci{1.8}{-2}\\[-1pt] \tiny\mdseries\unboldmath[\sci{1.2}{-2}, \sci{2.0}{-2}]} & \shortstack{\sci{2.7}{-2}\\[-1pt] \tiny\mdseries\unboldmath[\sci{2.2}{-2}, \sci{3.0}{-2}]} & \shortstack{\sci{2.1}{-1}\\[-1pt] \tiny\mdseries\unboldmath[\sci{1.7}{-1}, \sci{2.3}{-1}]} & \shortstack{\sci{4.1}{-2}\\[-1pt] \tiny\mdseries\unboldmath[\sci{3.2}{-2}, \sci{6.4}{-2}]} & \shortstack{\sci{1.5}{-1}\\[-1pt] \tiny\mdseries\unboldmath[\sci{1.0}{-1}, \sci{1.6}{-1}]} & \shortstack{\sci{1.5}{-1}\\[-1pt] \tiny\mdseries\unboldmath[\sci{1.1}{-1}, \sci{3.0}{-1}]} \\
 & 5 & {\boldmath\bfseries \shortstack{\sci{2.1}{-2}\\[-1pt] \tiny\mdseries\unboldmath[\sci{2.0}{-2}, \sci{2.8}{-2}]}} & \shortstack{\sci{2.3}{-2}\\[-1pt] \tiny\mdseries\unboldmath[\sci{1.5}{-2}, \sci{2.5}{-2}]} & \shortstack{\sci{2.4}{-2}\\[-1pt] \tiny\mdseries\unboldmath[\sci{1.4}{-2}, \sci{2.7}{-2}]} & \shortstack{\sci{1.8}{-1}\\[-1pt] \tiny\mdseries\unboldmath[\sci{1.5}{-1}, \sci{2.3}{-1}]} & NaN$^{a}$ & \shortstack{\sci{5.0}{-2}\\[-1pt] \tiny\mdseries\unboldmath[\sci{4.4}{-2}, \sci{5.6}{-2}]} & \shortstack{\sci{6.2}{-2}\\[-1pt] \tiny\mdseries\unboldmath[\sci{3.3}{-2}, \sci{2.4}{-1}]} \\
 & 7 & \shortstack{\sci{2.8}{-2}\\[-1pt] \tiny\mdseries\unboldmath[\sci{2.7}{-2}, \sci{2.8}{-2}]} & {\boldmath\bfseries \shortstack{\sci{1.2}{-2}\\[-1pt] \tiny\mdseries\unboldmath[\sci{1.2}{-2}, \sci{1.8}{-2}]}} & \shortstack{\sci{1.8}{-2}\\[-1pt] \tiny\mdseries\unboldmath[\sci{1.2}{-2}, \sci{1.8}{-2}]} & \shortstack{\sci{1.4}{-1}\\[-1pt] \tiny\mdseries\unboldmath[\sci{1.2}{-1}, \sci{1.9}{-1}]} & NaN$^{a}$ & \shortstack{\sci{1.3}{-1}\\[-1pt] \tiny\mdseries\unboldmath[\sci{4.2}{-2}, \sci{3.3}{-1}]} & \shortstack{\sci{5.0}{-2}\\[-1pt] \tiny\mdseries\unboldmath[\sci{4.7}{-2}, \sci{7.5}{-1}]} \\
\bottomrule
\end{tabular}
\end{sidewaystable}

\subsection{Gradient-Enhanced PINNs}\label{app:gpinn}
To examine Sobolev-type regularization, we trained gradient-enhanced PINNs \citep{Yu22}, which add the
penalty $w\,\|\nabla_x r\|^2$ on the gradient of the PDE residual to the loss, i.e.\ they measure the
residual in an $H^1$-type norm; this differs from Sobolev training \citep{Czarnecki17}, which supervises
derivatives of the target. We used $w=10^{-2}$ for tanh, GELU and MRePU ($p=2,3$) on the Poisson and
Helmholtz equations with $L=5$ and $7$, and compared each run with the run without the penalty that
uses the same seed, collocation points and training rule. The penalty did not systematically reduce the
error (Table~\ref{tab:gpinn}): it did so in 19 of the 48 paired runs, and the geometric mean of the error
ratio was $0.91$ for GELU, $1.09$ for tanh, $1.19$ for MRePU$_2$ and $1.21$ for MRePU$_3$, with individual
ratios between $0.38$ and $2.2$, comparable to the variation across seeds without the penalty.
The ordering of the activations was unchanged. Since the plain PINN already reaches errors of
$10^{-3}$--$10^{-2}$ on these smooth solutions and only one penalty weight was tested, these results
do not allow conclusions about Sobolev-type regularization in general.
\begin{table}[htp]
\small
\setlength{\tabcolsep}{2.5pt}
\begin{tabular}{llcccccccc}
\toprule
 & & \multicolumn{2}{c}{tanh} & \multicolumn{2}{c}{GELU} & \multicolumn{2}{c}{MRePU$_2$} & \multicolumn{2}{c}{MRePU$_3$} \\
\cmidrule(lr){3-4}\cmidrule(lr){5-6}\cmidrule(lr){7-8}\cmidrule(lr){9-10}
PDE & $L$ & error & ratio & error & ratio & error & ratio & error & ratio \\
\midrule
\multirow{2}{*}{Poisson} & 5 & \sci{1.3}{-2} & 1.13 (1/3) & \sci{1.4}{-3} & 0.89 (2/3) & \sci{4.5}{-3} & 1.56 (1/3) & \sci{3.4}{-3} & 1.23 (0/3) \\
 & 7 & \sci{2.1}{-2} & 0.97 (1/3) & \sci{1.0}{-3} & 1.05 (1/3) & \sci{6.1}{-3} & 0.99 (2/3) & \sci{8.7}{-3} & 1.21 (0/3) \\
\midrule
\multirow{2}{*}{Helmholtz} & 5 & \sci{1.4}{-2} & 1.38 (0/3) & \sci{1.5}{-3} & 0.67 (3/3) & \sci{2.8}{-3} & 0.99 (2/3) & \sci{4.0}{-3} & 1.50 (0/3) \\
 & 7 & \sci{1.7}{-2} & 0.94 (2/3) & \sci{1.9}{-3} & 1.09 (1/3) & \sci{7.4}{-3} & 1.30 (1/3) & \sci{8.1}{-3} & 0.97 (2/3) \\
\bottomrule
\end{tabular}
\centering
\caption{gPINN ($w=10^{-2}$): relative $L^2$ error (median over 3 seeds) and the geometric mean over seeds of the
error ratio to the same run without the penalty.
In parentheses: seeds for which the penalty reduced the error.}
\label{tab:gpinn}
\end{table}

\section{Additional Results for the CIFAR-10 Experiments}\label{app:cifar}
Table~\ref{tab:cifar-sweep} gives the validation accuracies of the learning-rate selection runs (seed $0$, 8 epochs, 45{,}000/5{,}000 split)
for the backbone of Section~5.3.2 and for two variants: width 64 with $2,2,1,1$ blocks, and width 64 with $2,2,2,2$ blocks (18 weight layers and 17 activations on the longest path).
For each combination we give the best validation accuracy over the three learning rates.
The pattern of Table~\ref{tab:cifar} holds for all three backbones.
MRePU diverges in Post-BN and Post-BN-$\gamma$ and stays at chance level in Pre-BN; with $p=2$ it trains in Pre-BN-crit and NF-$\gamma$, and with $p=3$ in NF-$\gamma$ but not in Pre-BN-crit.
RePU diverges in NF-$\gamma$ and stays at chance level elsewhere, except in Pre-BN-crit at width 128.
GELU and SiLU lose $0.17$--$0.23$ in the crit variants relative to the corresponding architecture with unit BN scales.
 
For MRePU, the best accuracy was attained at the smallest learning rate in every combination in which it trained, except Post-BN-$\gamma$-crit with $p=2$ (best at $10^{-3}$).
At width 128, MRePU failed at $2\times10^{-3}$ (validation accuracy below $0.1$ or divergence) in all four combinations in which it trained at a smaller rate;
at width 64 it failed at $2\times10^{-3}$ with $2,2,2,2$ blocks, and with $2,2,1,1$ blocks its accuracy at $2\times10^{-3}$ was lower than at $5\times10^{-4}$ by $0.08$--$0.22$.
The learning rates selected for the runs in Table~\ref{tab:cifar} were $10^{-3}$ or $2\times10^{-3}$ for ReLU, GELU and SiLU,
and $5\times10^{-4}$ for MRePU except Post-BN-$\gamma$-crit with $p=2$ ($10^{-3}$).
 
\begin{table}[htp]
\centering
\scriptsize
\setlength{\tabcolsep}{4pt}
\begin{tabular}{llcccccc}
\toprule
 & & ReLU & GELU & SiLU & RePU & MRePU & MRePU \\
Backbone & Architecture & & & & $(p{=}2)$ & $(p{=}2)$ & $(p{=}3)$ \\
\midrule
\multirow{6}{*}{\shortstack[l]{width 128\\$2,2,1,1$}}
 & Post-BN               & $0.900$ & $0.906$ & $0.901$ & $0.099$ & div. & div. \\
 & Post-BN-$\gamma$      & $0.893$ & $0.902$ & $0.891$ & $0.100$ & div. & div. \\
 & Post-BN-$\gamma$-crit & $0.901$ & $0.731$ & $0.670$ & $0.100$ & $0.771$ & $0.096^{\ast}$ \\
 & Pre-BN                & $0.900$ & $0.907$ & $0.899$ & $0.101$ & $0.103$ & $0.103$ \\
 & Pre-BN-crit           & $0.899$ & $0.731$ & $0.698$ & $0.706$ & $0.786$ & $0.097$ \\
 & NF-$\gamma$           & $0.866$ & $0.906$ & $0.905$ & div. & $0.770^{\ast}$ & $0.737^{\ast}$ \\
\midrule
\multirow{6}{*}{\shortstack[l]{width 64\\$2,2,1,1$}}
 & Post-BN               & $0.892$ & $0.899$ & $0.893$ & $0.103$ & div. & div. \\
 & Post-BN-$\gamma$      & $0.887$ & $0.896$ & $0.886$ & $0.103$ & div. & div. \\
 & Post-BN-$\gamma$-crit & -- & -- & -- & -- & -- & -- \\
 & Pre-BN                & $0.893$ & $0.899$ & $0.892$ & $0.101$ & $0.103$ & $0.104$ \\
 & Pre-BN-crit           & $0.893$ & $0.726$ & $0.672$ & $0.099$ & $0.766$ & $0.097$ \\
 & NF-$\gamma$           & $0.853$ & $0.893$ & $0.888$ & div. & $0.787$ & $0.762$ \\
\midrule
\multirow{4}{*}{\shortstack[l]{width 64\\$2,2,2,2$}}
 & Post-BN               & $0.896$ & $0.904$ & $0.900$ & $0.102$ & div. & div. \\
 & Post-BN-$\gamma$      & $0.894$ & $0.897$ & $0.893$ & $0.102$ & div. & div. \\
 & Pre-BN                & $0.899$ & $0.902$ & $0.894$ & $0.098$ & $0.103$ & $0.104$ \\
 & Pre-BN-crit           & $0.896$ & $0.712$ & $0.668$ & $0.097$ & $0.780$ & $0.098$ \\
\bottomrule
\end{tabular}
\caption{Validation accuracy of the learning-rate selection runs on CIFAR-10 (seed $0$, 8 epochs), best over the learning rates $\{5\times10^{-4},10^{-3},2\times10^{-3}\}$.
div.: diverged at all three learning rates; $^{\ast}$diverged at one or two of them; --: not run.}
\label{tab:cifar-sweep}
\end{table}

\vskip 0.2in
\bibliography{sample}

@Article{rosenblatt58,
  author = 	 {F. Rosenblatt},
  title = 	 {The perceptron: A probabilistic model for information storage and organization in the brain},
  journal = 	 {Psychological Review},
  year = 	 {1958},
  volume = 	 {65},
  number = 	 {6},
  pages = 	 {386--408}}

@Article{cybenko89,
  author = 	 {G. Cybenko},
  title = 	 {Approximation by superpositions of a sigmoidal function},
  journal = 	 {Mathematics of Control and Signals and Systems},
  year = 	 {1989},
  volume = 	 {2},
  pages = 	 {303--315}}

@Article{Rumelhart86,
  author = 	 {D. E. Rumelhart and G. E. Hinton and R. J. Williams},
  title = 	 {Learning representations by back-propagating errors},
  journal = 	 {Nature},
  year = 	 {1986},
  volume = 	 {323},
  pages = 	 {533--536}}

@Article{Narayan97,
  author = 	 {S. Narayan},
  title = 	 {The generalized sigmoid activation function: Competitive supervised learning},
  journal = 	 {Information Sciences},
  year = 	 {1997},
  volume = 	 {99},
  number = {1-2},
  pages = 	 {69--82}}

@Article{Nair10,
  author = 	 {V. Nair and G. E. Hinton},
  title = 	 {Rectified Linear Units Improve Restricted Boltzmann Machines},
  journal = 	 {Proceedings of the 27th International Conference on Machine Learning (ICML)},
  year = 	 {2010}}

@Article{Glorot11,
  author = 	 {X. Glorot and A. Bordes and Y. Bengio},
  title = 	 {Deep Sparse Rectifier Neural Networks},
  journal = 	 {Proceedings of the 14th International Conference on Artificial Intelligence and Statistics (AISTATS)},
  year = 	 {2011}}

@Article{He15,
  author = 	 {K. He and X. Zhang and S. Ren and J. Sun},
  title = 	 {Delving Deep into Rectifiers: Surpassing Human-Level Performance on ImageNet Classification},
  journal = 	 {Proceedings of the IEEE International Conference on Computer Vision (ICCV)},
  year = 	 {2015}}

@Article{Hendrycks16,
  author = 	 {D. Hendrycks and K. Gimpel},
  title = 	 {Gaussian Error Linear Units (GELUs)},
  journal = 	 {Arxiv},
  year = 	 {2016}}

@Article{Clervert15,
  author = 	 {DA. Clevert and T. Unterthiner and S. Hochreiter},
  title = 	 {Fast and Accurate Deep Network Learning by Exponential Linear Units (ELUs)},
  journal = 	 {International Conference on Learning Representations (ICLR)},
  year = 	 {2015}}

@Article{Ramachandran17,
  author = 	 {P. Ramachandran and B. Zoph and Q. V. Le},
  title = 	 {Searching for Activation Functions},
  journal = 	 {Arxiv},
  year = 	 {2017}}

@Article{Sun24,
  author = 	 {H. Sun and Z. Wu and B. Xia and P. Chang and Z. Dong and Y. Yuan and Y. Chang and X. Wang},
  title = 	 {A Method on Searching Better Activation Functions},
  journal = 	 {Arxiv},
  year = 	 {2024}}

@Article{Li20,
  author = 	 {B. Li and S. Tang and H. Yu},
  title = 	 {PowerNet: Efficient Representations of Polynomials and Smooth Functions by Deep Neural Networks with Rectified Power Units},
  journal = 	 {J. Math. Study},
  year = 	 {2020},
  volume = 	 {53},
  pages = 	 {159--191}}

@Article{E18,
  author = 	 {W. E and B. Yu},
  title = 	 {The Deep Ritz Method: A Deep Learning-Based Numerical Algorithm for Solving Variational Problems},
  journal = 	 {Communications in Mathematics and Statistics},
  year = 	 {2018},
  volume = 	 {6},
  pages = 	 {1--12}}

@Article{Abdeljawad22,
  author = 	 {A. Abdeljawad and P. Grohs},
  title = 	 {Integral representations of shallow neural network with rectified power unit activation function},
  journal = 	 {Neural Networks},
  year = 	 {2022},
  volume = 	 {155},
  pages = 	 {536--550}}

@Article{Shen23,
  author = 	 {G. Shen and Y. Jiao and Y. Lin and and J. Huang},
  title = 	 {Differentiable Neural Networks with RePU Activation: with Applications to Score Estimation and Isotonic Regression},
  journal = 	 {Arxiv},
  year = 	 {2023}}

@Book{Neal96,
  author = 	 {R. M. Neal},
  title = 	 {Bayesian Learning for Neural Networks},
  publisher = 	 {Springer New York},
  year = 	 {1996},
  address = 	 {NY}
}

@Book{Rasmussen04,
  author = 	 {C. E. Rasmussen},
  title = 	 {Gaussian Processes in Machine Learning},
  publisher = 	 {Springer},
  year = 	 {2004},
  address = 	 {Berlin}
}

@Book{Roberts22,
  author = 	 {D. A. Roberts and S. Yaida and B. Hanin},
  title = 	 {The Principles of Deep Learning Theory},
  publisher = 	 {Cambridge University Press},
  year = 	 {2022}
}

@Book{LeCun02,
  author = 	 {Y. A. LeCun and L. Bottou and G. B. Orr and KR Müller},
  title = 	 {Efficient BackProp},
  publisher = 	 {Springer},
  year = 	 {2002},
  address = 	 {Berlin}
}

@Article{Parascandolo16,
  author = 	 {G. Parascandolo and H. Huttunen and T. Virtanen},
  title = 	 {Taming the waves: sine as activation function in deep neural networks},
  journal = 	 {Arxiv},
  year = 	 {2016}}

@Article{Sitzmann20,
  author = 	 {V. Sitzmann and J. N. P. Martel and A. W. Bergman and D. B. Lindell and G. Wetzstein},
  title = 	 {Implicit Neural Representations with Periodic Activation Functions},
  journal = 	 {Proceedings of the Neural Information Processing Systems Conference (NeurIPS)},
  year = 	 {2020}}

@Article{Maas13,
  author = 	 {A. L. Maas and A. Y. Hannun and A. Y. Ng},
  title = 	 {Rectifier Nonlinearities Improve Neural Network Acoustic Models},
  journal = 	 {Proceedings of the 30th International Conference on Machine Learning (ICML)},
  year = 	 {2013}}

@Book{Bishop06,
  author = 	 {C. M. Bishop},
  title = 	 {Pattern Recognition and Machine Learning},
  publisher = 	 {Springer New York},
  year = 	 {2006},
  address = 	 {NY}
}

@Article{Banta24,
  author = 	 {I. Banta and T. Cai and N. Craig and Z. Zhang},
  title = 	 {Structures of neural network effective theories},
  journal = 	 {Phys. Rev. D},
  year = 	 {2024},
  volume = 	 {109},
  pages = 	 {105007}}

@Article{Halverson21,
  author = 	 {J. Halverson and A. Maiti and K. Stoner},
  title = 	 {Neural networks and quantum field theory},
  journal = 	 {Mach. Learn.: Sci. Technol.},
  year = 	 {2021},
  volume = 	 {2},
  pages = 	 {035002}}

@Article{Raissi19,
  author = 	 {M. Raissi and P. Perdikaris and G.E. Karniadakis},
  title = 	 {Physics-informed neural networks: A deep learning framework for solving forward and inverse problems involving nonlinear partial differential equations},
  journal = 	 {Journal of Computational Physics},
  year = 	 {2019},
  volume = 	 {378},
  pages = 	 {686--707}}

@Article{Wang22,
  author = 	 {Sifan Wang and Xinling Yu and Paris Perdikaris},
  title = 	 {When and why PINNs fail to train: A neural tangent kernel perspective},
  journal = 	 {Journal of Computational Physics},
  year = 	 {2022},
  volume = 	 {449},
  pages = 	 {110768}}

@Article{Aditi21,
  author = 	 {Aditi S. Krishnapriyan and Amir Gholami and Shandian Zhe and Robert M. Kirby and Michael W. Mahoney},
  title = 	 {Characterizing possible failure modes in physics-informed neural networks},
  journal = 	 {Proceedings of the Neural Information Processing Systems Conference (NeurIPS)},
  year = 	 {2021}}

@Article{He15b,
  author = 	 {K. He and X. Zhang and S. Ren and J. Sun},
  title = 	 {Deep Residual Learning for Image Recognition},
  journal = 	 {Proceedings of the IEEE Conference on Computer Vision and Pattern Recognition (CVPR)},
  year = 	 {2016}}

@Article{Kassam05,
  author = 	 {A.-K. Kassam and L. N. Trefethen},
  title = 	 {Fourth-Order Time-Stepping for Stiff {PDE}s},
  journal = 	 {SIAM Journal on Scientific Computing},
  volume = 	 {26},
  number = 	 {4},
  pages = 	 {1214--1233},
  year = 	 {2005}}

@Article{Yu22,
  author = 	 {J. Yu and L. Lu and X. Meng and G. E. Karniadakis},
  title = 	 {Gradient-Enhanced Physics-Informed Neural Networks for Forward and Inverse {PDE} Problems},
  journal = 	 {Computer Methods in Applied Mechanics and Engineering},
  volume = 	 {393},
  pages = 	 {114823},
  year = 	 {2022}}

@InProceedings{Czarnecki17,
  author = 	 {W. M. Czarnecki and S. Osindero and M. Jaderberg and G. Swirszcz and R. Pascanu},
  title = 	 {Sobolev Training for Neural Networks},
  booktitle = {Advances in Neural Information Processing Systems},
  volume = 	 {30},
  year = 	 {2017}}

@inproceedings{He16identity,
  author    = {He, Kaiming and Zhang, Xiangyu and Ren, Shaoqing and Sun, Jian},
  title     = {Identity Mappings in Deep Residual Networks},
  booktitle = {European Conference on Computer Vision (ECCV)},
  series    = {Lecture Notes in Computer Science},
  volume    = {9908},
  pages     = {630--645},
  publisher = {Springer},
  year      = {2016}
}

@inproceedings{Loshchilov19,
  author    = {Loshchilov, Ilya and Hutter, Frank},
  title     = {Decoupled Weight Decay Regularization},
  booktitle = {International Conference on Learning Representations (ICLR)},
  year      = {2019}
}

@inproceedings{Jacot18, author={Jacot, Arthur and Gabriel, Franck and Hongler, Cl{\'e}ment},
title={Neural Tangent Kernel: Convergence and Generalization in Neural Networks},
booktitle={Advances in Neural Information Processing Systems (NeurIPS)}, volume={31}, year={2018}}

@inproceedings{HaninNica20, author={Hanin, Boris and Nica, Mihai},
title={Finite Depth and Width Corrections to the Neural Tangent Kernel},
booktitle={International Conference on Learning Representations (ICLR)}, year={2020}}

@article{Leshno93,
  author  = {Leshno, Moshe and Lin, Vladimir Ya. and Pinkus, Allan and Schocken, Shimon},
  title   = {Multilayer feedforward networks with a nonpolynomial activation function can approximate any function},
  journal = {Neural Networks},
  volume  = {6},
  number  = {6},
  pages   = {861--867},
  year    = {1993},
  doi     = {10.1016/S0893-6080(05)80131-5}
}
\nocite{Abdeljawad22,Aditi21,Banta24,Bishop06,Clervert15,cybenko89,E18,Glorot11,Halverson21,He15,Hendrycks16,LeCun02,Li20,Maas13,Nair10,Narayan97,Neal96,Parascandolo16,Raissi19,Ramachandran17,Rasmussen04,Roberts22,rosenblatt58,Rumelhart86,Shen23,Sitzmann20,Sun24,Wang22}

\end{document}